\documentclass[10pt]{article} 
\usepackage[preprint]{tmlr}

\usepackage{tmlr}
\usepackage{graphicx}
\usepackage{subcaption}
\usepackage{booktabs}
\usepackage{color}
\usepackage{lipsum}
\usepackage{xspace}
\usepackage{threeparttable}
\usepackage{todonotes}
\usepackage{multirow}
\usepackage{enumitem}
\usepackage{wrapfig}

\definecolor{cobalt}{rgb}{0.0, 0.0, 0.75}
\usepackage[hidelinks,colorlinks=true,linkcolor=cobalt,citecolor=cobalt]{hyperref}

\newcommand{\layernorm}{layernorm\xspace}
\newcommand{\cifarcaps}{cifar-captions\xspace}

\newcommand{\bx}{{\pmb{x}}}

\newcommand{\bc}{{\pmb{c}}}

\newcommand{\fid}{FID\xspace}
\newcommand{\clipfid}{Clip-FID\xspace}
\newcommand{\clipscore}{Clip-score\xspace}
\newcommand{\cifarcap}{cifar-captions\xspace}
\newcommand{\flops}{FLOPs\xspace}
\newcommand{\geneval}{{GenEval}\xspace}

\newcommand{\ourdit}{{\texttt{MicroDiT}}\xspace}

\newcommand{\eightH}{{8$\times$H100}\xspace}
\newcommand{\eightA}{{8$\times$A100}\xspace}

\newcommand{\dittiny}{{\texttt{DiT-Tiny/2}}\xspace}
\newcommand{\ditxl}{{\texttt{DiT-Xl/2}}\xspace}
\newcommand{\ourmasking}{{deferred}\xspace}

\newcommand{\beps}{{\pmb{\epsilon}}}
\newcommand{\hyph}{{$-$}}

\usepackage{amsmath,amsfonts,bm}

\def\eqref#1{equation~\ref{#1}}

\def\1{\bm{1}}

\DeclareMathAlphabet{\mathsfit}{\encodingdefault}{\sfdefault}{m}{sl}
\SetMathAlphabet{\mathsfit}{bold}{\encodingdefault}{\sfdefault}{bx}{n}

\author{{{Vikash Sehwag$^1$\thanks{Corresponding author: \textit{vikash.sehwag@sony.com}}~, Xianghao Kong$^2$, Jingtao Li$^1$, Michael Spranger$^1$, Lingjuan Lyu$^1$}}}
\newcommand{\affiliations}{$^1$ Sony AI, $^2$ University of California, Riverside}

\def\month{MM}  
\def\year{YYYY} 
\def\openreview{\url{https://openreview.net/forum?id=XXXX}} 

\title{Stretching Each Dollar: Diffusion Training from Scratch on a Micro-Budget}
\date{March 2024}

\begin{document}

\maketitle
\vspace{-25pt}
{\small \affiliations}

\begin{abstract}
    As scaling laws in generative AI push performance, they also simultaneously concentrate the development of these models among actors with large computational resources. With a focus on text-to-image (T2I) generative models, we aim to address this bottleneck by demonstrating very low-cost training of large-scale T2I diffusion transformer models. As the computational cost of transformers increases with the number of patches in each image, we propose to randomly mask up to 75\% of the image patches during training. We propose a deferred masking strategy that preprocesses all patches using a patch-mixer before masking, thus significantly reducing the performance degradation with masking, making it superior to model downscaling in reducing computational cost. We also incorporate the latest improvements in transformer architecture, such as the use of mixture-of-experts layers, to improve performance and further identify the critical benefit of using synthetic images in micro-budget training. Finally, using only 37M publicly available real and synthetic images, we train a 1.16 billion parameter sparse transformer with only \textit{\$1,890} economical cost and achieve a 12.7 FID in zero-shot generation on the COCO dataset. Notably, our model achieves competitive FID and high-quality generations while incurring 118$\times$ lower cost than stable diffusion models and 14$\times$ lower cost than the current state-of-the-art approach that costs \$28,400. We aim to release our end-to-end training pipeline to further democratize the training of large-scale diffusion models on micro-budgets.
\end{abstract}
\begin{figure}
    \centering
     \begin{subfigure}[b]{0.495\linewidth}
        \centering
        \scalebox{-1}[1]{
        \includegraphics[width=\linewidth]{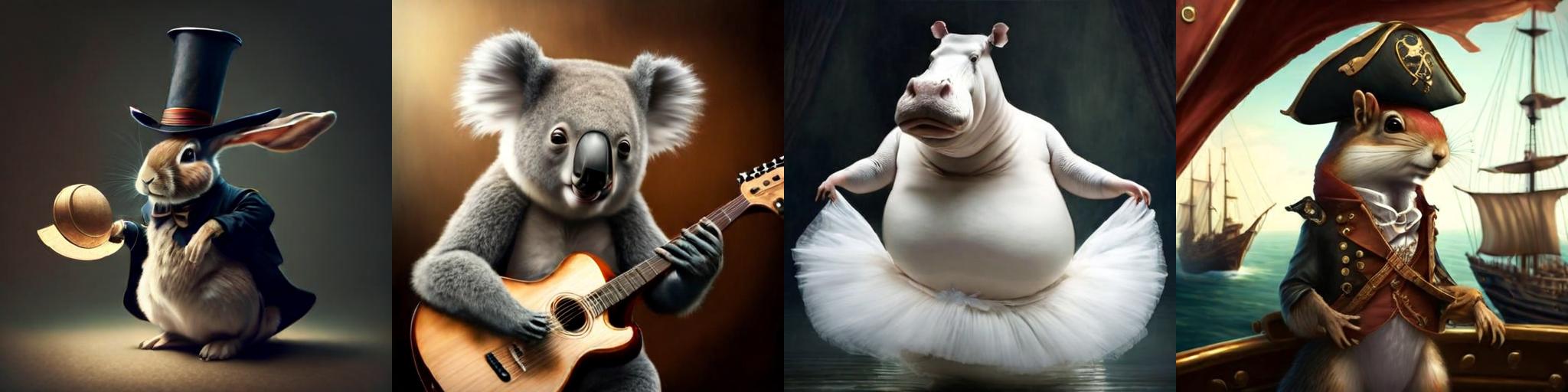}}
        \vspace{-18pt}
        \caption*{{\tiny An elegant squirrel pirate on a ship.; A hippo in a ballet performance in a white swan dress; A photo realistic koala bear playing a guitar; A rabbit magician pulling a hat. \par}}
    \end{subfigure}
    \hfill
    \begin{subfigure}[b]{0.495\linewidth}
        \centering
        \includegraphics[width=\linewidth]{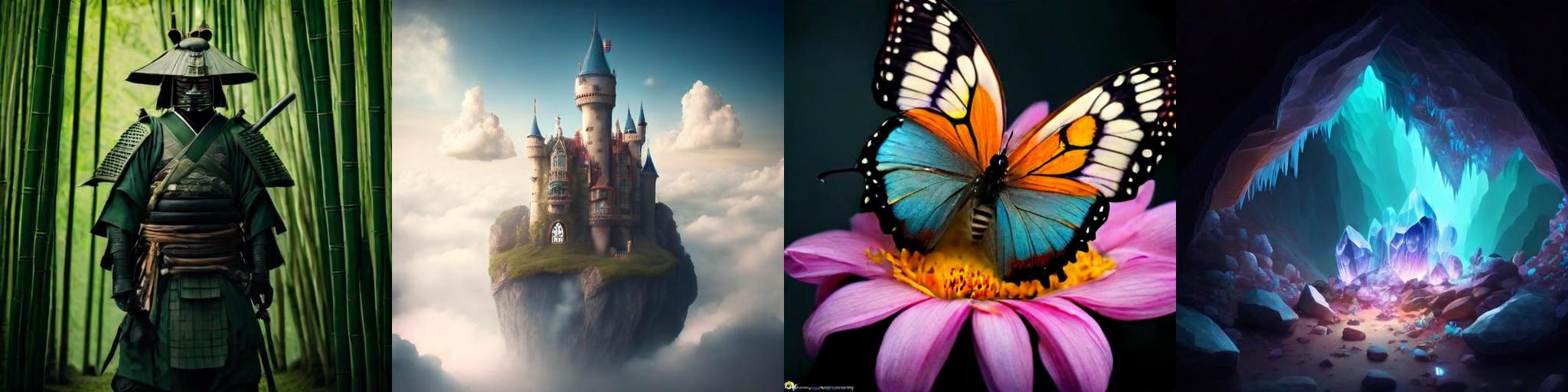}
        \vspace{-18pt}
        \caption*{{\tiny A real life samurai standing in a green bamboo grove; A fairy tale castle in the clouds; A real looking vibrant butterfly on a photorealistic flower; A mystical cave with glowing crystals.\par}}
    \end{subfigure}
    \begin{subfigure}[b]{0.495\linewidth}
        \centering
        \includegraphics[width=\linewidth]{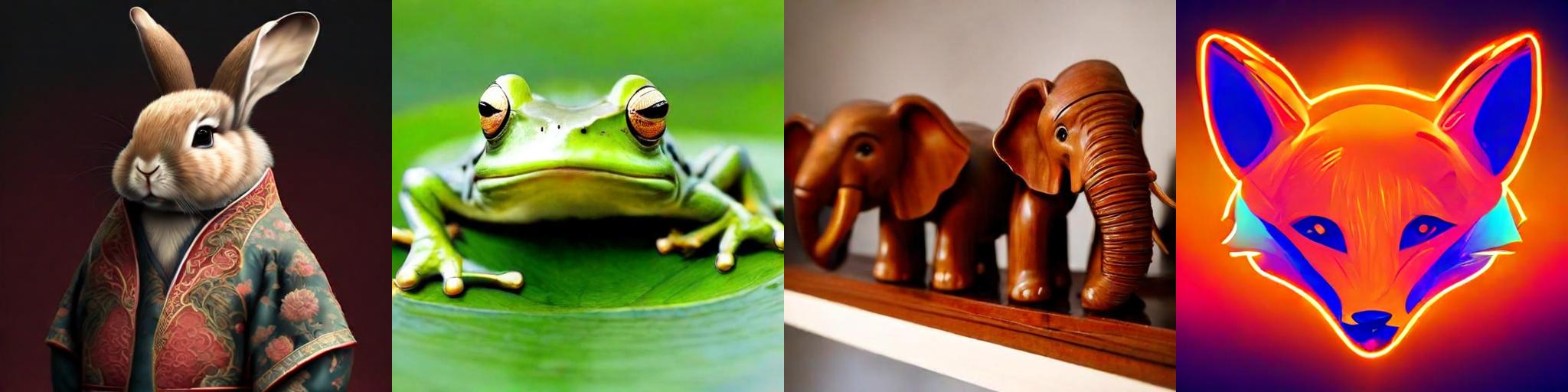}
        \vspace{-18pt}
        \caption*{{\tiny A cute rabbit in a stunning, detailed chinese coat; A frog on a lotus leaf.; Toy elephants made of Mahogany wood on a shelf; A logo of a fox in neon light.\par}}
    \end{subfigure}
    \hfill
    \begin{subfigure}[b]{0.495\linewidth}
        \centering
        \includegraphics[width=\linewidth]{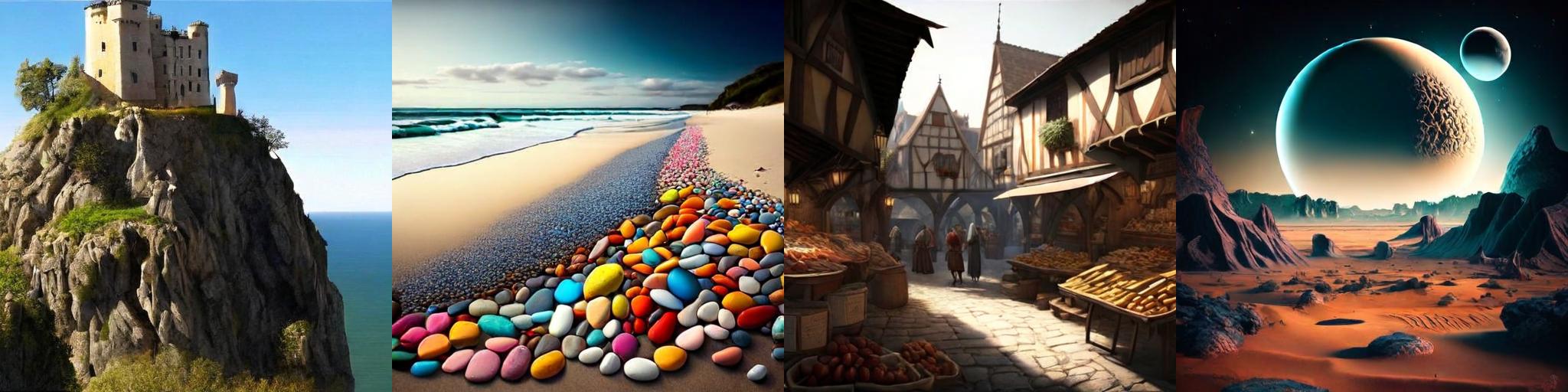}
        \vspace{-18pt}
        \caption*{{\tiny Ancient castle of a cliff edge; Serene beach with colourful pebble stones; Photo realistic market place in a medieval town; Alien landscape with two moons.\par}}
    \end{subfigure}

    \begin{subfigure}[b]{0.495\linewidth}
        \centering
        \includegraphics[width=\linewidth]{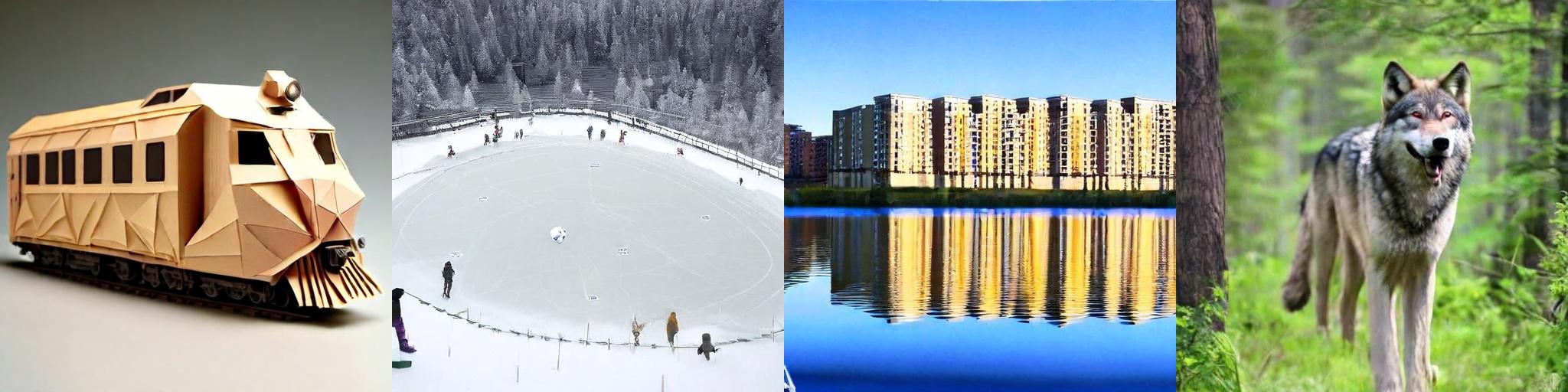}
        \vspace{-18pt}
        \caption*{{\tiny A realistic train made of origami.; A soccer plaground made of snow and timber.; Pre war apartment complexes reflected in the water of a beautifull lake.; A wolf coming out of the woods.\par}}
    \end{subfigure}
    \hfill
    \begin{subfigure}[b]{0.495\linewidth}
        \centering
        \includegraphics[width=\linewidth]{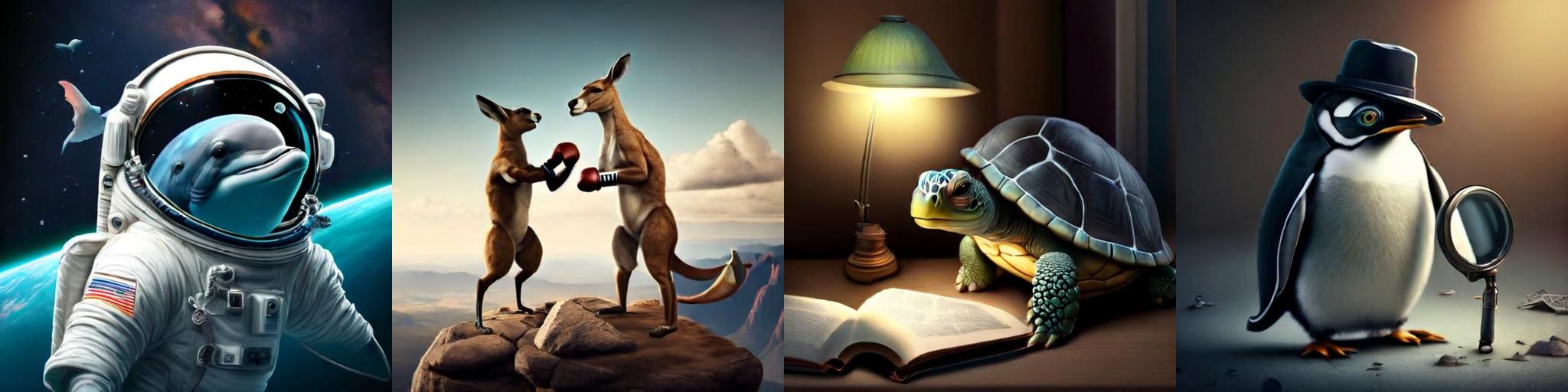}
        \vspace{-18pt}
        \caption*{{\tiny A dolphin astronaut in space in astronaut suit; Kangaroo boxing champions on a mountain top; A photo realistic turtle reading a book using a table lamp; A realistic penguin detective with magnifying glass in a crime scene.\par}}
    \end{subfigure}
    \begin{subfigure}[b]{0.495\linewidth}
        \centering
        \includegraphics[width=\linewidth]{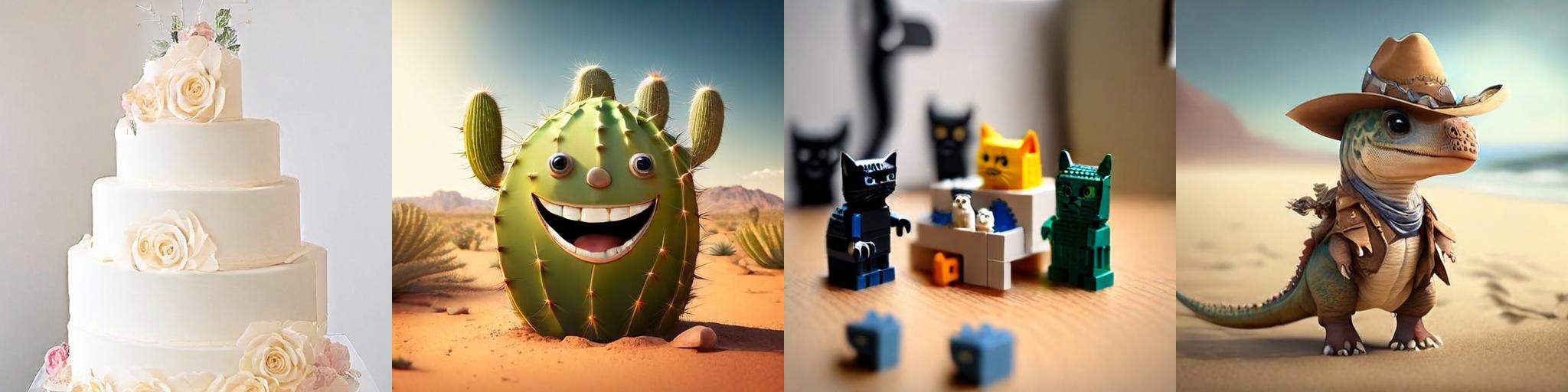}
        \vspace{-18pt}
        \caption*{{\tiny A four layer wedding cake; A cactus smiling in a desert; Miniature cat made of lego playing with other cats in a room; An adorable baby dinosaur dressed as cowboy on a beach\par}}
    \end{subfigure}
    \hfill
    \begin{subfigure}[b]{0.495\linewidth}
        \centering
        \includegraphics[width=\linewidth]{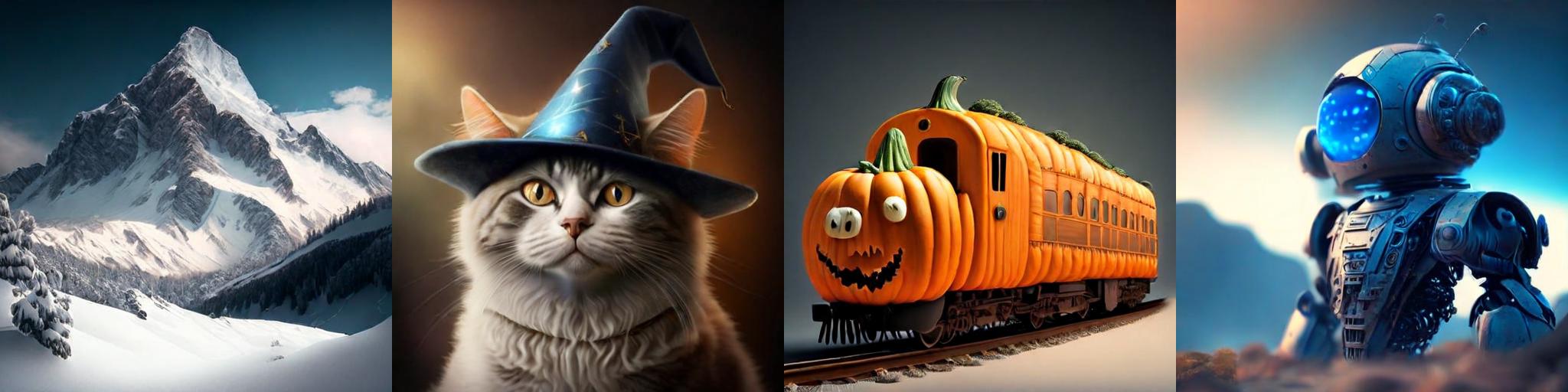}
        \vspace{-18pt}
        \caption*{{\tiny Mountain covered in snow; A cat wearing a wizard hat; A train made of realistic pumpkins; A robot on a blue planet. Close up shot.  \hspace{500pt} \par}}
    \end{subfigure}
    
    \begin{subfigure}[b]{0.245\linewidth}
        \centering
        \includegraphics[width=\linewidth]{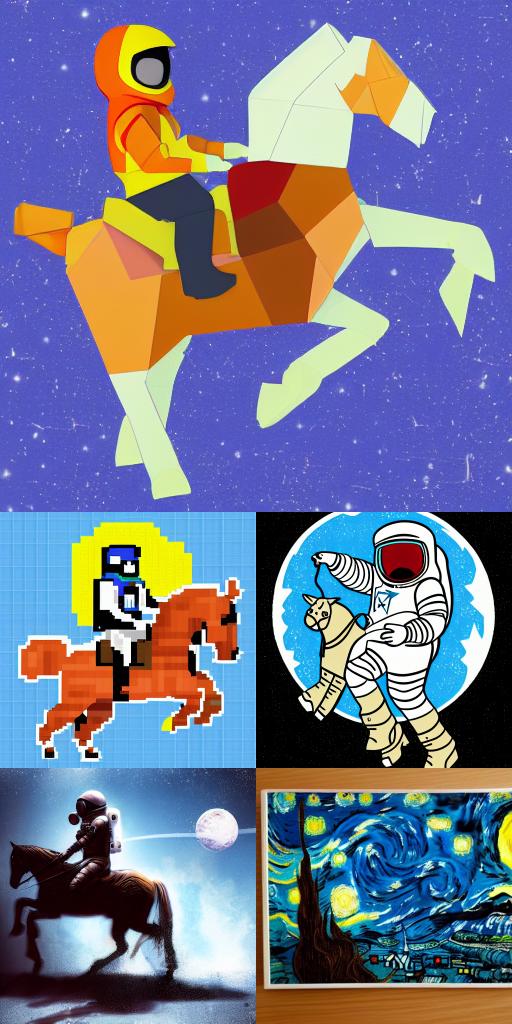}
        \caption{Stable-Diffusion-1.5}
    \end{subfigure}
    \begin{subfigure}[b]{0.245\linewidth}
        \centering
        \includegraphics[width=\linewidth]{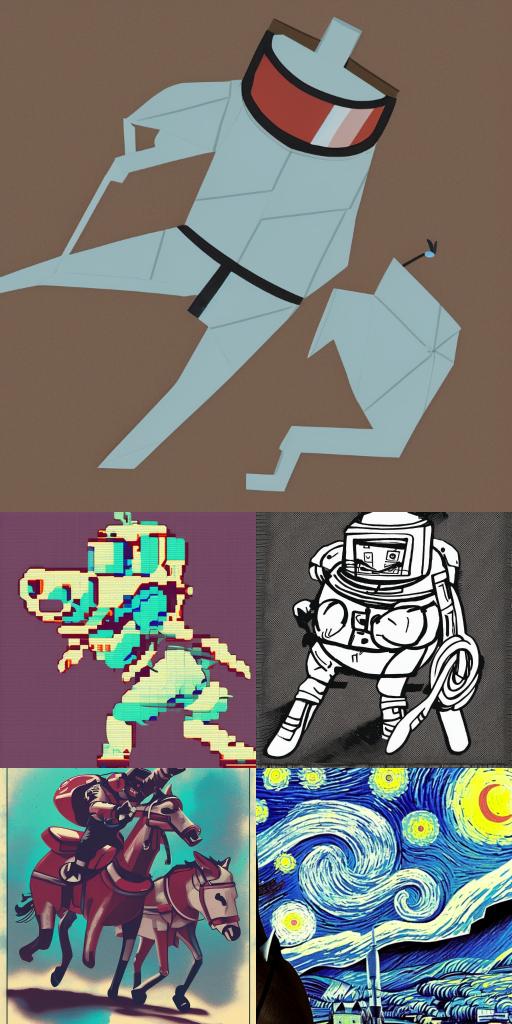}
        \caption{Stable-Diffusion-2.1}
    \end{subfigure}
    \begin{subfigure}[b]{0.245\linewidth}
        \centering
        \includegraphics[width=\linewidth]{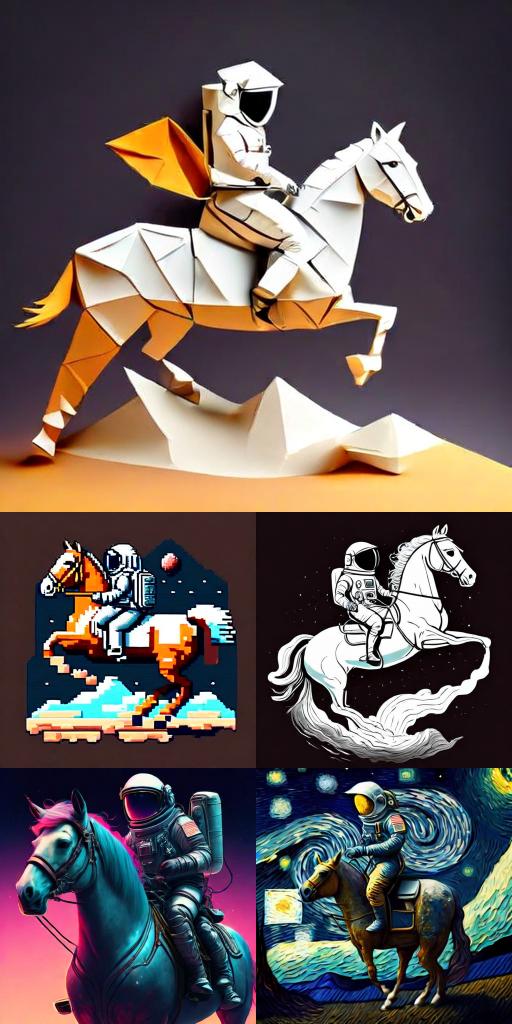}
        \caption{Ours}
    \end{subfigure}
    \hfill
    \begin{subfigure}[b]{0.245\linewidth}
        \centering
        \includegraphics[width=\linewidth]{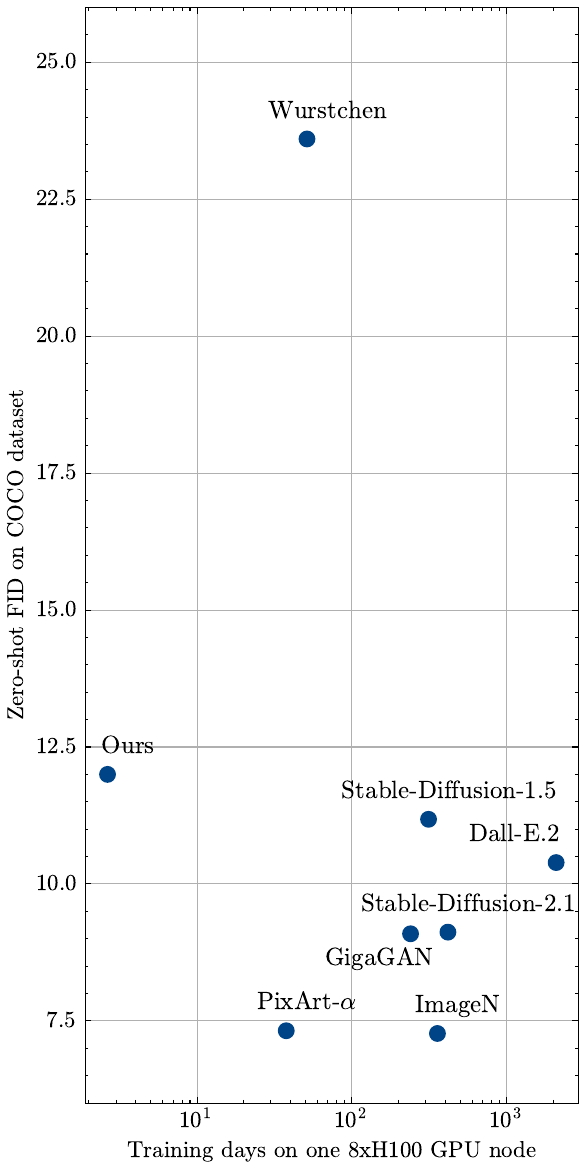}
        \caption{FID vs. training time}
    \end{subfigure}
    \caption{Qualitative evaluation of the image generation capabilities of our model ($512\times512$ image resolution) . Our model is trained in $2.6$ days on a single \eightH machine (amounting to only \$1,890 in GPU cost) without any proprietary or billion image dataset. In (a)-(c) we examine diverse style generation capabilities using prompt \textit{`Image of an astronaut riding a horse in \rule{0.5cm}{0.15mm} style'}, with following styles: Origami, Pixel art, Line art, Cyberpunk, and Van Gogh Starry Night. In (d) we compare training cost and fidelity, measured using FID~\citep{heusel2017FIDgans} for zero-shot image generation on COCO~\citep{lin2014microsoftCOCO} dataset, of all models.}
    
\end{figure}

    

\section{Introduction}
While modern visual generative models excel at creating photorealistic visual content and have propelled the generation of more than a billion images each year~\citep{AIBillionImages2023}, the cost and effort of training these models from scratch remain very high~\citep{rombach2022high, ramesh2022DallE2, saharia2022ImageN}. In text-to-image diffusion models (T2I)~\citep{song2020denoising, ho2020denoising, song2020score}, some previous works have successfully scaled down the computational cost compared to the 200,000 A100 GPU hours used by Stable Diffusion 2.1~\citep{rombach2022high, junsongchenPixArtAlphaFast2023, pabloperniasWuerstchenEfficientArchitecture2023, gokaslan2024commoncanvas}. However, the computational cost of even the state-of-the-art approaches (18,000 A100 GPU hours) remains very high~\citep{junsongchenPixArtAlphaFast2023}, requiring more than a \textit{month} of training time on the leading \eightH GPU machine. Furthermore, previous works either leverage larger-scale datasets spanning on the order of a billion images or use proprietary datasets to enhance performance~\citep{rombach2022high, junsongchenPixArtAlphaFast2023, yu2022Parti, saharia2022ImageN}. The high training cost and the dataset requirements create an inaccessible barrier to participating in the development of large-scale diffusion models. 
In this work, we aim to address this issue by developing a low-cost end-to-end pipeline for competitive text-to-image diffusion models that achieve more than an order of magnitude reduction in training cost than the state-of-the-art while not requiring access to billions of training images or proprietary datasets\footnote{Our code will be available at \url{https://github.com/SonyResearch/micro_diffusion}}.

We consider vision transformer based latent diffusion models for text-to-image generation~\citep{dosovitskiy2020ViT, peebles2023DiT}, particularly because of their simplified design and wide adoption across the latest large-scale diffusion models~\citep{betker2023DallE3, esser2024sd3, junsongchenPixArtAlphaFast2023}. To reduce the computational cost, we exploit the strong dependence of the transformer's computational cost on the input sequence size, i.e., the number of patches per image. We aim to reduce the effective number of patches processed per image by the transformer during training. This objective can be easily achieved by randomly masking a fraction of the tokens at the input layer of the transformer. However, existing masking approaches fail to scale beyond a 50\% masking ratio without drastically degrading performance~\citep{zheng2024maskDit}, particularly because, at high masking ratios, a large fraction of input patches are completely unobserved by the diffusion transformer.

To mitigate the massive performance degradation with masking, we propose a deferred masking strategy where all patches are preprocessed by a lightweight patch-mixer before being transferred to the diffusion transformer. The patch mixer comprises a fraction of the number of parameters in the diffusion transformer. In contrast to naive masking, masking after patch mixing allows non-masked patches to retain semantic information about the whole image and enables reliable training of diffusion transformers at very high masking ratios, while incurring no additional computational cost compared to existing state-of-the-art masking~\citep{zheng2024maskDit}. We also demonstrate that under an identical computational budget, our deferred masking strategy achieves better performance than model downscaling, i.e., reducing the size of the model. Finally, we incorporate recent advancements in transformer architecture, such as layer-wise scaling~\citep{mehta2024openelm} and sparse transformers using mixture-of-experts~\citep{shazeer2017outrageouslySparse, zoph2022stMoe, zhou2022edMoe}, to improve performance in large-scale training.

Our low-cost training pipeline naturally reduces the overhead of ablating experimental design choices at scale. One such ablation we run is examining the impact of the choice of training datasets on the performance of micro-budget training. In addition to using only real images, we consider combining additional synthetic images in the training dataset. We find that training on a combined dataset significantly improves image quality and alignment with human visual preferences.
Furthermore, our combined dataset comprises only 37M images, significantly fewer than most existing large-scale models, thus providing strong evidence of high-quality generations from moderate-sized datasets in micro-budget training.

In summary, we make the following key contributions.
\begin{itemize}[leftmargin=10pt,itemindent=10pt,topsep=0pt,itemsep=0pt]
    \item We propose a novel deferred patch masking strategy that preprocesses all patches before masking by appending a lightweight patch-mixer model to the diffusion transformer.
    \item We conduct extensive ablations of deferred masking and demonstrate that it enables reliable training at high masking ratios and achieves significantly better performance compared to alternatives such as other masking strategies and model downscaling. 
    \item We demonstrate that rather than limiting to only real image datasets, combining additional synthetic images in micro-budget training significantly improves the quality of generated images.
    \item Using a micro-budget of only \$1,890, we train a 1.16 billion parameter sparse diffusion transformer on 37M images and a 75\% masking ratio that achieves a 12.7 FID in zero-shot generation on the COCO dataset. The wall-clock time of our training is only 2.6 days on a single $8\times$H100 GPU machine, 14$\times$ lower than the current state-of-the-art approach that would take 37.6 training days (\$28,400 GPU cost).
\end{itemize}
\section{Methodology}
In this section, we first provide a background on diffusion models, followed by a review of common approaches, particularly those based on patch masking, to reduce computational costs in training diffusion transformers. Then, we provide a detailed description of our proposed deferred patch masking strategy.

\subsection{Background: Diffusion-based generative models and diffusion transformers}
We denote the data distribution by $\mathcal{D}$, such that $(\bx, \bc) \sim \mathcal{D}$, where $\bx$ and $\bc$ represent the image and the corresponding natural language caption, respectively. We assume that $p(\bx; \sigma)$ is the image distribution obtained after adding $\sigma^2$-variance Gaussian noise. 

Diffusion-based probabilistic models use a forward-reverse process-based approach, where the forward process corrupts the input data and the reverse process tries to reconstruct the original signal by learning to reconstruct the data degradation at each step. Though multiple variations of diffusion models exists~\citep{ho2020denoising, song2020scoreSde, bansal2024coldDiff}, it is common to use time-dependent Gaussian noise ($\sigma(t)$) for model data corruption. Under deterministic sampling, both the forward and reverse processes in diffusion models can be modeled using an ordinary differential equation (ODE). 
\begin{align}\label{eq: score-ode}
    \text{d}\bx = -\dot{\sigma}(t)\sigma(t) \nabla_\bx \log p(\bx; \sigma(t)) \text{d}t,
\end{align}
where $\dot{\sigma}(t)$ represents the time derivative and $\nabla_\bx \log p(\bx; \sigma(t))$ is the score of the underlying density function. Thus the move towards (or away from) higher density regions in the diffusion process is proportional to both the scores of the density function and changes in the noise distribution ($\sigma(t)$) with time. Going beyond the deterministic sampling based on the ODE formulation in Equation~\ref{eq: score-ode}, the sampling process in diffusion models can also be formulated as a stochastic differential equation,
\begin{align}\label{eq: score-sde}
    \text{d}\bx = \underbrace{-\dot{\sigma}(t)\sigma(t) \nabla_\bx \log p(\bx; \sigma(t)) \text{d}t}_{\text{Probability flow ODE (Eq.~\ref{eq: score-ode})}} -  \underbrace{\beta(t)\sigma(t)^2  \nabla_\bx \log p(\bx; \sigma(t)) \text{d}t}_{\text{Deterministic noise decay}} + \underbrace{\sqrt{2\beta(t)} \sigma(t)}_{\text{Noise injection}}  \text{d}w_t
\end{align}
where $\text{d}w_t$ is standard the Wiener process and the parameter $\beta(t)$ controls the rate of injection of additional noise. The generation process in diffusion models starts by sampling an instance from $p(\bx, \sigma_{\text{max}})$ and iteratively denoising it using the ODE or SDE-based formulation.

\textbf{Training.} Diffusion model training is based on parameterizing the score of the density function, which doesn't depend on the generally intractable normalization factor, using a denoiser, i.e., $\nabla_\bx \log p(\bx; \sigma(t)) \approx \left(F_\theta(\bx, \sigma(t)) - \bx \right) / \sigma(t)^2$. The denoising function $F_\theta(\bx, \sigma(t))$ is generally modeled using a deep neural network. For text-to-image generation, the denoiser is conditioned on both noise distribution and natural language captions and trained using the following loss function,
\begin{align}\label{eq: loss}
     \mathcal{L}= \mathbb{E}_{(\bx, \bc) \sim \mathcal{D}} \mathbb{E}_{\beps \sim \mathcal{N}(\pmb{0}, \sigma(t)^2\mathbf{I})} \| F_{\theta}(\bx + \beps; \sigma(t), \bc) - \bx \|_2^2.
\end{align}

The noise distribution during training is lognormal ($\text{ln}(\sigma) \sim \mathcal{N}(P_{\text{mean}}, P_{\text{std}})$), where the choice of the mean and standard deviation of the noise distribution strongly influences the quality of generated samples. For example, the noise distribution is shifted to the right to sample larger noise values on higher resolution images, as the signal-to-noise ratio increases with resolution~\citep{teng2023relaydiffusion}. 

\textbf{Classifier-free guidance.} To increase alignment between input captions and generated images, classifier-free guidance~\citep{ho2022FreeGuidance} modifies the image denoiser to output a linear combination of denoised samples in the presence and absence of input captions.
\begin{align}
    \hat{F}_{\theta}(\bx; \sigma(t), \bc) = F_{\theta}(\bx; \sigma(t)) + w \cdot (F_{\theta}(\bx; \sigma(t), \bc) - F_{\theta}(\bx; \sigma(t))),
\end{align}
where $w \geq 1$ controls the strength of guidance. During training, a fraction of image captions (commonly set to 10\%) are randomly dropped to learn unconditional generation. 

\textbf{Latent diffusion models.} In contrast to modeling higher dimensional pixel space ($\mathbb{R}^{h \times w \times 3}$), latent diffusion models~\citep{rombach2022high} are trained in a much lower dimensional compressed latent space ($\mathbb{R}^{\frac{h}{n} \times \frac{w}{n} \times c}$), where $n$ is the compression factor and $c$ is the number of channels in the latent space. The image-to-latent encoding and its reverse mapping are performed by a variational autoencoder. Latent diffusion models achieve faster convergence than training in pixel space and have been heavily adopted in large-scale diffusion models~\citep{dustinpodellSDXLImprovingLatent2023, betker2023DallE3, junsongchenPixArtAlphaFast2023, balajiyogeshEDiffITexttoImageDiffusion2022}.

\textbf{Diffusion transformers.} We consider that the transformer model ($F_{\theta}$) consists of $k$ hierarchically stacked transformer blocks. Each block consists of a multi-head self-attention layer, a multi-head cross-attention layer, and a feed-forward layer. For simplicity, we assume that all images are converted to a sequence of patches to be compatible with the transformer architecture. In the transformer architecture, we refer to the width of all linear layers as $d$. 

\subsection{Common approaches towards minimizing the computational cost of training diffusion transformers}
\begin{figure}[!htb]
    \centering
    \includegraphics[width=\linewidth]{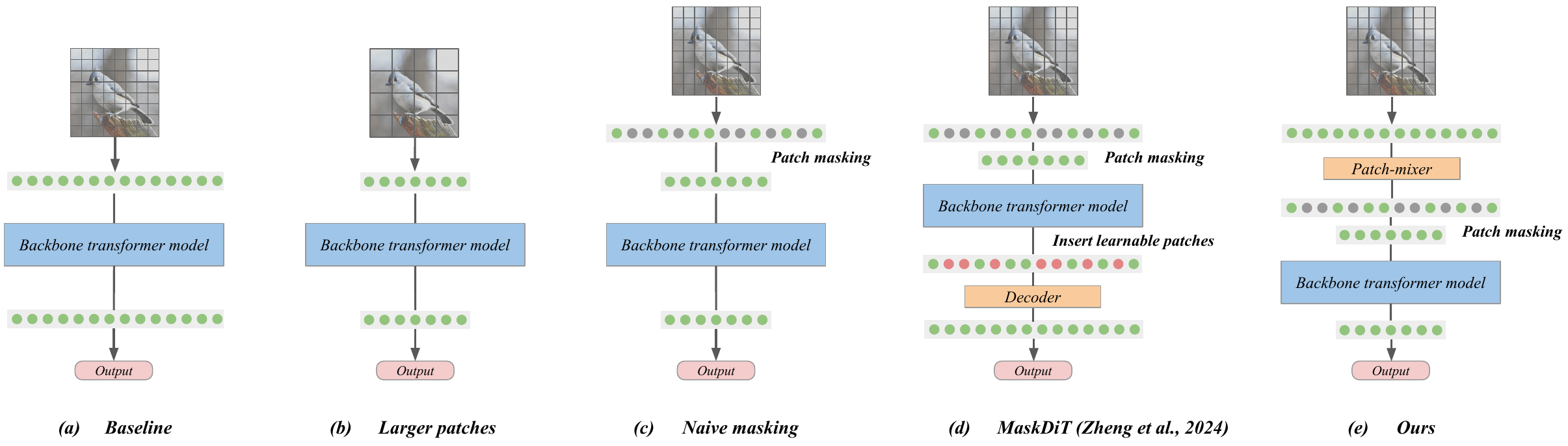}
    \caption{\textbf{Compressing patch sequence to reduce computational cost.} As the training cost of diffusion transformers is proportional to sequence size, i.e., number of patches, it is desirable to reduce the sequence size without degrading performance. It can be achieved by a) using larger patches, b) naively masking a fraction of patches at random, or c) using MaskDiT~\citep{zheng2024maskDit} that combines naive masking with an additional autoencoding objective. We find all three approaches lead to significant degradation in image generation performance, especially at high masking ratios. To alleviate this issue, we propose a straightforward deferred masking strategy, where we mask patches after they are processed by a patch-mixer. Our approach is analogous to naive masking in all aspects except the use of the patch-mixer. In comparison to MaskDiT, our approach doesn't require optimizing any surrogate objectives and has nearly identical computational costs.
    }
    \label{fig: masking_arch_compare}
\end{figure}

Assuming transformer architectures where linear layers account for the majority of the computational cost, the total training cost is proportional to $M \times N \times S$, where $M$ is the number of samples processed, $N$ is the number of model parameters, and $S$ is the number of patches derived from one image, i.e., sequence size for transformer input (Figure~\ref{fig: masking_arch_compare}a). As our goal is to train an open-world model, we believe that a large number of parameters are required to support diverse concepts during generation; thus, we opt to not decrease the parameter count significantly. Since diffusion models converge slowly and are often trained for a large number of steps, even for small-scale datasets, we keep this choice intact~\citep{rombach2022high, balajiyogeshEDiffITexttoImageDiffusion2022}. Instead, we aim to exploit any redundancy in the input sequence to reduce computational cost while simultaneously not drastically degrade performance.

\textit{A. Using larger patch size (Figure~\ref{fig: masking_arch_compare}b).} In a vision transformer, the input image is first transformed into a sequence of non-overlapping patches with resolution $p \times p$, where $p$ is the patch size~\citep{dosovitskiy2020ViT}. Though using a higher patch size quadratically reduces the number of patches per image, it can significantly degrade performance due to the aggressive compression of larger regions of the image into a single patch. 

\textit{B. Using patch masking (Figure~\ref{fig: masking_arch_compare}c).} A contemporary approach is to keep the patch size intact but drop a large fraction of patches at the input layer of the transformer~\citep{he2022mae}. Naive token masking is similar to training on random crops in convolutional UNets, where often upsampling diffusion models are trained on smaller random crops rather than the full image~\citep{nichol2021glide, saharia2022ImageN}. However, in contrast to random crops of the images, patch masking allows training on non-continuous regions of the image\footnote{We observe consistent degradation in the quality of image generation with block masking, i.e., retaining continuous regions of the image (Appendix~\ref{app: ablation}). Thus, we argue that random patch masking in transformers offers a more powerful masking paradigm compared to random image crops in a convolutional network under identical computational cost.}. Due to its effectiveness, masking-based training of transformers has been adopted across both vision and language domains~\citep{devlin2018bert, he2022mae}. 

\textit{C. Masking patches with autoencoding (MaskDiT - Figure~\ref{fig: masking_arch_compare}d).} In order to also encourage representation learning from masked patches, \citet{zheng2024maskDit} adds an auxiliary autoencoding loss that encourages the reconstruction of masked patches. This formulation was motivated by the success of the autoencoding loss in learning self-supervised representations~\citep{he2022mae}. MaskDiT first appends a lightweight decoder, a small-scale transformer network, to the larger backbone transformer. Before the decoder, it expands the backbone transformer output by inserting spurious learnable patches in place of masked patches. 

Assuming that the binary mask $m$ represents the masked patches, the final training loss is the following,
\begin{align}
    &\mathcal{L}_{\text{diff}} = \mathbb{E}_{(\bx, \bc)  \sim \mathcal{D}} \mathbb{E}_{\beps \sim \mathcal{N}(\pmb{0}, \sigma(t)^2\mathbf{I})} \| \left(\bar{F}_{\theta}((\bx + \beps) \odot (1 - m); \sigma, \bc) - \bx \right) \odot (1 - m) \|_2^2 \\
    &\mathcal{L}_{\text{mae}} = \mathbb{E}_{(\bx, \bc) \sim \mathcal{D}} \mathbb{E}_{\beps \sim \mathcal{N}(\pmb{0}, \sigma(t)^2\mathbf{I})} \| \left(\bar{F}_{\theta}((\bx + \beps) \odot (1 - m); \sigma, \bc) - (\bx + \beps) \right) \odot m \|_2^2 \\
    & \mathcal{L} = \mathcal{L}_{\text{diff}} + \gamma \mathcal{L}_{\text{mae}}
\end{align}
where $\bar{F}_{\theta}$ represents the sequential backbone transformer and decoder module, and $\gamma$ is a hyperparameter to balance the influence of the masked autoencoding loss with respect to the diffusion training loss.

\subsection{Alleviating performance bottleneck in patch masking using \ourmasking masking}
To drastically reduce the computational cost, patch masking requires dropping a large fraction of input patches before they are fed into the backbone transformer, thus making the information from masked patches unavailable to the transformer. High masking ratios, e.g., 75\% masking, significantly degrade the overall performance of the transformer. Even with MaskDiT, we only observe a marginal improvement over naive masking, as this approach also drops the majority of image patches at the input layer itself. 

\textbf{Deferred masking to retain semantic information of all patches.} As high masking ratios remove the majority of valuable learning signals from images, we are motivated to ask whether it is necessary to mask at the input layer. As long as the computational cost is unchanged, it is merely a design choice and not a fundamental constraint. In fact, we uncover a significantly better masking strategy, with nearly identical cost to the existing MaskDiT approach. As the patches are derived from non-overlapping image regions in diffusion transformers~\citep{peebles2023DiT, dosovitskiy2020ViT}, each patch embedding does not embed any information about other patches in the image. Thus, we aim to preprocess the patch embeddings before masking, such that non-masked patches can embed information about the whole image. We refer to the preprocessing module as \textit{patch-mixer}.

\textbf{Training diffusion transformers with a patch-mixer.} We consider a patch mixer to be any neural architecture that can fuse the embeddings of individual patches. In transformer models, this objective is naturally achieved with a combination of attention and feedforward layers. So we use a lightweight transformer comprising only a few layers as our patch mixer. We mask the input sequence tokens after they have been processed by the patch mixer (Figure~\ref{fig: masking_arch_compare}e). Assuming a binary mask $m$ for masking, we train our model using the following loss function,
\begin{align}
    &\mathcal{L} = \mathbb{E}_{(\bx, \bc) \sim \mathcal{D}} \mathbb{E}_{\beps \sim \mathcal{N}(\pmb{0}, \sigma(t)^2\mathbf{I})} \| F_{\theta}(M_\phi(\bx + \beps; \sigma(t), \bc) \odot (1 - m); \sigma(t), \bc) - \bx  \odot (1 - m) \|_2^2
\end{align}
where $M_\phi$ is the patch-mixer model and $F_\theta$ is the backbone transformer. Note that compared to MaskDiT, our approach also simplifies the overall design by not requiring an additional loss function and corresponding hyperparameter tuning between two losses during training. We don't mask any patches during inference.

\textbf{Unmasked finetuning.} 
As a very high masking ratio can significantly reduce the ability of diffusion models to learn global structure in the images and introduce a train-test distribution shift in sequence size, we consider a small degree of unmasked finetuning after the masked pretraining. The finetuning can also mitigate any undesirable generation artifacts due to the use of patch masking. Thus, it has been crucial in previous work~\citep{zheng2024maskDit} to recover the sharp performance drop from masking, especially when classifier-free guidance is used in sampling. However, we don't find it completely necessary, as even in masked pretraining, our approach achieves comparable performance to the baseline unmasked pretraining. We only employ it in large-scale training to mitigate any unknown-unknown generation artifacts from a high degree of patch masking.

\begin{figure}[!htb]
  \begin{minipage}[c]{0.45\textwidth}
    \includegraphics[width=\linewidth]{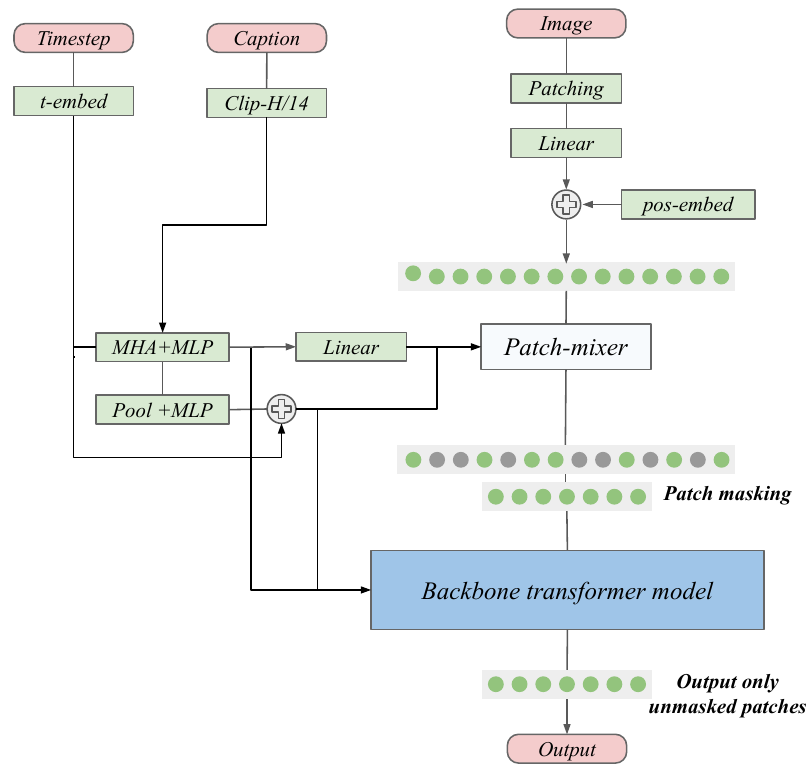}
  \end{minipage}
  \hspace{10pt}
  \begin{minipage}[c]{0.45\textwidth}
    \caption{\textbf{Overall architecture of our diffusion transformer.} We prepend the backbone transformer model with a lightweight patch-mixer that operates on all patches in the input image before they are masked. Following contemporary works~\citep{betker2023DallE3, esser2024sd3}, we process the caption embeddings using an attention layer before using them for conditioning. We use sinusoidal embeddings for timesteps. Our model only denoises unmasked patches, thus the diffusion loss (Eq.~\ref{eq: 
 loss}) is calculated only on these patches. We modify the backbone transformer using layer-wise scaling on individual layers and use mixture-of-expert layers in alternate transformer blocks.}
    \label{subfig: patch_mixer_full_arch}
  \end{minipage}
\end{figure}

\textbf{Improving backbone transformer architecture with mixture-of-experts (MoE) and layer-wise scaling.} We also leverage innovations in transformer architecture design to boost the performance of our model under computational constraints. We use mixture-of-experts layers as they increase the parameters and expressive power of our model without significantly increasing the training cost~\citep{shazeer2017outrageouslySparse, fedus2022switchTransformer}. We use a simplified MoE layer based on expert-choice routing~\citep{zhou2022edMoe}, where each expert determines the tokens routed to it, as it doesn't require any additional auxiliary loss functions to balance the load across experts~\citep{shazeer2017outrageouslySparse, zoph2022stMoe}. We also consider a layer-wise scaling approach, which has recently been shown to outperform canonical transformers in large-language models~\citep{mehta2024openelm}. It linearly increases the width, i.e., hidden layer dimension in attention and feedforward layers, of transformer blocks. Thus, deeper layers in the network are assigned more parameters than earlier layers. We argue that since deeper layers in visual models tend to learn more complex features~\citep{zeiler2014visualizingfeatures}, using higher parameters in deeper layers would lead to better performance. We provide background details on both techniques in Appendix~\ref{app: background}. We describe the overall architecture of our diffusion transformer in Figure~\ref{subfig: patch_mixer_full_arch}.

\section{Experimental Setup}
We provide key details in our experimental setup below and additional details in Appendix~\ref{app: exp_setup}.

We use two variants of Diffusion Transformer (DiT) architecture from \citet{peebles2023DiT}: \dittiny and \ditxl, both comprising a patch size of $2$. When training sparse models, we replace every alternate transformer block with an 8-expert mixture-of-experts block. We train all models using the AdamW optimizer with cosine learning rate decay and high weight decay. We provide an exhaustive list of our training hyperparameters in Table~\ref{tab: big_hparams_table} in Appendix~\ref{app: exp_setup}. We use a four-channel variational autoencoder (VAE) from the Stable-Diffusion-XL~\citep{dustinpodellSDXLImprovingLatent2023} model to extract image latents. We also consider the latest 16-channel VAEs for test their performance in large-scale micro-budget training~\citep{esser2024sd3}. We use the EDM framework from \citet{karras2022elucidating} as a unified training setup for all diffusion models. We refer to our diffusion trasnformer as \ourdit. We use Fréchet inception distance (FID)~\citep{heusel2017FIDgans}, both with the original Inception-v3 model~\citep{szegedy2016rethinking} and a CLIP model~\citep{radford2021Clip}, and CLIP score~\citep{hessel2021clipscore} to measure the performance of the image generation models. 

\textbf{Choice of text encoder.} To convert natural language captions to higher-dimensional feature embeddings, we use the text encoder from state-of-the-art CLIP models\footnote{We use the DFN-5B text encoder: \url{https://huggingface.co/apple/DFN5B-CLIP-ViT-H-14-378}}~\citep{ilharco2021OpenClip, fang2023DFN}. We favor CLIP models over T5-xxl~\citep{raffel2020T5xxl, saharia2022ImageN} for text encoding, despite the better performance of the latter on challenging tasks like text synthesis, as the latter incurs much higher computational and storage costs at scale, thus not suitable for micro-budget training (Table~\ref{tab: t5_vs_clip_cost} in Appendix~\ref{app: exp_setup}). We consider the computation of text embeddings for captions and latent compression of images as a one-time cost that amortizes over multiple training runs of a diffusion model. Thus, we compute them offline and do not account for these costs in our estimation of training costs. We use a \$30/hour cost estimate for an \eightH GPU node\footnote{\url{https://cloud-gpus.com/}} (equivalent to \$12/hour for an \eightA GPU node) to measure the economical cost. 

\textbf{Training datasets.} We use the following three real image datasets comprising $22M$ image-text pairs: Conceptual Captions (CC12M)~\citep{changpinyo2021conceptualcaptions12M}, Segment Anything (SA1B)~\citep{kirillov2023segmentanthing1B}, and TextCaps~\citep{sidorov2020textcaps}. As SA1B does not provide real captions, we use synthetic captions generated from an LLaVA model~\citep{liu2023Llava15, junsongchenPixArtAlphaFast2023}. We also add two synthetic image datasets comprising $15M$ image-text pairs in large-scale training: JourneyDB~\citep{sun2024journeydb} and DiffusionDB~\citep{wang2022diffusiondb}. For small-scale ablations, we construct a text-to-image dataset, named \cifarcap, by subsampling images of ten CIFAR-10 classes from the larger COYO-700M~\citep{kakaobrain2022coyo700m} dataset. Thus, \cifarcap serves as a drop-in replacement for any image-text dataset and simultaneously enables fast convergence due to being closed-domain. We provide additional details on construction of this dataset in Appendix~\ref{app: exp_setup} and samples images in Figure~\ref{fig: cifar_cap_examples}.

\section{Evaluating Effectiveness of Deferred Masking}
In this section, we validate the effectiveness of our deferred masking approach and investigate the impact of individual design choices on image generation performance. We use the \dittiny model and the \cifarcap dataset ($256\times256$ image resolution) for all experiments in this section. We train each model for $60$K optimization steps, with the AdamW optimizer and exponential moving average with a $0.995$ smoothing coefficient enabled for the last $10$K steps.

\subsection{Out-of-the-box performance: Making high masking ratios feasible with deferred masking}
A key limitation of patch masking strategies is poor performance when a large fraction of patches are masked. For example, \citet{zheng2024maskDit} observed large degradation in even MaskDiT performance beyond a 50\% masking ratio. Before optimizing the performance of our approach with a rigorous ablation study, we evaluate its out-of-the-box performance with common training parameters for up to $87.5\%$ masking ratios. As a baseline, we train a network with no patch-mixer, i.e., naive masking (Figure~\ref{fig: masking_arch_compare}c) for each masking ratio. For deferred masking, we use a four-transformer block patch-mixer that comprises less than $10$\% of the parameters of the backbone transformer.

\begin{figure}[!htb]
    \centering
    \includegraphics[width=0.75\linewidth]{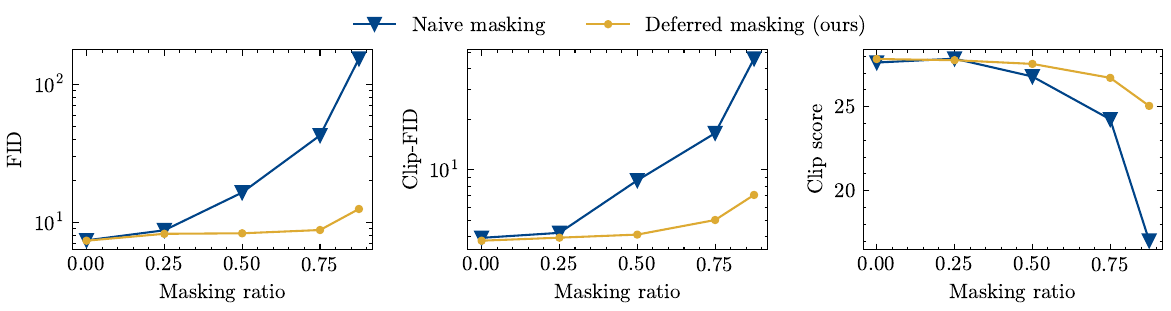}
    \caption{\textbf{Out-of-the-box performance of deferred masking.} Without any hyperparameter optimization, we compare the performance of our deferred masking with a naive masking strategy. We find that deferred masking, i.e., using a patch-mixer before naive masking, tremendously improves image generation performance, particularly at high masking ratios.}
    \label{fig: mixer_success}
\end{figure}

We use commonly used default hyperparameters to simulate the out-of-the-box performance for both our approach and the baseline. We train both models with the AdamW optimizer with an identical learning rate of $1.6 \times 10^{-4}$, $0.01$ weight decay, and a cosine learning rate schedule. We set $(P_{\text{mean}}, P_{\text{std}})$ to $(-1.2, 1.2)$ following the original work~\citep{karras2022elucidating}. We provide our results in Figure~\ref{fig: mixer_success}. Our deferred masking approach achieves much better performance across all three performance metrics. Furthermore, the performance gaps widen with an increase in masking ratio. For example, with a 75\% masking ratio, naive masking degrades the FID score to $16.5$ while our approach achieves $5.03$, much closer to the FID score of $3.79$ with no masking.

\subsection{Ablation study of our training pipeline}
We ablate design choices across masking, patch mixer, and training hyperparameters. We summarize the techniques that improved out-of-the-box performance in Table~\ref{tab: ablation_summary}. We provide supplementary results and discussion of the ablation study in Appendix~\ref{app: ablation}. 

\textbf{Comparison with training hyperparameters of LLMs.} Since the diffusion architectures are very similar to transformers used in large language models (LLMs), we compare the hyperparameter choices across the two tasks. Similar to common LLM training setups~\citep{touvron2023llama, jiang2023mistral7B, chowdhery2022palm}, we find that SwiGLU activation~\citep{shazeer2020Swiglu} in the feedforward layer outperforms the GELU~\citep{hendrycks2016GeLU} activation function. Similarly, higher weight decay leads to better image generation performance. However, we observe better performance when using a higher running average coefficient for the AdamW second moment ($\beta_2$), in contrast to large-scale LLMs where $\beta_2 \approx 0.95$ is preferred. As we use a small number of training steps, we find that increasing the learning rate to the maximum possible value until training instabilities also significantly improves image generation performance. 

\textbf{Design choices in masking and patch-mixer.} We observe a consistent improvement in performance with a larger patch mixer. However, despite the better performance of larger patch-mixers, we choose to use a small patch mixer to lower the computational budget spent by the patch mixer in processing the unmasked input. We also update the noise distribution $(P_{\text{mean}}, P_{\text{std}})$ to $(-0.6, 1.2)$ as it improves the alignment between captions and generated images. By default, we mask each patch at random. We ablate the masking strategy to retain more continuous regions of patches using block masking (Figure~\ref{fig: block_masking}). However, we find that increasing the block size leads to a degradation in the performance of our approach, thus we retain the original strategy of masking each patch at random.

\begin{figure}
    \centering
    \begin{subfigure}[b]{0.48\linewidth}
        \renewcommand*{\arraystretch}{1.15}
        \resizebox{\linewidth}{!}{
        \begin{tabular}{cccc} 
            \toprule
             & \fid ($\downarrow$) & \clipfid ($\downarrow$) & \clipscore ($\uparrow$) \\ \midrule
           Out-of-box          &  $8.82$ & $5.03$ & $26.72$ \\
           Higher weight decay & $8.38$ & $4.90$ & $27.00$ \\ 
           Sigma distribution & $8.49$ & $4.93$ & $27.47$ \\
           Larger patch-mixer & $7.40$ & $4.57$ & $27.79$ \\
           Higher learning rate & $7.09$ & $4.10$ & $28.24$ \\
           \bottomrule
        \end{tabular}}
         \caption{Summarizing the factors that led to improved performance during our ablation study. We conduct this ablation study at $75\%$ masking ratio. Detailed results provided in Table~\ref{tab: ablation_large} in Appendix~\ref{app: ablation}.}
         \label{tab: ablation_summary}
    \end{subfigure}
    \hfill
    \begin{subfigure}[b]{0.5\linewidth}
        \includegraphics[width=\linewidth]{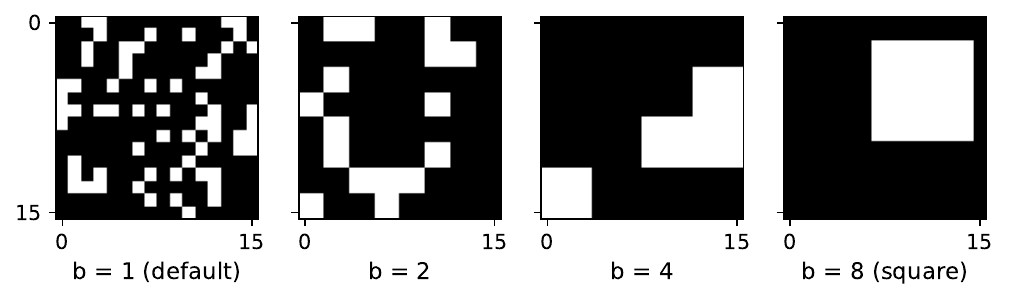}
        \caption{\textbf{Is block masking of patches more effective?} We retain a continuous region of image patches using block masking to test its impact on image generation performance. In this figure, we illustrate the masks for $75\%$ masking ratio with varying block sizes. By default, we mask each patch randomly, i.e., block size ($b$) = 1.}
        \label{fig: block_masking}
    \end{subfigure}
    \vspace{-10pt}
\end{figure}

\subsection{Validating improvements in diffusion transformer architecture}
We measure the effectiveness of the two modifications in transformer architecture in improving image generation performance. 

\begin{minipage}{\textwidth}
\begin{wraptable}{r}{0.4\linewidth}
\vspace{-12pt}
\caption{Layer-wise scaling of transformer architecture is a better fit for masked training in diffusion transformers. We validate its effectiveness in the canonical naive masking with $75\%$ masking ratio.}
\label{tab: layer_wise_scaling}
\vspace{-5pt}
\resizebox{\linewidth}{!}{
\begin{tabular}{cccc}
    \toprule
    Arch & \fid ($\downarrow$) & \clipfid ($\downarrow$) & \clipscore ($\uparrow$) \\
    \midrule
    Constant width & $19.6$ & $9.9$ & $26.7$ \\
    Layer-wise scaling & $\mathbf{15.9}$ & $\mathbf{7.4}$ & $\mathbf{27.1}$ \\ \bottomrule
\end{tabular}
}
\vspace{-5pt}
\end{wraptable} 
\textbf{Layer-wise scaling.} We investigate the impact of this design choice in a standalone experiment where we train two variants of a \texttt{DiT-Tiny} architecture, one with a constant width transformer and the other with layer-wise scaling of each transformer block. We use naive masking for both methods.
We select the width of the constant-width transformer such that its computational footprint is identical to layer-wise scaling. We train both models for an identical number of training steps and wall-clock time. We find that the layer-wise scaling approach outperforms the baseline constant width approach across all three performance metrics (Table~\ref{tab: layer_wise_scaling}), demonstrating a better fit of the layer-wise scaling approach in masked training of diffusion transformers. 
\end{minipage}

\textbf{Mixture-of-experts layers.} We train a \dittiny transformer with 
expert-choice routing based mixture-of-experts (MoE) layers in each alternate transformer block. On the small-scale setup, it achieves similar performance to baseline model trained without MoE blocks. While it slightly improves \clipscore from $28.11$ to $28.66$, it degrades \fid score from $6.92$ to $6.98$. We hypothesize that the lack of improvement is due to the small number of training steps ($60$K) as using multiple experts effectively reduces the number of samples observed by each router. In top-2 routing for 8-experts, each expert is trained effectively for one fourth number of epochs over the dataset. We observe a significantly improvement in performance with MoE blocks when training on scale for longer training schedules (Section~\ref{sec: ablation_scale}).

\begin{figure}[!htb]
    \centering
    \includegraphics[width=\linewidth]{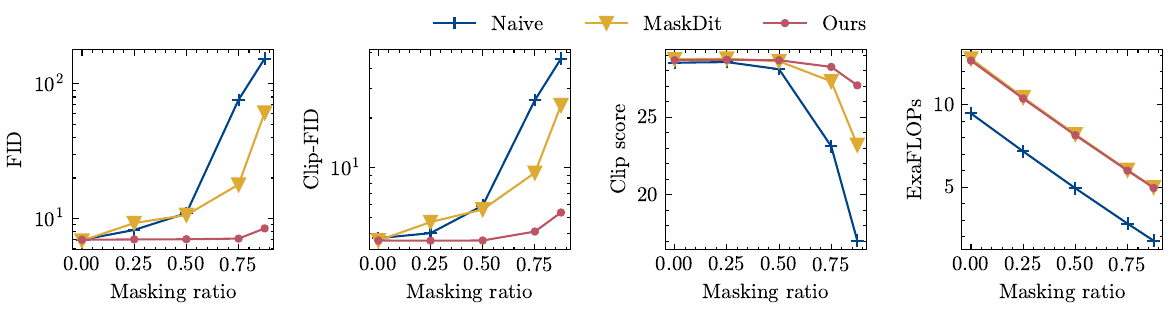}
    \caption{\textbf{Comparing performance of patch masking strategies.} Using a lightweight patch-mixer before patch masking in our deferred masking approach significantly improves image generation performance over baseline masking strategies. Our approach incurs near identical training cost as the MaskDiT~\citep{zheng2024maskDit} baseline. However, both approaches incur slightly higher cost than naive masking due the use of an additional lightweight transformer along with the backbone diffusion transformer.}
\end{figure}

\section{Comparing the Effectiveness of Deferred Masking with Baselines}
In this section, we provide a concrete comparison with multiple baselines. First, we compare with techniques aimed at reducing computational cost by reducing the transformer patch sequence size. Next, we consider reducing network size, i.e., downscaling, while keeping the patch sequence intact. Under isoflops budget, i.e., identical wall-clock training time and \flops, we show that our approach achieves better performance than model downscaling.

\textbf{Comparing different masking strategies.} We first compare our approach with the strategy of using larger patches. We increase the patch size from two to four, equivalent to 75\% patch masking. However, it underperforms compared to deferred masking and only achieves $9.38$, $6.31$, and $26.70$ \fid, \clipfid, and \clipscore, respectively. In contrast, deferred masking achieves $7.09$, $4.10$, and $28.24$ \fid, \clipfid, and \clipscore, respectively. Next, we compare all three masking strategies (Figure~\ref{fig: masking_arch_compare}). We train each model for $60$K steps and report total training flops for each approach to indicate its computational cost. We find that naive masking significantly degrades model performance, and using additional proxy loss functions~\citep{zheng2024maskDit} only marginally improves performance. We find that simply deferring the masking and using a small transformer as a patch-mixer significantly bridges the gap between masked and unmasked training of diffusion transformers.

\begin{figure}[!htb]
    \centering
    \begin{subfigure}[b]{0.48\linewidth}
        \centering
        \resizebox{\linewidth}{!}{
        \begin{tabular}{ccccc} \toprule
            Masking ratio & Masking ratio & \fid & \clipfid & \clipscore \\ \midrule
           Downscaled network  & - & $\mathbf{6.60}$ & $3.85$ & $28.49$ \\ 
           Deferred masking  & $0.5$ & $6.74$ & $\mathbf{3.45}$ & $\mathbf{28.86}$ \\ \midrule
           Downscaled network   & - & $7.09$ & $4.56$ & $27.68$ \\ 
           Deferred masking    & $0.75$ &  $\mathbf{6.96}$ & $\mathbf{4.17}$ & $\mathbf{28.16}$ \\ \midrule
           Downscaled network  & - &  $\mathbf{7.52}$ &  $\mathbf{5.03}$ & $26.98$ \\ 
           Deferred masking   & $0.875$ &  $8.35$ &  $5.26$ & $\mathbf{27.00}$ \\ \bottomrule
        \end{tabular}}
        \caption{\textbf{Iso\flops training.} Under identical training setup and wall-clock time, we compare the effectiveness of model downscaling and our deferred masking approaches. We find that except at extremely high masking ratios, deferred masking achieves better performance across at least two performance metrics. Based on this finding, we do not use a masking ratio higher than 75\% in our models.}
        \label{tab: isoflops}
    \end{subfigure}
    \hfill
    \begin{subfigure}[b]{0.48\linewidth}
        \centering
        \includegraphics[width=\linewidth]{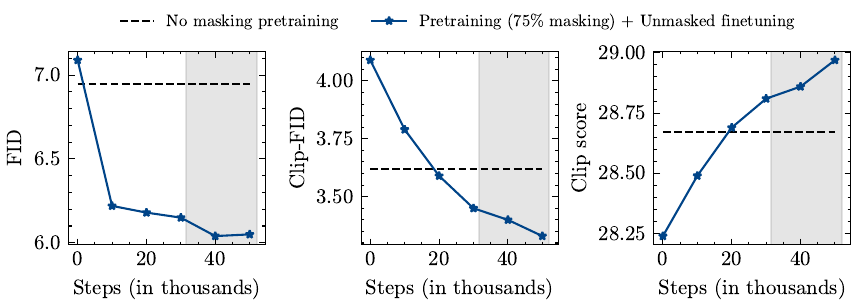}
        \caption{\textbf{Unmasked finetuning}. After pretraining with 75\% deferred masking, we increase the number of unmasked finetuning steps and compare its performance with another model trained completely without masking. The shaded region represents steps after the isoflops threshold, where deferred masking pretraining and finetuning have higher computation cost. At the isoflops threshold, the finetuned model achieves better performance than training the diffusion model without any masking.}
    \end{subfigure}
    \caption{\textbf{Deferred masking vs. model downscaling to reduce training cost.} Both patch masking and model downscaling, i.e., reducing the size of the model, reduce the computational cost in training and can be complementary to each other. However, it is natural to ask how the two paradigms compare to each other in training diffusion transformers. In figure (a), we show that reducing the model size, thus reducing the total computational budget of training, achieves inferior performance compared to our approach at masking ratios under 75\%. Using a patch-mixer in deferred masking effectively enables our model to simultaneously observe all patches in input images while reducing the effective number of patches processed per image. This result encourages using a larger model with masking rather than training a smaller model with no masking for equivalent computational cost. In figure (b), we further highlight the advantage of using deferred masking with finetuning over training a model with no masking for a given computational budget.}
\end{figure}

\textbf{Iso\flops study - Why bother using masking instead of smaller transformers?} Using patch masking is a complementary approach to downscaling transformer size to reduce computation cost in training. We compare the two approaches in Table~\ref{tab: isoflops}. At each masking ratio, we reduce the size of the unmasked network to match \flops and total wall-clock training time of masked training. We train both networks for $60$K training steps. Until a masking ratio of 75\%, we find that deferred masking outperforms network down-scaling across at least two out of three metrics. However, at extremely high masking ratios, deferred masking tends to achieve lower performance. This is likely because the information loss from masking is too high at these ratios. For example, for $32 \times 32$ resolution latent and a patch size of two, only $32$ patches are retained (out of $256$ patches) at an 87.5\% masking ratio.

\textbf{Deferred masking as pretraining $+$ unmasked finetuning.} We show deferred masking also acts as a strong pretraining objective and using it with an unmasked finetuning schedule achieves better performance than training an unmasked network under an identical computational budget. We first train a network without any masking and another identical network with 75\% deferred masking. We finetune the latter with no masking and measure performance as we increase the number of finetuning steps. We mark the \textit{isoflops} threshold when the combined cost of masked pretraining and unmasked finetuning is identical to the model trained with no masking. We find that at the isoflops threshold, the finetuned model achieves superior performance across all three performance metrics. The performance of the model also continues to improve with unmasked finetuning steps beyond the isoflops threshold.

\begin{table}[!htb]
    \centering
    \caption{Computational cost of the two-stage training of our large-scale model. Our total computational cost is $3.45 \times 10^{20}$ \flops, amounting to a total cost of \$1,890 and $2.6$ training days on an \eightH GPU machine.}
    \vspace{-10pt}
    \label{tab: comp_cost_large}
    \resizebox{.8\linewidth}{!}{
    \begin{tabular}{ccccccc} \toprule
        Resolution &  Masking ratio & Training steps & Total \flops & $8\times$A100 days &  $8\times$H100 days & Cost (\$) \\ \midrule
        \multirow{2}{*}{$256\times256$} & $0.75$ & $250000$ & $1.47 \times 10^{20}$ & $2.77$ & $1.11$ & $800$  \\
         & $0.00$ & $30000$ & $4.53 \times 10^{19}$ & $0.94$ & $0.38$  &  $271$ \\ \midrule
         \multirow{2}{*}{$512\times512$} & $0.75$ & $50000$ & $1.18 \times 10^{20}$ & $2.18$ & $0.88$ & $630$ \\
         & $0.00$ & $5000$ & $3.48 \times 10^{19}$ & $0.65$ & $0.26$ & $189$ \\ \bottomrule
    \end{tabular}}
\end{table}

\section{Micro-budget Training of Large-scale Models}
In this section, we validate the effectiveness of our approach in training open-world text-to-image generative models on a micro-budget. Moving beyond the small-scale \cifarcap dataset, we train the large-scale model on a mixture of real and synthetic image datasets comprising $37$M images. We demonstrate the high-quality image generation capability of our model and compare its performance with previous works across multiple quantitative metrics.

\textbf{Training process.} We train a \ditxl transformer with eight experts in alternate transformer blocks. We provide details on training hyperparameters in Table~\ref{tab: big_hparams_table} in the Appendix~\ref{app: exp_setup}. We conduct the training in follwoing two phases. We refer to any diffusion transformer trained using our micro-budget training as \ourdit.
\begin{itemize}[leftmargin=5pt, itemindent=10pt, topsep=0pt]
    \item \textit{Phase-1:} In this phase, we pretrain the model on $256 \times 256$ resolution images. We train for $250$K optimization steps with $75\%$ patch masking and finetune for another $30$K steps without any patch masking. 
    \item \textit{Phase-2:} In this phase, we finetune the Phase-1 model on $512 \times 512$ resolution images. We first finetune the model for $50$K steps with $75\%$ patch masking followed by another $5$K optimization steps with no patch masking.
\end{itemize}

\textbf{Computational cost.} We provide a breakdown of computational cost, including both training \flops and economical cost, across each training phase in Table~\ref{tab: comp_cost_large}. Our Phase-1 and Phase-2 training consumes $56\%$ and $44\%$ of the total computational cost, respectively. The total wall-clock training time of our model is $2.6$ days on an $8 \times$H100 GPU cluster, equivalent to $6.6$ days on an $8 \times$A100 GPU cluster.

\begin{figure}[!htb]
    \begin{subfigure}[c]{0.48\linewidth}
        \includegraphics[width=\linewidth]{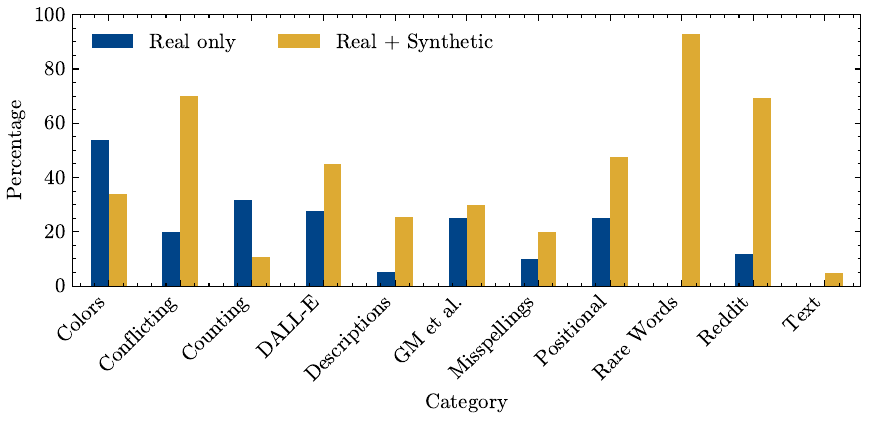}
        \caption{DrawBench}
    \end{subfigure}
    \hfill
    \begin{subfigure}[c]{0.48\linewidth}
        \includegraphics[width=\linewidth]{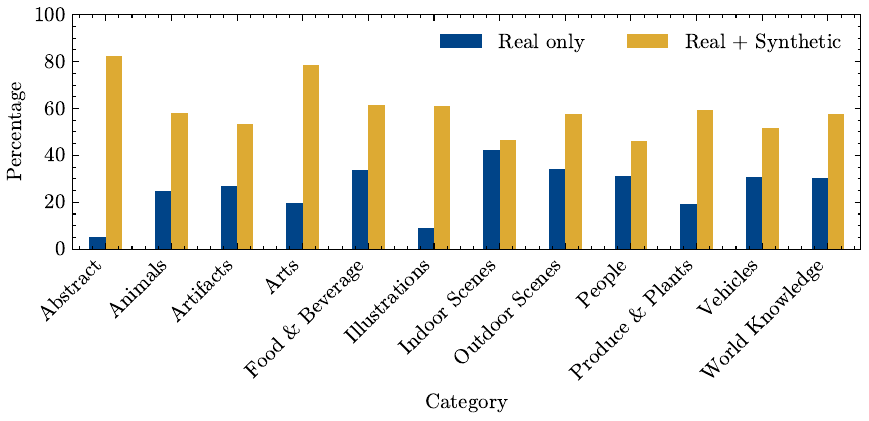}
        \caption{PartiPrompts}
    \end{subfigure}
    \caption{\textbf{Assesssing image quality with GPT-4o.} We supply the following prompt to the GPT-4o model: \textit{Given the prompt `\{prompt\}', which image do you prefer, Image A or Image B, considering factors like image details, quality, realism, and aesthetics? Respond with 'A' or 'B' or 'none' if neither is preferred.} For each comparison, we also shuffle the order of images to remove any ordering bias. We evaluate the performance on two prompt databases: DrawBench~\citep{saharia2022ImageN} and PartiPrompts~\citep{yu2022Parti}. The y-axis in the bar plots indicates the percentage of comparisons in which image from a model is preferred. We breakdown the comparison across individual image category in each prompt database.}
    \label{fig: real_vs_all_gpt4}
    \vspace{10pt}
    \centering
    \includegraphics[width=\linewidth]{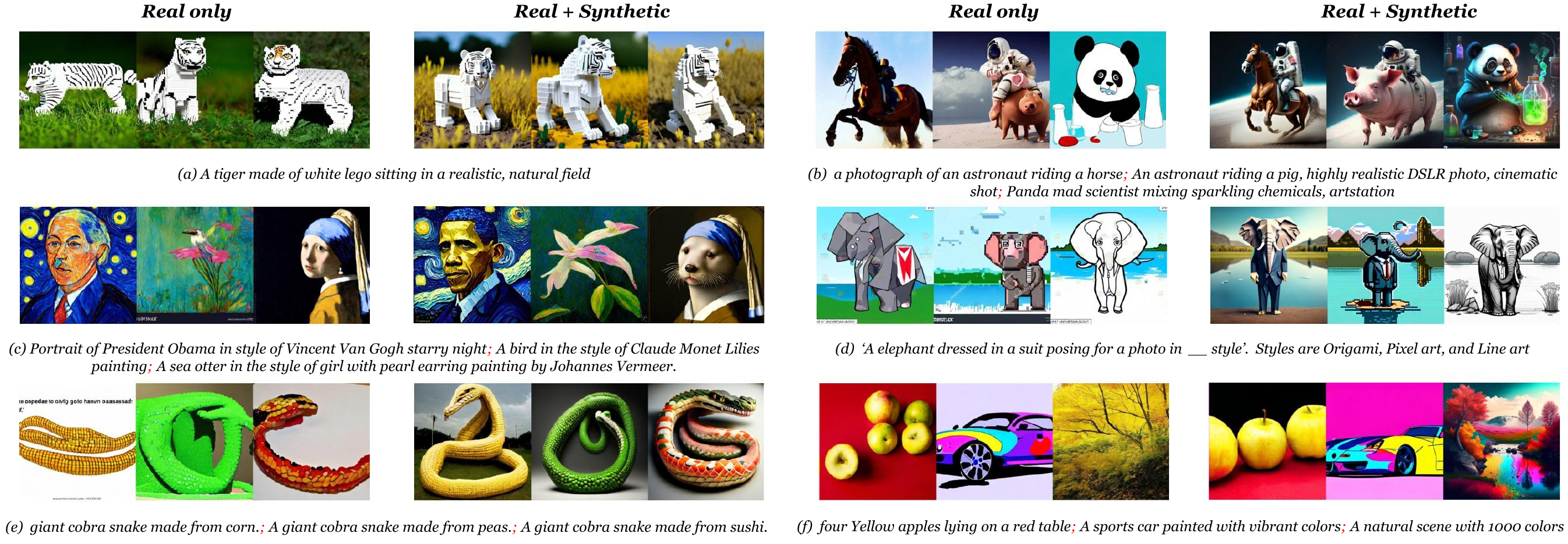}
    \caption{Synthesized images from a model trained only on real data (on left) and combined real and synthetic data (on right). Real data includes SA1B, CC12M, and TextCaps datasets while synthetic data includes JourneyDB and DiffusionDB datasets.}
    \label{fig: real_vs_all}
\end{figure}

\subsection{Benefit of additional synthetic data in training} 
Instead of training only on real images, we expand the training data to include 40\% additional synthetic images. We compare the performance of two \ourdit models trained on only real images and combined real and synthetic images, respectively, using identical training processes.

Under canonical performance metrics, both models apparently achieve similar performance. For example, the model trained on real-only data achieved an FID score of 12.72 and a CLIP score of 26.67, while the model trained on both real and synthetic data achieved an FID score of 12.66 and a CLIP score of 28.14. Even on \geneval~\citep{ghosh2024geneval}, a benchmark that evaluates the ability to generate multiple objects and modelling object dynamics in images, both models achieved an identical score of 0.46. These results seemingly suggest that incorporating a large amount of synthetic samples didn't yield any meaningful improvement in image generation capabilities.

However, we argue that this observation is heavily influenced by the limitations of our existing evaluation metrics. In a qualitative evaluation, we found that the model trained on the combined dataset achieved much better image quality (Figure~\ref{fig: real_vs_all}). The real data model often fails to adhere to the prompt, frequently hallucinating key details and often failing to synthesize the correct object. Metrics, such as FID, fail to capture this difference because they predominantly measure distribution similarity~\citep{pabloperniasWuerstchenEfficientArchitecture2023}. Thus, we focus on using human visual preference as an evaluation metric. To automate the process, we use GPT-4o~\citep{gpt4o2024}, a state-of-the-art multimodal model, as a proxy for human preference. We supply the following prompt to the model: \textit{Given the prompt `\{prompt\}', which image do you prefer, Image A or Image B, considering factors like image details, quality, realism, and aesthetics? Respond with 'A' or 'B' or 'none' if neither is preferred.} For each comparison, we also shuffle the order of images to remove any ordering bias. We generate samples using DrawBench~\citep{saharia2022ImageN} and PartiPrompts (P2)~\citep{yu2022Parti}, two commonly used prompt databases (Figure~\ref{fig: real_vs_all_gpt4}). On the P2 dataset, images from the combined data model are preferred in 63\% of comparisons while images from the real data model are only preferred 21\% of the time (16\% of comparisons resulted in no preference). For the DrawBench dataset, the combined data model is preferred in 40\% of comparisons while the real data model is only preferred in 21\% of comparisons. Overall, using a human preference-centric metric clearly demonstrates the benefit of additional synthetic data in improving overall image quality.

\begin{figure}[!htb]
    \centering
    \begin{subfigure}[c]{0.44\linewidth}
        \caption{Measuring fidelity and prompt alignment of generated images on COCO dataset.}
        \resizebox{\linewidth}{!}{
        \begin{tabular}{cccc}
            \toprule
            Channels  & \fid-30K ($\downarrow$) & \clipfid-30K ($\downarrow$) & \clipscore ($\uparrow$) \\
            \midrule
            $4$  & $\mathbf{12.65}$ & $\mathbf{5.96}$ & $\mathbf{28.14}$ \\
            $16$ & $13.04$ & $6.84$ & $25.63$ \\ \bottomrule
        \end{tabular}
        }
    \end{subfigure}
    \hfill
    \begin{subfigure}[c]{0.53\linewidth}
        \caption{Measuring performance on the \geneval benchmark.}
        \resizebox{\linewidth}{!}{
        \begin{tabular}{cccccccc}
            \toprule
             &  & \multicolumn{2}{c}{Objects} &  &  &  &  \\ \cline{3-4}
            Channels  & Overall & Single & Two & Counting & Colors & Position & \begin{tabular}[c]{@{}c@{}} Color \\ attribution\end{tabular} \\
            \midrule
            4  & $\mathbf{0.46}$ & $\mathbf{0.97}$ & $\mathbf{0.47}$ & $\mathbf{0.33}$ & $\mathbf{0.78}$ & $\mathbf{0.09}$ & $\mathbf{0.20}$ \\
            16  & 0.40 & 0.96 & 0.36 & 0.27 & 0.72 & 0.07 & 0.09 \\
            \bottomrule
            \end{tabular}
        }
    \end{subfigure}
    \caption{We ablate the dimension of latent space by training our \ourdit models in four and sixteen channel latent space, respectively. Even though training in higher dimensional latent space is being adopted across large-scale models~\citep{esser2024sd3, dai2023emu} we find that it underperforms when training on a micro-budget.}
    \label{tab: fid_4_vs_16}
\end{figure}

\begin{table}[!htb]
\resizebox{\linewidth}{!}{
\begin{threeparttable}
    \centering
    \caption{\textbf{Zero-shot FID on COCO2014 validation split}~\citep{lin2014microsoftCOCO}. We report total training time in terms of number of days required to train the model on a machine with eight A100 GPUs. We observe a $2.5\times$ reduction in training time when using H100 GPUs. Our micro-budget training takes $14.2\times$ less training time than state-of-the-art low-cost training approach while simultaneously achieving competitive FID compared to some open-source models.}
    \label{tab: fid_30k}
    \begin{tabular}{c|cccccc}
        \toprule
        Model & Params ($\downarrow$) & \begin{tabular}[c]{@{}c@{}}Sampling \\ steps ($\downarrow$)\end{tabular} & \begin{tabular}[c]{@{}c@{}}Open-source \end{tabular} & \begin{tabular}[c]{@{}c@{}} Training \\images($\downarrow$) \end{tabular} & \begin{tabular}[c]{@{}c@{}}8$\times$A100 \\ GPU days ($\downarrow$) \end{tabular} & \begin{tabular}[c]{@{}c@{}}FID-30K ($\downarrow$) \end{tabular} \\ \midrule
        CogView2~\citep{ding2022cogview2} & $6.00$B & \hyph  & \hyph &  \hyph & \hyph  & $24.0$ \\ 
        Dall-E~\citep{ramesh2021DallE} & $12.0$B & $256$  & \hyph &  \hyph & \hyph  & $17.89$ \\
        Glide~\citep{nichol2021glide} & $3.50$B & $250$  & \hyph &  \hyph & \hyph  & $12.24$ \\
        Parti-750M~\citep{yu2022Parti} & $0.75$B & $1024$  & \hyph &  $3690$M & \hyph  &  $10.71$ \\
        Dall-E.2~\citep{ramesh2022DallE2} & $6.50$B & \hyph  & \hyph & $650$M & $5208.3$  & $10.39$ \\
        Make-a-Scene~\citep{gafni2022makeAscene} & $4.00$B & $1024$ & \hyph &  \hyph & \hyph  & $11.84$ \\
        GigaGAN~\citep{kang2023GigaGAN} & $1.01$B & $1$  & \hyph & $980$M & $597.8$  & $9.09$ \\
        ImageN~\citep{saharia2022ImageN} & $3.00$B & \hyph  & \hyph & $860$M & $891.5$  & $7.27$ \\
        Parti-20B~\citep{yu2022Parti} & $20.0$B & $1024$  & \hyph &  $3690$M & \hyph  &  $7.23$ \\ 
        eDiff-I~\citep{balajiyogeshEDiffITexttoImageDiffusion2022} & $9.10$B & $25$ & \hyph &  $11470$M & \hyph  &  $6.95$ \\ 
        \midrule
        Stable-Diffusion-2.1\tnote{a}~\citep{rombach2022high}  & $0.86$B & $50$  & $\checkmark$ & $3900$M &  $1041.6$ & $9.12$ \\
        Stable-Diffusion-1.5~\citep{rombach2022high} & $0.86$B & $50$  & $\checkmark$ & $4800$M & $781.2$ & $11.18$ \\
        Würstchen~\citep{pabloperniasWuerstchenEfficientArchitecture2023} & $0.99$B & $60$  & $\checkmark$ & $1420$M & $128.1$ & $23.60$ \\
        PixArt-$\alpha$~\citep{junsongchenPixArtAlphaFast2023} & $0.61$B & $20$  & $\checkmark$ &  $25$M\tnote{b} & $94.1$\tnote{c}  &  $7.32$ \\ 
        \textbf{MicroDiT (our work)}  & $1.16$B & $30$  & $\checkmark$ & $37$M & $\mathbf{6.6}$ & $12.66$ \\
        \bottomrule
    \end{tabular}
    \begin{tablenotes}
        \item[a] As the FID scores for the stable diffusion models are not officially reported~\citep{rombach2022high}, we calculate them using the official release of each model. We achieve slightly better FID scores compared to the scores reported in Würstchen~\citep{pabloperniasWuerstchenEfficientArchitecture2023}. We use our FID scores to represent the best performance of these models.
        \item[b] Includes 10M proprietary high-quality images.
         \item[c] PixArt-$\alpha$ training takes 85 days on an \eightA machine when only training till $512\times512$ resolution.

    \end{tablenotes}
\end{threeparttable}
}
\end{table}

\begin{table}[!htb]
    \centering
    \caption{Comparing performance on compositional image generation using \geneval~\citep{ghosh2024geneval} benchmark. Table adopted from~\citet{esser2024sd3}. Higher performance is better.}
    \label{tab: gen_eval_large}
    \resizebox{0.9\linewidth}{!}{
    \begin{tabular}{lcccccccc}
        \toprule
         &   &   &  \multicolumn{2}{c}{Objects} &  &  &  &  \\ \cline{4-5}
        Model & Open-source & Overall & Single & Two & Counting & Colors & Position & \begin{tabular}[c]{@{}c@{}} Color \\ attribution\end{tabular} \\
        \midrule
        DaLL-E.2~\citep{ramesh2022DallE2} & $-$  & 0.52 & 0.94 & 0.66 & 0.49 & 0.77 & 0.10 & 0.19 \\
        DaLL-E.3~\citep{betker2023DallE3} & $-$ & 0.67 & 0.96 & 0.87 & 0.47 & 0.83 & 0.43 & 0.45 \\ \midrule
        minDALL-E~\citep{kakaobrain2021minDALLE} & $\checkmark$ & 0.23 & 0.73 & 0.11 & 0.12 & 0.37 & 0.02 & 0.01 \\
        Stable-Diffusion-1.5~\citep{rombach2022high} & $\checkmark$  & 0.43 & 0.97 & 0.38 & 0.35 & 0.76 & 0.04 & 0.06 \\
        PixArt-$\alpha$~\citep{junsongchenPixArtAlphaFast2023} & $\checkmark$ & 0.48 & 0.98 & 0.50 & 0.44 & 0.80 & 0.08 & 0.07 \\
        Stable-Diffusion-2.1~\citep{rombach2022high} & $\checkmark$ & 0.50 & 0.98 & 0.51 & 0.44 & 0.85 & 0.07 & 0.17 \\
        Stable-Diffusion-XL~\citep{dustinpodellSDXLImprovingLatent2023} & $\checkmark$ & 0.55 & 0.98 & 0.74 & 0.39 & 0.85 & 0.15 & 0.23 \\
        Stable-Diffusion-XL-Turbo~\citep{sauer2023sdxlTurbo} & $\checkmark$ & 0.55 & 1.00 & 0.72 & 0.49 & 0.80 & 0.10 & 0.18 \\
        IF-XL & $\checkmark$ & 0.61 & 0.97 & 0.74 & 0.66 & 0.81 & 0.13 & 0.35 \\
        Stable-Diffusion-3~\citep{esser2024sd3} & $\checkmark$ 
 & 0.68 & 0.98 & 0.84 & 0.66 & 0.74 & 0.40 & 0.43 \\
        \textbf{MicroDiT (our work)} & $\checkmark$ & 0.46 & 0.97 & 0.47 & 0.33 & 0.78 & 0.09 & 0.20 \\
        \bottomrule
        \end{tabular}
        }
\end{table}

\subsection{Ablating design choices on scale} \label{sec: ablation_scale}
We first examine the effect of using a higher dimensional latent space in micro-budget training. We replace the default four-channel autoencoder with one that has sixteen channels, resulting in a $4\times$ higher dimensional latent space. Recent large-scale models have adopted high dimensional latent space as it provides significant improvements in image generation abilities~\citep{dai2023emu, esser2024sd3}. Note that the autoencoder with higher channels itself has superior image reconstruction capabilities, which further contributes to overall success. Intriguingly, we find that using a higher dimensional latent space in micro-budget training hurts performance. For two \ourdit models trained with identical computational budgets and training hyperparameters, we find that the model trained in four-channel latent space achieves better \fid, \clipscore, and \geneval scores (Table~\ref{tab: fid_4_vs_16}). We hypothesize that even though an increase in latent dimensionality allows better modeling of data distribution, it also simultaneously increases the training budget required to train higher quality models. We also ablate the choice of mixture-of-experts (MoE) layers by training another \ourdit model without them under an identical training setup. We find that using MoE layers improves zero-shot FID-30K from 13.7 to 12.7 on the COCO dataset. Due to the better performance of the sparse transformers with MoE layers, we use MoE layers by default in large-scale training.

\subsection{Comparison with previous works}
\textbf{Comparing zero-shot image generation on COCO dataset (Table~\ref{tab: fid_30k}).} We follow the evaluation methodology in previous work~\citep{saharia2022ImageN, yu2022Parti, balajiyogeshEDiffITexttoImageDiffusion2022} and sample 30K random images and corresponding captions from the COCO 2014 validation set. We generate 30K synthetic images using the captions and measure the distribution similarity between real and synthetic samples using \fid (referred to as FID-30K in Table~\ref{tab: fid_30k}). Our model achieves a competitive 12.66 FID-30K, while trained with $14\times$ lower computational cost than the state-of-the-art low-cost training method and without any proprietary or humongous dataset to boost performance. Our approach also outperforms Würstchen~\citep{pabloperniasWuerstchenEfficientArchitecture2023}, a recent low-cost training approach for diffusion models, while requiring $19\times$ lower computational cost.

\textbf{Fine-grained image generation comparison (Table~\ref{tab: gen_eval_large}).} We use the \geneval~\citep{ghosh2024geneval} framework to evaluate the compositional image generation properties, such as object positions, co-occurrence, count, and color. Similar to most other models, our model achieves near perfect accuracy on single-object generation. Across other tasks, our model performs similar to early Stable-Diffusion variants, notably outperforming Stable-Diffusion-1.5. On the color attribution task, our model also outperforms Stable-Diffusion-XL-turbo and PixArt-$\alpha$ models.

\section{Related work}
The landscape of diffusion models~\citep{sohl2015deepEquilibrium, ho2020denoising, song2020denoising, song2020scoreSde} has rapidly evolved in the last few years, with modern models trained on web-scale datasets~\citep{betker2023DallE3, esser2024sd3, rombach2022high, saharia2022ImageN}. In contrast to training in pixel space~\citep{saharia2022ImageN, nichol2021glide}, the majority of large-scale diffusion models are trained in a compressed latent space, thus referred to as latent diffusion models~\citep{rombach2022high}. Similar to auto-regressive models, transformer architectures~\citep{vaswani2017attentionAllyouNeed} have also been recently adopted for diffusion-based image synthesis. While earlier models commonly adopted a fully convolutional or hybrid UNet network architecture~\citep{dhariwal2021diffusionbeatGan, nichol2021glide, rombach2022high, saharia2022ImageN}, recent works have demonstrated that diffusion transformers~\citep{peebles2023DiT} achieve better performance~\citep{esser2024sd3, junsongchenPixArtAlphaFast2023, betker2023DallE3}. Thus, we also use diffusion transformers for modeling latent diffusion models.

Since the image captions in web-scale datasets are often noisy and of poor quality~\citep{kakaobrain2022coyo700m, schuhmann2022laion5B}, recent works have started to recaption them using vision-language models~\citep{liu2023Llava15, wang2023cogvlm, li2024DataCompSyn}. Using synthetic captions leads to significant improvements in the diffusion models' image generation capabilities~\citep{betker2023DallE3, esser2024sd3, gokaslan2024commoncanvas, junsongchenPixArtAlphaFast2023}. While text-to-image generation models are the most common application of diffusion models, they can also support a wide range of other conditioning mechanisms, such as segmentation maps, sketches, or even audio descriptions~\citep{zhang2023Controlnet, yariv2023audio2ImageDiffusion}. Sampling from diffusion models is also an active area of research, with multiple novel solvers for ODE/SDE sampling formulations to reduce the number of iterations required in sampling without degrading performance~\citep{song2020denoising, jolicoeur2021gottagoFast, liu2022pndmSolver, karras2022elucidating}. Furthermore, the latest approaches enable single-step sampling from diffusion models using distillation-based training strategies~\citep{song2023consistencymodels, sauer2023sdxlTurbo}. The sampling process in diffusion models also employs an additional guidance signal to improve prompt alignment, either based on an external classifier~\citep{dhariwal2021diffusionbeatGan, nichol2021glide, sehwag2022Lowdensity} or self-guidance~\citep{ho2022FreeGuidance, karras2024cfgBadVersion}. The latter classifier-free guidance approach is widely adopted in large-scale diffusion models and has been further extended to large-language models~\citep{zhao2024mitigating, sanchez2023stay}.

Since the training cost of early large-scale diffusion models was noticeably high~\citep{rombach2022high, ramesh2022DallE2}, multiple previous works focused on bringing down this cost. \citet{gokaslan2024commoncanvas} showed that using common tricks from efficient deep learning can bring the cost of stable-diffusion-2 models under \$50K. \citet{junsongchenPixArtAlphaFast2023} also reduced this cost by training a diffusion transformer model on a mixture of openly accessible and proprietary image datasets. Cascaded training of diffusion models is also used by some previous works~\citep{saharia2022ImageN, pabloperniasWuerstchenEfficientArchitecture2023, guo2024makecheapscalingselfcascade}, where multiple diffusion models are employed to sequentially upsample the low-resolution generations from the base diffusion model. A key limitation of cascaded training is the strong influence of the low-resolution base model on overall image fidelity and prompt alignment. Most recently, \citet{pabloperniasWuerstchenEfficientArchitecture2023} adopted the cascaded training approach (Würstchen) while training the base model in a $42\times$ compressed latent space. Though Würstchen achieves low training cost due to extreme image compression, it also achieves significantly lower image generation performance on the FID evaluation metric. Alternatively, patch masking has been recently adopted as a means to reduce the computational cost of training diffusion transformers~\citep{zheng2024maskDit, gao2023mdtv2}, taking inspiration from the success of patch masking in contrastive models~\citep{he2022mae}. Patch masking is straightforward to implement in transformers, and the diffusion transformer successfully generalizes to unmasked images in inference, even when patches for each image were masked randomly during training.
\section{Discussion and Future Work}
As the overarching objective of our work is to reduce the overhead of training large-scale diffusion models, we not only aim to reduce the training cost but also align design choices in the training setup with this objective. For example, we deliberately choose datasets that are openly accessible and show that using only these datasets suffices to generate high-quality samples, thus omitting the need for proprietary or enormous datasets. We also favor CLIP-based text embeddings which, although underperforming compared to T5 model embeddings (especially in text generation), are much faster to compute and require six times less disk storage.

While we observe competitive image generation performance with micro-budget training, it inherits the limitations of existing text-to-image generative models, especially in text rendering. It is surprising that the model learns to successfully adhere to complex prompts demonstrating strong generalization to novel concepts while simultaneously failing to render even simple words, despite the presence of a dedicated optical character recognition dataset in training. Similar to a majority of other open-source models, our model struggles with controlling object positions, count, and color attribution, as benchmarked on the \geneval dataset.

We believe that the lower overhead of large-scale training of diffusion generative models can dramatically enhance our understanding of training dynamics and accelerate research progress in alleviating the shortcomings of these models. It can enable a host of explorations on scale, in directions such as understanding end-to-end training dynamics~\citep{agarwal2022estimating, choromanska2015loss, jiang2020neurips}, measuring the impact of dataset choice on performance~\citep{xie2024doremi}, data attribution for generated samples~\citep{zheng2023intriguing, dai2023training, park2023trak, koh2017understanding}, and adversarial interventions in the training pipeline~\citep{chou2023backdoor, shafahi2018poison}. Future work can also focus on alleviating current limitations of micro-budget training, such as poor rendering of text and limitations in capturing intricate relationships between objects in detailed scenes. While we focus mainly on algorithmic efficiency and dataset choices, the training cost can be significantly reduced by further optimizing the software stack, e.g., by adopting native 8-bit precision training~\citep{mellempudi2019mixed8bitTrain}, using dedicated attention kernels for H100 GPUs~\citep{shah2024flashattention3}, and optimizing data loading speed.

\section{Conclusion}
In this work, we focus on patch masking strategies to reduce the computational cost of training diffusion transformers. We propose a deferred masking strategy to alleviate the shortcomings of existing masking approaches and demonstrate that it provides significant improvements in performance across all masking ratios. With a deferred masking ratio of 75\%, we conduct large-scale training on commonly used real image datasets, combined with additional synthetic images. Despite being trained at an order of magnitude lower cost than the current state-of-the-art, our model achieves competitive zero-shot image generation performance. We hope that our low-cost training mechanism will empower more researchers to participate in the training and development of large-scale diffusion models.

\section{Acknowledgements}
We would like to thank Peter Stone, Felancarlo Garcia, Yihao Zhan, and Naohiro Adachi for their valuable feedback and support during the project. We also thank Weiming Zhuang, Chen Chen, Zhizhong Li, Sina Sajadmanesh, Jiabo Huang, Nidham Gazagnadou, and Vivek Sharma for their constructive feedback throughout the progress of this project. 

\bibliography{ref.bib}
\bibliographystyle{tmlr}

\appendix
\section{Additional details on experimental setup} \label{app: exp_setup}
We use \dittiny and \ditxl diffusion transformer architectures for small and large scale training setups, respectively. We use four and six transformer blocks in the patch-mixer for the \dittiny and \ditxl architectures, respectively. The patch-mixer comprises approximately 13\% of the parameters in the backbone diffusion transformer.
We use half-precision layernorm normalization and SwiGLU activation in the feedforward layers for all transformers. Initially, half-precision \layernorm leads to training instabilities after $100$K steps of training. Thus, we further apply layer normalization to query and key embeddings in the attention layers~\citep{dehghani2023ViT22B}, which stabilizes the training. We reduce the learning rate for expert layers by half as each expert now processes a fraction of all patches. We provide an exhaustive list of our training hyperparameters in Table~\ref{tab: big_hparams_table}. We use the default configuration ($\sigma_{\text{max}}=80, \sigma_{\text{min}}=0.002, S_{\text{max}}=\infty, S_{\text{noise}}=1, S_{\text{min}}=0, S_{\text{churn}}=0$), except that we increase $\sigma_{data}$ to match the standard deviation of our image datasets in latent space. We generate images using deterministic sampling from Heun's $2^{nd}$ order method~\citep{fourier1822fourier, karras2022elucidating} with $30$ sampling steps. Unless specified, we use classifier-free guidance of $3$ and sample $30$K images in quantitative evaluation on the \cifarcaps dataset. We reduce the guidance scale to $1.5$ for large-scale models. We find that these guidance values achieve the best \fid score under the \fid-\clipscore tradeoff. For all qualitative generations, we recommend a guidance scale of $5$ to better photorealism and prompt adherence. By default, we use $512 \times 512$ pixel resolution when generating synthetic images from large-scale models and  $256 \times 256$ pixel resolution with small-scale model trained on \cifarcap dataset.

We conduct all experiments on an 8$\times$H100 GPU node. We train and infer all models using \texttt{bfloat16} mixed-precision mode. We did not observe a significant speedup when using FP8 precision with the Transformer Engine library\footnote{\url{https://github.com/NVIDIA/TransformerEngine}}. We use PyTorch 2.3.0~\citep{paszke2019pytorch} with PyTorch native flash-attention-2 implementation. We use the DeepSpeed~\citep{rasley2020deepspeed} flops profiler to estimate total \flops in training. We also use just-in-time compilation (\texttt{torch.compile}) to achieve a 10-15\% speedup in training time. We use StreamingDataset~\citep{mosaicml2022streaming} to enable fast dataloading.

\begin{table}[!htb]
    \centering
    \caption{Hyperparameter across both phases of our large scale training setup.}
    \label{tab: big_hparams_table}
    \resizebox{0.8\linewidth}{!}{
    \begin{tabular}{cccccc} \toprule
       Resolution  & \multicolumn{2}{c}{$256 \times 256$ (Phase-I)} &  & \multicolumn{2}{c}{$512 \times 512$ (Phase-II)} \\ \cmidrule{2-3} \cmidrule{5-6}
       Masking ratio  & $0.75$ & $0$ &  & $0.75$ & $0$ \\ 
       Training steps & $250000$ & $30000$ &  &  $50000$ & $5000$ \\ 
       Batch size & $2048$ & $2048$  &   & $2048$  & $2048$  \\
       Learning rate & $2.4 \times 10^{-4}$ & $8 \times 10^{-5}$ & & $8 \times 10^{-5}$ & $8 \times 10^{-5}$ \\
       Weight decay & $0.1$ & $0.1$ & & $0.1$ & $0.1$ \\
       Optimizer & AdamW & AdamW &  & AdamW & AdamW \\
       Momentum & $\beta_1=0.9, \beta_2=0.999$ & $\beta_1=0.9, \beta_2=0.999$ &  & $\beta_1=0.9, \beta_2=0.999$ & $\beta_1=0.9, \beta_2=0.999$ \\
       Optimizer epsilon & $1 \times 10^{-8}$ & $1 \times 10^{-8}$ &  & $1 \times 10^{-8}$ & $1 \times 10^{-8}$ \\
       Lr scheduler & Cosine & Constant &  & Constant & Constant \\
       Warmup steps & $2500$ & $0$ &  & $500$ & $0$ \\
       Gradient clip norm & $0.25$ & $0.25$ &  & $0.5$ & $0.5$ \\
       EMA & no ema & no ema &  & $0.99975$ & $0.9975$ \\
       Hflip & False & False &  & False & False \\
       Precision & bf16 & bf16 &  & bf16 & bf16 \\
       Layernorm precision & bf16 & bf16 &  & bf16 & bf16 \\
       QK-normalization & True & True &  & True & True \\
       $(P_{\text{mean}}, P_{\text{std}})$ & ($-0.6$, $1.2$) & ($-0.6$, $1.2$)  &  &  ($0$, $0.6$) & ($0$, $0.6$) \\
       \bottomrule
    \end{tabular}}
\end{table}

\noindent \textbf{Choice of text encoder.} To convert discrete token sequences in captions to high-dimensional feature embeddings, two common choices of text encoders are CLIP~\citep{radford2021Clip, ilharco2021OpenClip} and T5~\citep{raffel2020T5xxl}, with T5-xxl\footnote{\url{https://huggingface.co/DeepFloyd/t5-v1_1-xxl}} embeddings narrowly outperforming equivalent CLIP model embeddings~\citep{saharia2022ImageN}. However, T5-xxl embeddings pose both compute and storage challenges: 1) Computing T5-xxl embeddings costs an order of magnitude more time than a CLIP ViT-H/14 model while also requiring $6\times$ more disk space for pre-computed embeddings. Overall, using T5-xxl embeddings is more demanding (Table~\ref{tab: t5_vs_clip_cost}). Following the observation that large text encoders trained on higher quality data tend to perform better~\citep{saharia2022ImageN}, we use state-of-the-art CLIP models as text encoders~\citep{fang2023DFN}\footnote{We use the DFN-5B text encoder: \url{https://huggingface.co/apple/DFN5B-CLIP-ViT-H-14-378}}.

\textbf{Cost analysis.} We translate the wall-clock time of training to financial cost using a \$30/hour budget for an 8$\times$H100 GPU cluster. Since the cost of H100 fluctuates significantly across vendors, we base our estimate on the commonly used cost estimates for A100 GPUs~\citep{junsongchenPixArtAlphaFast2023}, in particular \$1.5/A100/hour\footnote{\url{https://cloud-gpus.com/}}. We observe a 2.5x reduction in wall-clock time on H100 GPUs, thus assume a \$3.75/H100/hour cost. We also report the wall-clock time to benchmark the training cost independent of the fluctuating GPU costs in the AI economy.

\textbf{Datasets.} We use the following five datasets to train our final models:
\begin{itemize}
    \item \textit{Conceptual Captions (real).} This dataset was released by Google and includes 12 million pairs of image URLs and captions~\citep{changpinyo2021conceptualcaptions12M}. Our downloaded version of the dataset includes $10.9$M image-caption pairs. 
    \item \textit{Segment Anything-1B (real).} Segment Anything comprises $11.1$M high-resolution images originally released by Meta for segmentation tasks~\citep{kirillov2023segmentanthing1B}. Since the images do not have corresponding real captions, we use synthetic captions generated by the LLaVA model~\citep{junsongchenPixArtAlphaFast2023, liu2023Llava15}.
    \item \textit{TextCaps (real).} This dataset comprises $28$K images with natural text in the images~\citep{sidorov2020textcaps}. Each image has five associated descriptions. We combine them into single captions using the Phi-3 language model~\citep{abdin2024phi3}.
    \item \textit{JourneyDB (synthetic).} JourneyDB is a synthetic image dataset comprising $4.4$M high-resolution Midjourney image-caption pairs~\citep{sun2024journeydb}. We use the train subset ($4.2$M samples) of this dataset.
    \item \textit{DiffusionDB (synthetic).} DiffusionDB is a collection of $14$M synthetic images generated primarily by Stable Diffusion models~\citep{wang2022diffusiondb}. We filter out poor quality images from this dataset, resulting in $10.7$M samples, and use this dataset only in the first phase of pretraining.
\end{itemize}

\textbf{CIFAR-Captions.} We construct a text-to-image dataset, named \cifarcap, that closely resembles the existing web-curated datasets and serves as a drop-in replacement of existing datasets in our setup. \cifarcap is closed-domain and only includes images of ten classes (airplanes, cars, birds, cats, deer, dogs, frogs, horses, ships, and trucks), imitating the widely used CIFAR-10 classification dataset~\citep{krizhevsky2009cifar10}. In contrast to other small-scale text-to-image datasets that are open-world, such as subsets of conceptual captions~\citep{changpinyo2021conceptualcaptions12M}, we observe fast convergence of diffusion models on this dataset. We create this dataset by first downloading $120$M images from the coyo-$700$M~\citep{kakaobrain2022coyo700m} dataset. We observe a success rate of approximately $60\%$ at the time of downloading the dataset. Next, we use a ViT-H-14~\citep{fang2023DFN} to measure CLIP score by averaging it over the eighteen prompt templates (such as `a photo of a \{\}.') used in the original CLIP model~\citep{radford2021Clip}. We select images with CLIP scores higher than $0.25$ (~1.25\% acceptance rate) that results in a total of $1.3$M images. As the real captions are highly noisy and uninformative, we replace them with synthetic captions generated with an LLaVA-1.5 model~\citep{liu2023Llava15}.

\begin{table}[!htb]
    \centering
    \caption{Comparing the computational cost and storage overhead of CLIP~\citep{radford2021Clip, fang2023DFN} and T5~\citep{raffel2020T5xxl, 2024PileT5} text encoders. We use the state-of-the-art CLIP model from \citet{fang2023DFN}. We report the compute and storage cost for $1M$ image captions. Even though T5 embeddings achieve better generation quality, especially for text generation, computing them is an order of magnitude slower than CLIP embeddings and even precomputing them for our dataset (37M images) poses high storage overheads (36.4 TB). We use one H100 GPU to estimate the time to process $1M$ image captions. We save the embeddings in \texttt{float16} precision.
    }
    \label{tab: t5_vs_clip_cost}
    \begin{tabular}{ccccc} \toprule
        Text encoder & Sequence length & Embedding size & Time (min:sec) & Storage (GB) \\ \midrule
        CLIP (\texttt{ViT-H-14}) & $77$  &  $1024$  &  3:20 &  $157$ \\
        T5 (\texttt{T5-xxl}) & $120$  &  $4096$  &  33:04 & $983$ \\ \bottomrule
    \end{tabular}
\end{table}

\textbf{Evaluation metrics.} We use the following evaluation metrics to assess the quality of synthetic images generated by the text-to-image models.
\begin{itemize}
    \item \textit{\fid}. Fréchet Inception Distance (FID) measures the 2-Wasserstein distance between real and synthetic data distributions in the feature space of a pretrained vision model. We use the clean-fid\footnote{\url{https://github.com/GaParmar/clean-fid}}~\citep{parmar2022cleanfid} implementation for a robust evaluation of the FID score.
    \item \textit{\clipfid}. Unlike \fid that uses an Inception-v3~\citep{szegedy2016rethinking} model trained on ImageNet~\citep{deng2009imagenet}, \clipfid uses a CLIP~\citep{radford2021Clip} model, trained on a much larger dataset than ImageNet, as an image feature extractor. We use the default ViT-B/32 CLIP model from the clean-fid library to measure \clipfid.
    \item \textit{\clipscore}. It measures the similarity between a caption and the generated image corresponding to the caption. In particular, it measures the cosine similarity between normalized caption and image embeddings calculated using a CLIP text and image encoder, respectively. 
\end{itemize}

\begin{figure}
    \centering
    \begin{subfigure}[c]{0.48\linewidth}
        \caption{\textit{Ablating the choice of $\beta_2$ in Adam optimizer}. Unlike the LLM training where $\beta_2$ if often set to $0.95$~\citep{touvron2023llama, brown2020GPT3}, we find that image generation quality consistently degrades as we reduce $\beta_2$.}
        \resizebox{\linewidth}{!}{
        \begin{tabular}{cccc}
            \toprule
            $\beta_2$  & \fid ($\downarrow$) & \clipfid ($\downarrow$) & \clipscore ($\uparrow$) \\
            \midrule
            $0.999$ & $\mathbf{8.53}$ & $\mathbf{4.85}$ & $\mathbf{26.88}$ \\
            $0.99$ & $8.63$ & $4.94$ & $26.75$ \\
            $0.95$ & $8.71$ & $5.02$ & $26.71$ \\
            $0.9$ & $8.81$ & $5.13$ & $26.61$ \\ \bottomrule
        \end{tabular}
        }
    \end{subfigure}
    \hfill
    \begin{subfigure}[c]{0.48\linewidth}
        \caption{\textit{Ablating the choice of weight-decay in \texttt{AdamW} optimizer}. In resonance with transformer training in LLMs, we observe improvement in performance with increase in weight decay regularization.}
        \resizebox{\linewidth}{!}{
        \begin{tabular}{cccc}
            \toprule
            $wd$  & \fid ($\downarrow$) & \clipfid ($\downarrow$) & \clipscore ($\uparrow$) \\
            \midrule
            $0.00$ & $8.77$ & $5.03$ & $26.82$ \\
            $0.01$ & $8.73$ & $5.03$ & $26.82$ \\
            $0.03$ & $8.53$ & $\mathbf{4.85}$ & $26.88$ \\
            $0.10$ & $\mathbf{8.38}$ & $4.90$ & $\mathbf{27.00}$ \\ \bottomrule
        \end{tabular}
        }
    \end{subfigure}
    \begin{subfigure}[c]{0.48\linewidth}
        \caption{\textit{Ablating the parameters of noise distribution}. We used patch-mixer with $0.999$ $\beta_2$ and $0.1$ weight decay. We observe a tradeoff between between \fid and \clipscore in first two choices and set $(m, s)$ to $(-0.6, 1.2)$ in all followup experiments.}
        \resizebox{\linewidth}{!}{
        \begin{tabular}{cccc}
            \toprule
            $(m, s)$  & \fid ($\downarrow$) & \clipfid ($\downarrow$) & \clipscore ($\uparrow$) \\
            \midrule
            $(-1.2, 1.2)$ & $\mathbf{8.38}$ & $\mathbf{4.90}$ & $27.00$ \\
            $(-0.6, 1.2)$ & $8.49$ & $4.93$ & $\mathbf{27.47}$ \\
            $(-0.6, 0.6)$ & $9.05$ & $6.72$ & $26.95$ \\
            $(-0.25, 0.6)$ & $10.44$ & $7.51$ & $27.46$ \\
            $(0.0, 0.6)$ & $12.76$ & $9.00$ & $27.40$ \\ \bottomrule
        \end{tabular}
        }
    \end{subfigure}
    \hfill
    \begin{subfigure}[c]{0.48\linewidth}
        \caption{\textit{Ablating the block-size in block sampling.} We observe consistent performance degradation with block masking. At block size of $8$ (latent image with $16\times16=256$ patches), block sampling sampling collapse to sampling a quadrant, thus we sample a single continuous square patch for it.}
        \resizebox{\linewidth}{!}{
        \begin{tabular}{cccc}
            \toprule
            $Block-size$  & \fid ($\downarrow$) & \clipfid ($\downarrow$) & \clipscore ($\uparrow$) \\
            \midrule
            $1$ & $\mathbf{8.49}$ & $\mathbf{4.93}$ & $\mathbf{27.47}$ \\
            $2$ & $8.89$ & $4.64$ & $27.42$ \\
            $4$ & $9.80$ & $5.09$ & $26.91$ \\
            square & $12.71$ & $7.11$ & $26.14$ \\ \bottomrule
        \end{tabular}
        }
    \end{subfigure}
    \begin{subfigure}[c]{0.48\linewidth}
        \caption{\textit{Ablating the size of patch-mixer}. We increase the width of path embeddings, while also varying the multipliers for attention layers ($m_a$) and feedforward layers ($m_f$). We also report the total wall-clock time (in hours:minutes) for training.}
        \resizebox{\linewidth}{!}{
        \begin{tabular}{ccccc}
            \toprule
            $(w, m_a, m_b)$  & Time & \fid ($\downarrow$) & \clipfid ($\downarrow$) & \clipscore ($\uparrow$) \\
            \midrule
            $(384, 0.5, 0.5)$ & $3$:$19$ & $8.49$ & $4.93$ & $27.47$ \\
            $(384, 1.0, 4.0)$ & $3$:$20$ & $7.72$ & $4.80$ & $27.72$ \\
            $(512, 1.0, 4.0)$ & $3$:$24$ & $7.40$ & $4.57$ & $\mathbf{27.79}$ \\ 
            $(768, 1.0, 4.0)$ & $3$:$39$ & $\mathbf{7.09}$ & $\mathbf{4.39}$ & $27.76$ \\ \bottomrule
        \end{tabular}
        }
    \end{subfigure}
    \hfill
    \begin{subfigure}[c]{0.48\linewidth}
        \caption{\textit{Ablating the choice of feedforward layers: GELU vs SwiGLU activation}. We find that replace GELU activation with a SwiGLU based activation in feedforward layer improves all three performance metrics. SwiGLU activation is also commonly used in transformers for large language models~\citep{touvron2023llama}.}
        \resizebox{\linewidth}{!}{
        \begin{tabular}{cccc}
            \toprule
            $Block-size$  & \fid ($\downarrow$) & \clipfid ($\downarrow$) & \clipscore ($\uparrow$) \\
            \midrule
            \textit{GELU} & $7.82$ & $4.61$ & $27.33$ \\ 
            \textit{SwiGLU} & $\mathbf{7.40}$ & $\mathbf{4.57}$ & $\mathbf{27.79}$ \\ \bottomrule
        \end{tabular}
        }
    \end{subfigure}
    \begin{subfigure}[c]{0.48\linewidth}
        \caption{\textit{Testing the cyclic learning rate scheduler}. We consider the cosine learning rate with warm restarts, while varying the duration of base cycle ($t$) and multiplier ($m_t$) where the duration of each subsequent cycle increase by the multiplier factor.}
        \label{tab: cyclic_lr}
        \resizebox{\linewidth}{!}{
        \begin{tabular}{cccc}
            \toprule
            $lr$ schedule  & \fid ($\downarrow$) & \clipfid ($\downarrow$) & \clipscore ($\uparrow$) \\
            \midrule
            1-cycle ($t=60,000, m_t=1$) & $\mathbf{7.40}$ & $4.57$ & $27.79$ \\
            3-cycles ($t=20,000, m_t=1)$ & $7.64$ & $4.52$ & $27.75$ \\ 
            4-cycles ($t=4,000, m_t=2)$ & $7.49$ & $\mathbf{4.43}$ & $\mathbf{27.81}$ \\ \bottomrule
        \end{tabular}
        }
    \end{subfigure}
    \hfill
    \begin{subfigure}[c]{0.48\linewidth}
        \caption{\textit{Using higher learning rate for each batch for better performance.} We observe training instabilities after ~$2$K steps for learning rates higher than $3.2\times10^{-4}$ in mixed-precision training.}
        \resizebox{\linewidth}{!}{
        \begin{tabular}{cccc}
            \toprule
            $lr$  & \fid ($\downarrow$) & \clipfid ($\downarrow$) & \clipscore ($\uparrow$) \\
            \midrule
            $1.6\times10^{-4}$ & $7.40$ & $4.57$ & $27.79$ \\
            $3.2\times10^{-4}$ & $\mathbf{7.09}$ & $\mathbf{4.10}$ & $\mathbf{28.24}$ \\ \bottomrule
        \end{tabular}
        }
    \end{subfigure}
    
    \caption{Ablating individual components in our training pipeline. In each subsequent ablation we use the overall best performing model from previous ablation. In each ablation, we train a \dittiny model for $60$K training steps using $75$\% masking ratio.}
    \label{tab: ablation_large}
\end{figure}

\section{Background on layer-wise scaling and mixture-of-experts} \label{app: background}
\textbf{Layer-wise scaling.} In contrast to using identical transformer blocks throughout the network, layer-wise scaling dynamically increases the transformer block size with the depth of the network. In a text-to-image transformer, each block consists of a self-attention layer, a cross-attention layer, and a feedforward layer. Layer-wise scaling increases the hidden dimension of feedforward layers by a factor $m_f$, thus linearly increasing the number of parameters and \flops of the feedforward block. Similarly, layer-wise scaling dynamically increases the number of heads in the attention layers by scaling the embedding size by a factor of $m_a$. By default, $m_a=1.0$ and $m_f=4.0$ lead to a canonical transformer network. We use $m_a \in [0.5, 1.0]$ and $m_f \in [0.5, 4.0]$ to reduce the size of transformer blocks. Note that we do not apply any scaling to cross-attention layers ($m_a=1.0$, $m_f=4.0$) to avoid degrading the controllability of captions on image generation. 

\begin{figure}[!htb]
    \centering
    \includegraphics[width=0.5\linewidth]{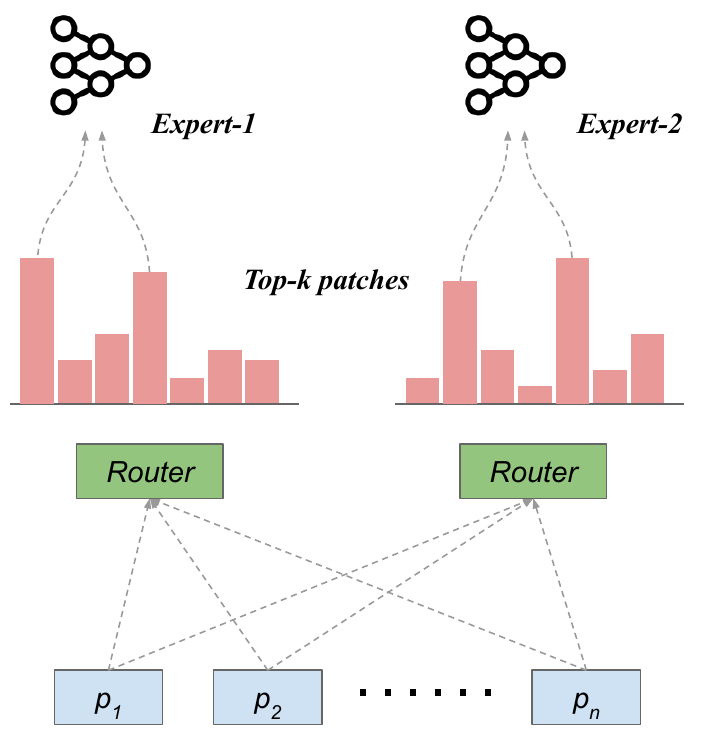}
    \caption{\textbf{MoE.} Expert-choice routing based mixture-of-experts~\citep{zhou2022edMoe}. Each patch is passed to a patch router that determines the top-k patches routed to each expert.}
    \label{fig: moe}
\end{figure}

\textbf{Mixture-of-experts.} Mixture-of-experts enables the construction of much larger models, referred to as sparse models, with minimal impact on the training and inference cost, thus making them highly applicable for micro-budget training. A sparse model modifies the transformer block to include replicas of the feedforward layer, referred to as experts. The input patch embeddings to the feedforward layers are first fed into a router that determines the configuration of patches processed by each expert. We use the expert-choice (EC) routing mechanism~\citep{zhou2022edMoe} where each expert selects the top-k patches using the importance score determined by the routing network for each expert (Figure~\ref{fig: moe}). We favor EC routing over conventional routing due to its simplicity, as it does not require any auxiliary loss to balance the load across experts~\citep{shazeer2017outrageouslySparse, zoph2022stMoe}. We use a linear layer as a router and it is trained jointly with the rest of the network.

\section{Detailed results on deferred masking and patch mixer ablation study} \label{app: ablation}
When ablating the learning rate, we observe better performance with higher learning rates. However, we run into training instabilities after approximately $2$K steps, despite using qk-normalization~\citep{dehghani2023ViT22B}, with mixed precision training. Overall, we recommend ablating the choice of learning rate for each network and using the maximum attainable value without causing training instability. We also consider a fast training strategy using cyclic learning schedules~\citep{smith2015cyclicalLR}. Cyclic learning has been previously observed to improve the convergence rate for image classifiers~\citep{smith2015cyclicalLR,smith2022generalCyclicLR}. We consider a cyclic cosine schedule with base cycle time $t$ and each subsequent cycle duration multiplied by a multiplier ($m_t$). We only observe a marginal benefit of cyclic schedules (Table~\ref{tab: cyclic_lr}). In favor of simplicity, we continue to use a single cycle schedule.

In ablating the choice of masking, we shift from masking patches randomly to retaining a continuous region of image patches with block masking~\ref{fig: block_masking}. We perform this ablation on $256 \times 256$ image resolution with a corresponding latent resolution of $32 \times 32$ and $256$ patches with a patch size of two in the \dittiny model. Note that at a block size of 8 and a $75$\% masking ratio, block sampling collapses to masking quadrants of image patches. Thus, for this configuration, we resort to sampling a single continuous square patch of non-masked patches. Overall, we find that any amount of block masking degrades performance compared to random masking of each patch. It is likely because despite latent compression and patching, there exists some redundancy in visual information between neighboring patches, and the diffusion transformer model receives more global semantic information about the image with random masking.

\begin{figure}
    \centering
    \begin{subfigure}[b]{\linewidth}
        \centering
        \includegraphics[width=\linewidth]{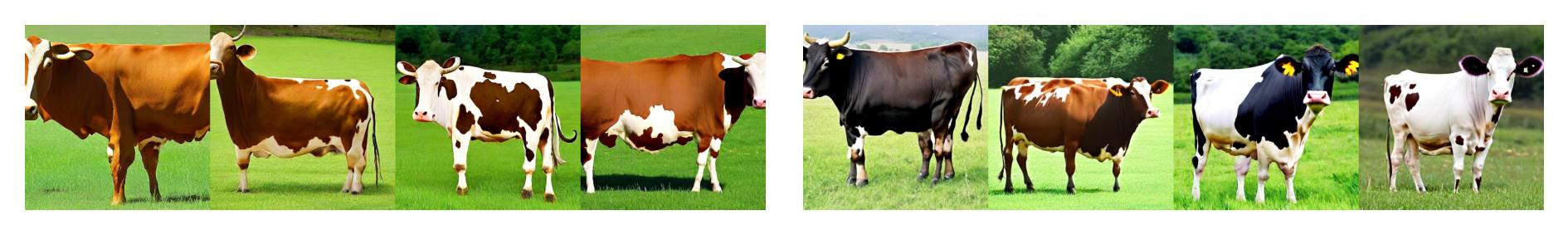}
        \caption{a photo of a cow}
    \end{subfigure}
    \begin{subfigure}[b]{\linewidth}
        \centering
        \includegraphics[width=\linewidth]{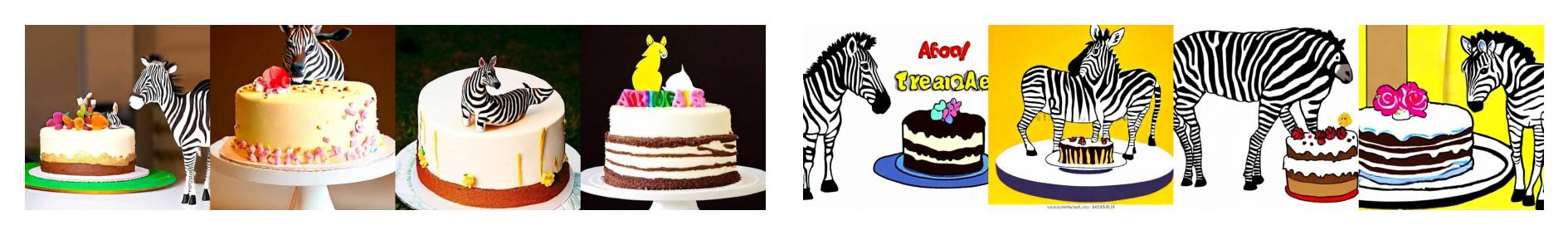}
        \caption{a photo of a cake and a zebra}
    \end{subfigure}
    \begin{subfigure}[b]{\linewidth}
        \centering
        \includegraphics[width=\linewidth]{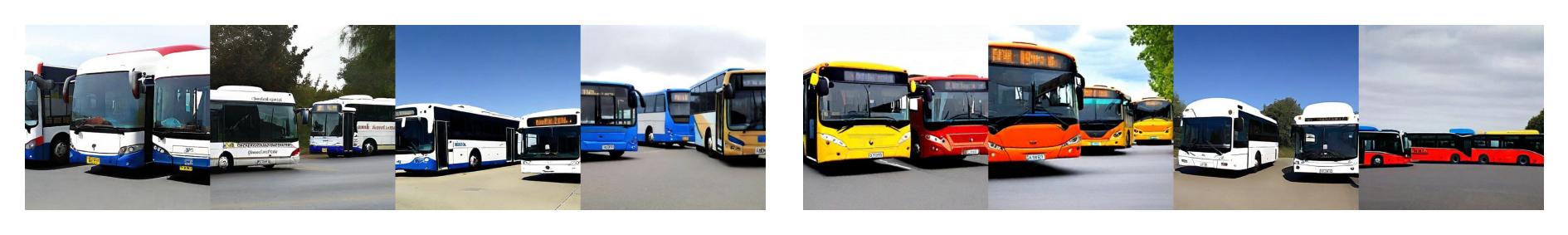}
        \caption{a photo of three buses}
    \end{subfigure}
    \begin{subfigure}[b]{\linewidth}
        \centering
        \includegraphics[width=\linewidth]{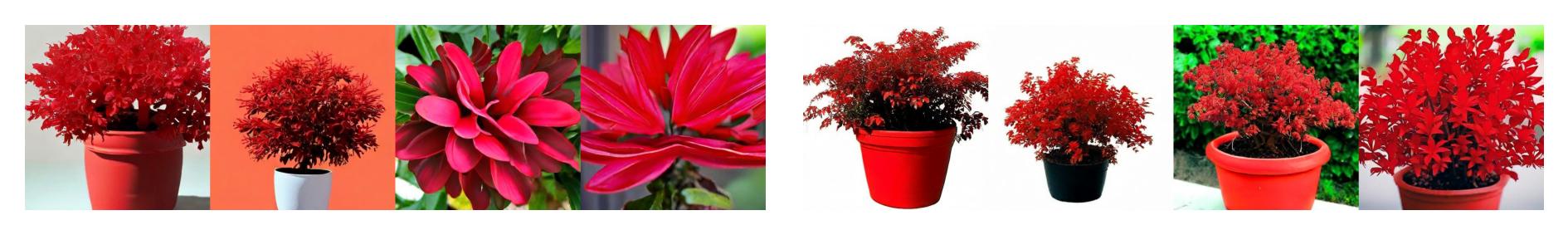}
        \caption{a photo of a red potted plant}
    \end{subfigure}
    \begin{subfigure}[b]{\linewidth}
        \centering
        \includegraphics[width=\linewidth]{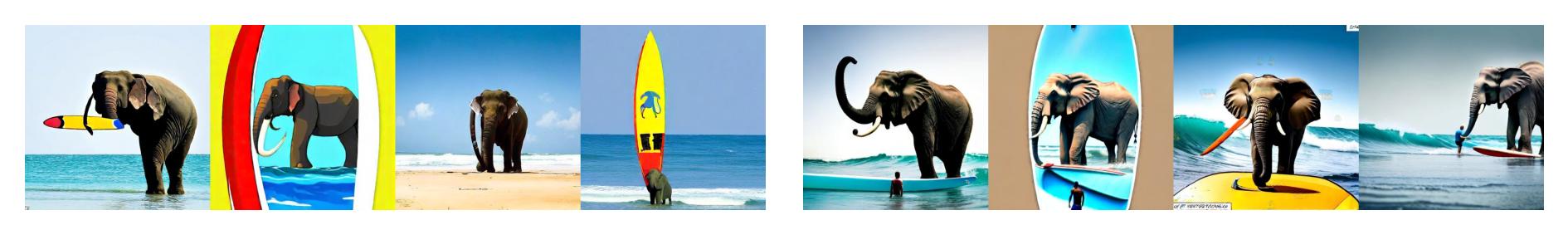}
        \caption{a photo of an elephant below a surfboard}
    \end{subfigure}
    \begin{subfigure}[b]{\linewidth}
        \centering
        \includegraphics[width=\linewidth]{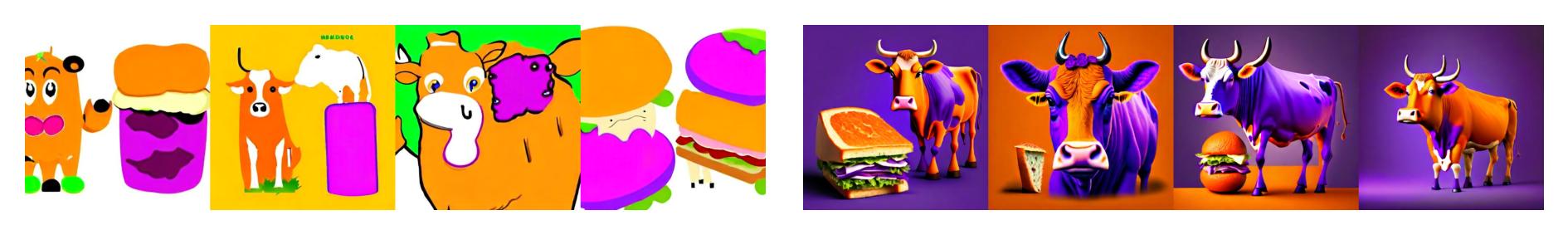}
        \caption{a photo of an orange cow and a purple sandwich}
    \end{subfigure}
    \caption{Generation from a model trained only on real data (on left) and combined real and synthetic data (on right) on GenEval benchmark prompts. Both models use identical random seed for generation.}
\end{figure}

\begin{figure}[!htb]
     \centering
     \includegraphics[width=\linewidth]{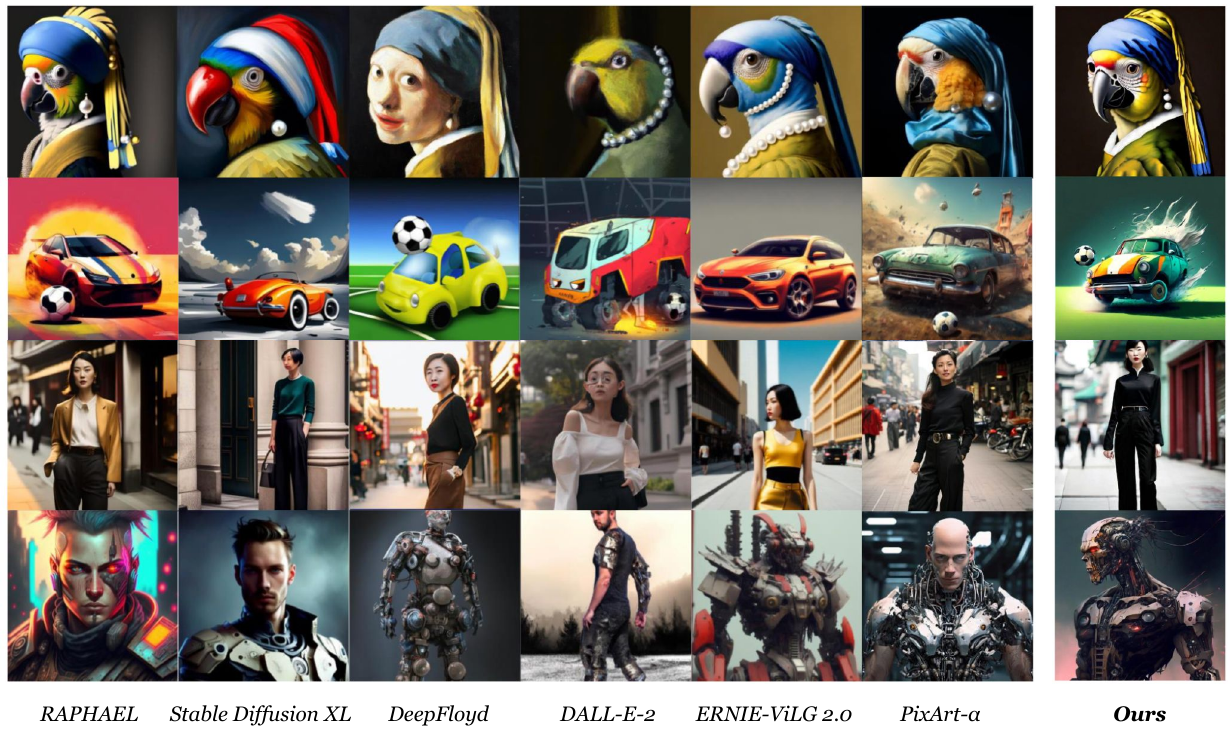}
     \begin{itemize}
        \item A parrot with a pearl earring, Vermeer style.
        \item A car playing soccer, digital art.
        \item Street shot of a fashionable Chinese lady in Shanghai, wearing black high-waisted trousers.
        \item Half human, half robot, repaired human, human flesh warrior, mech display, man in mech, cyberpunk.
    \end{itemize}
    \caption{Comparison with previous works. Figure adapted from~\citet{junsongchenPixArtAlphaFast2023}.}
\end{figure}

\begin{figure}[!htb]
     \centering
     \begin{subfigure}[b]{0.5\linewidth}
         \centering
         \includegraphics[width=\linewidth]{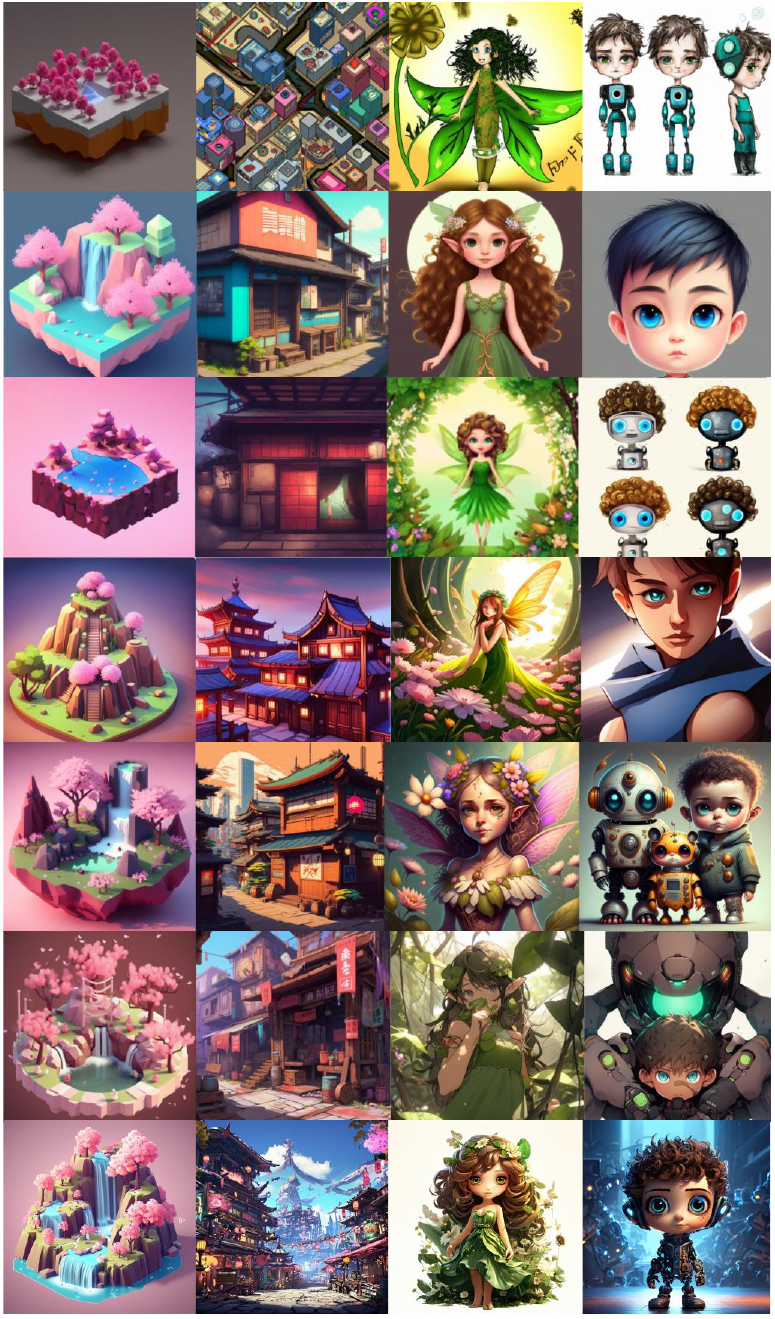}
         \caption{\textit{Previous works}: RAPHAEL, Stable Diffusion XL, DeepFloyd, DALL-E-2, ERNIE-ViLG 2.0, PixArt-$\alpha$}
     \end{subfigure}
     \\
     \begin{subfigure}[b]{0.5\linewidth}
         \centering
         \includegraphics[width=\linewidth]{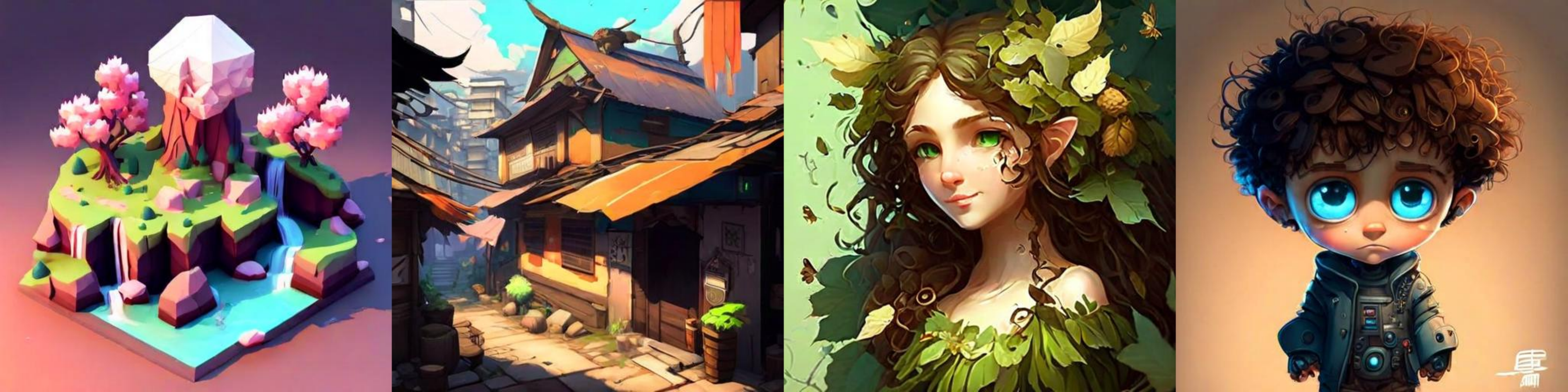}
         \caption{Our work}
     \end{subfigure}
     \begin{itemize}
        \tiny
        \item A cute little matte low poly isometric cherry blossom forest island, waterfalls, lighting, soft shadows, trending on Artstation, 3d render, monument valley, fez video game.
        \item A shanty version of Tokyo, new rustic style, bold colors with all colors palette, video game, genshin, tribe, fantasy, overwatch.
        \item Cartoon characters, mini characters, figures, illustrations, flower fairy, green dress, brown hair,curly long hair, elf-like wings,many flowers and leaves, natural scenery, golden eyes, detailed light and shadow , a high degree of detail. 
        \item Cartoon characters, mini characters, hand-made, illustrations, robotkids, color expressions, boy, short brown hair, curly hair,blue eyes, technological age, cyberpunk, big eyes, cute, mini, detailed light and shadow, high detail.
    \end{itemize}
    \caption{Comparison with previous works. Figure adapted from~\citet{junsongchenPixArtAlphaFast2023}.}
\end{figure}

\begin{figure}
    \centering
    \includegraphics[width=0.8\linewidth]{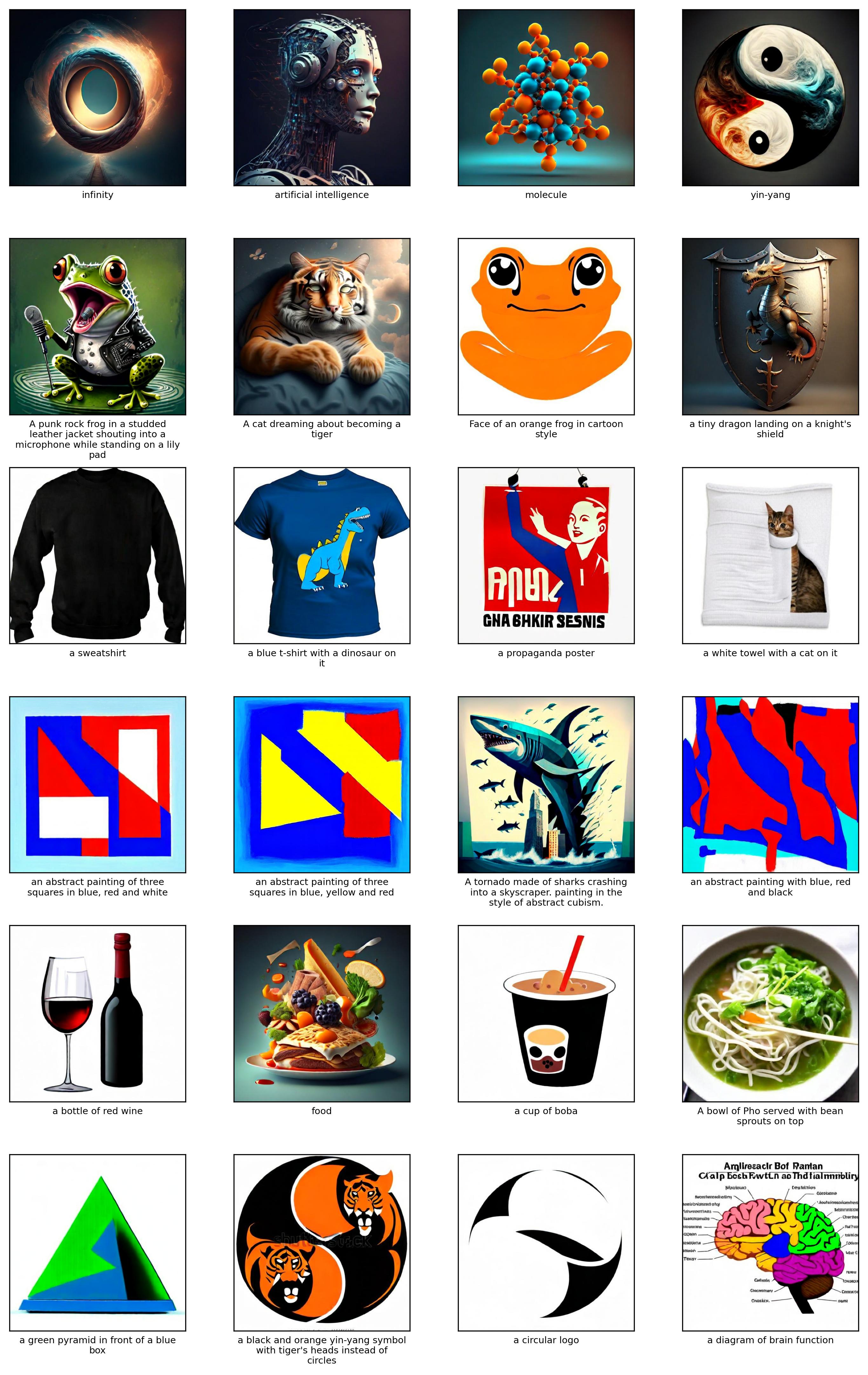}
    \caption{Synthesized images by our model using randomly selected prompts from PartiPrompts~\citep{yu2022Parti}. Rows correspond to following categories: Abstract, Animals, Artifacts, Arts, Food \& Beverage, and Illustrations.}
\end{figure}

\begin{figure}
    \centering
    \includegraphics[width=0.8\linewidth]{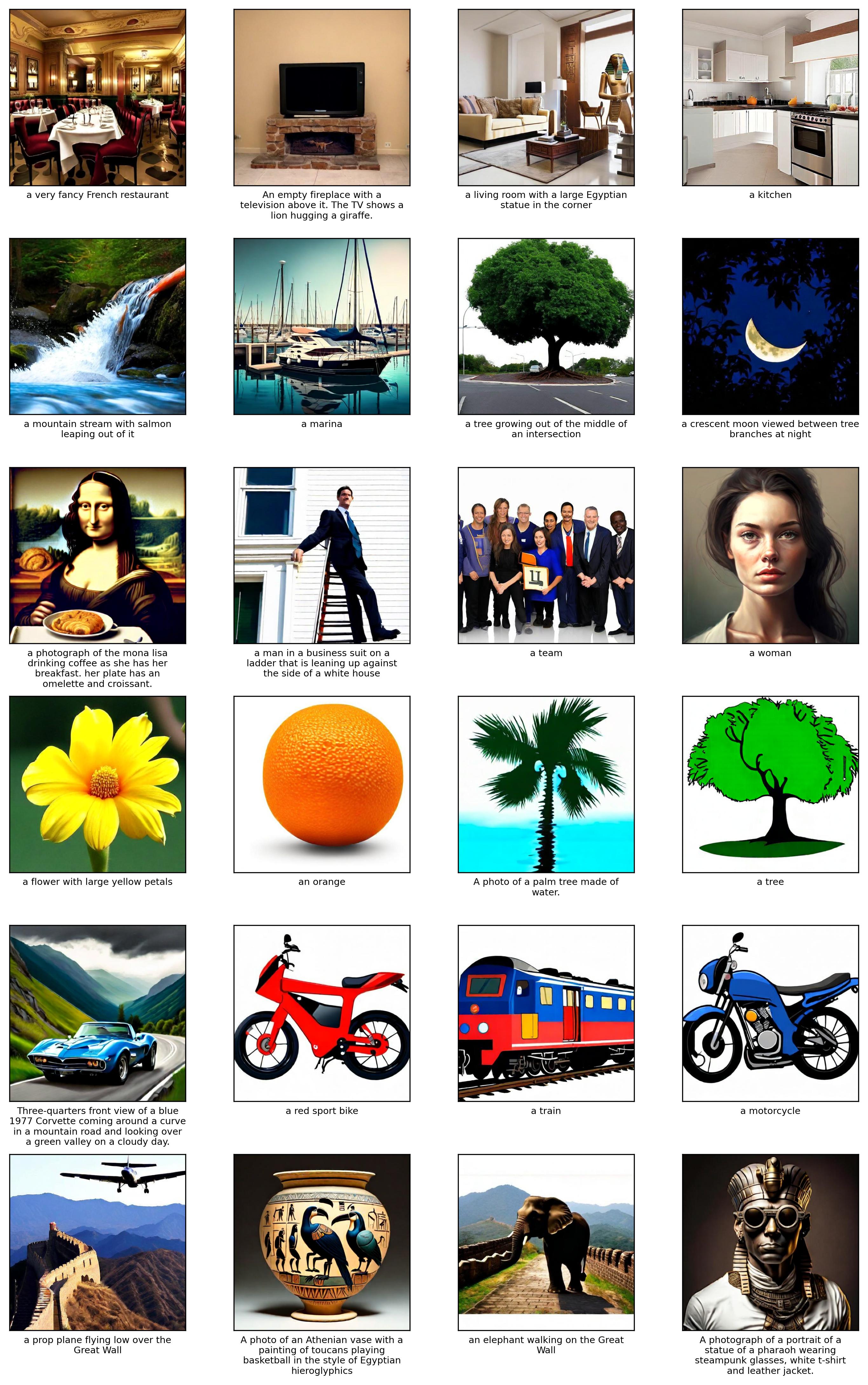}
    \caption{Synthesized images by our model using randomly selected prompts from PartiPrompts~\citep{yu2022Parti}. Rows correspond to following categories: Indoor scenes, Outdoor scenes, People, Produce \& Plants, Vehicles, and World knowledge.}
\end{figure}

\begin{figure}
    \centering
    \includegraphics[width=0.8\linewidth]{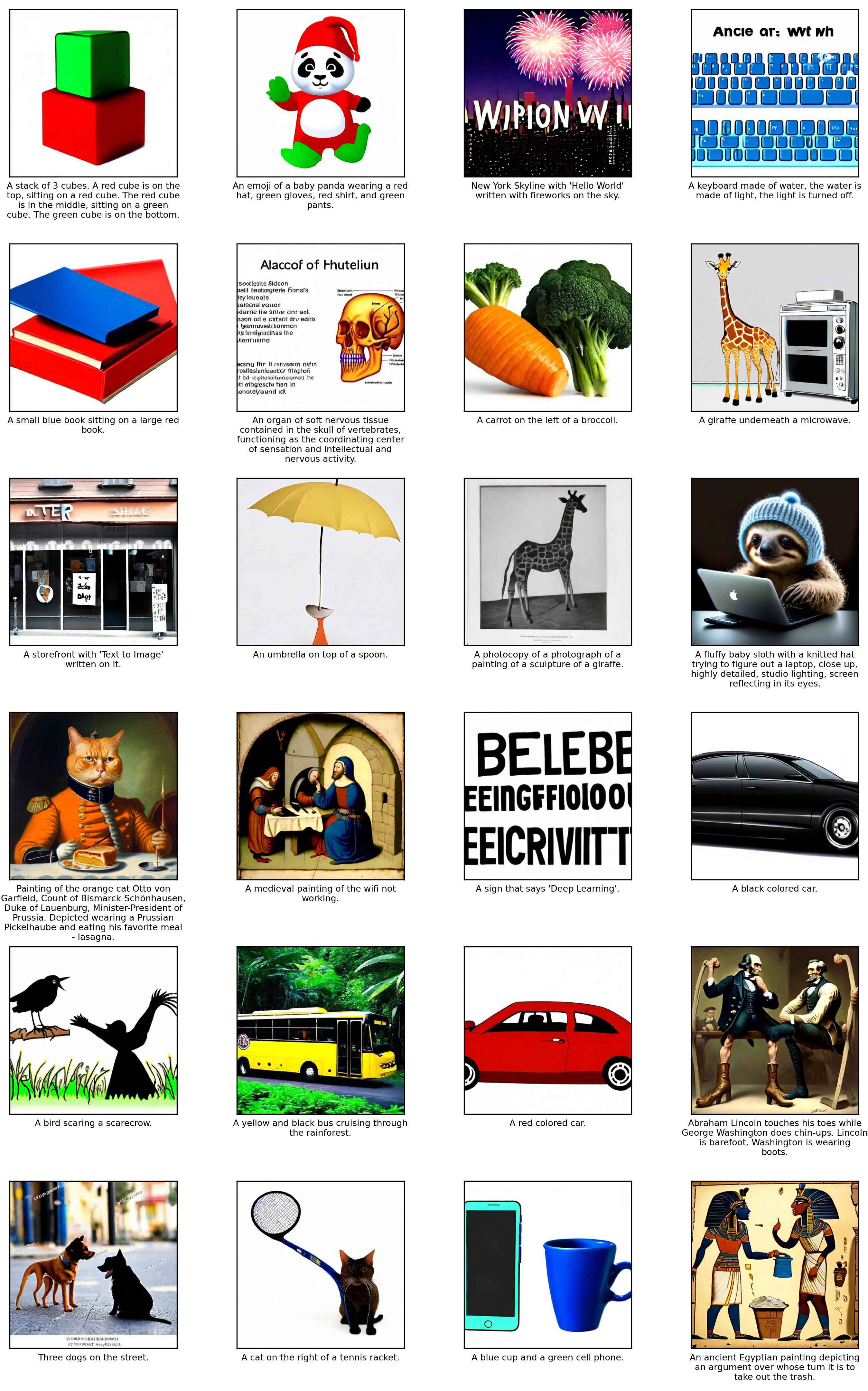}
    \caption{Synthesized images by our model using randomly selected prompts from DrawBench~\citep{saharia2022ImageN}.}
\end{figure}

\begin{figure}
    \centering
    \includegraphics[width=\linewidth]{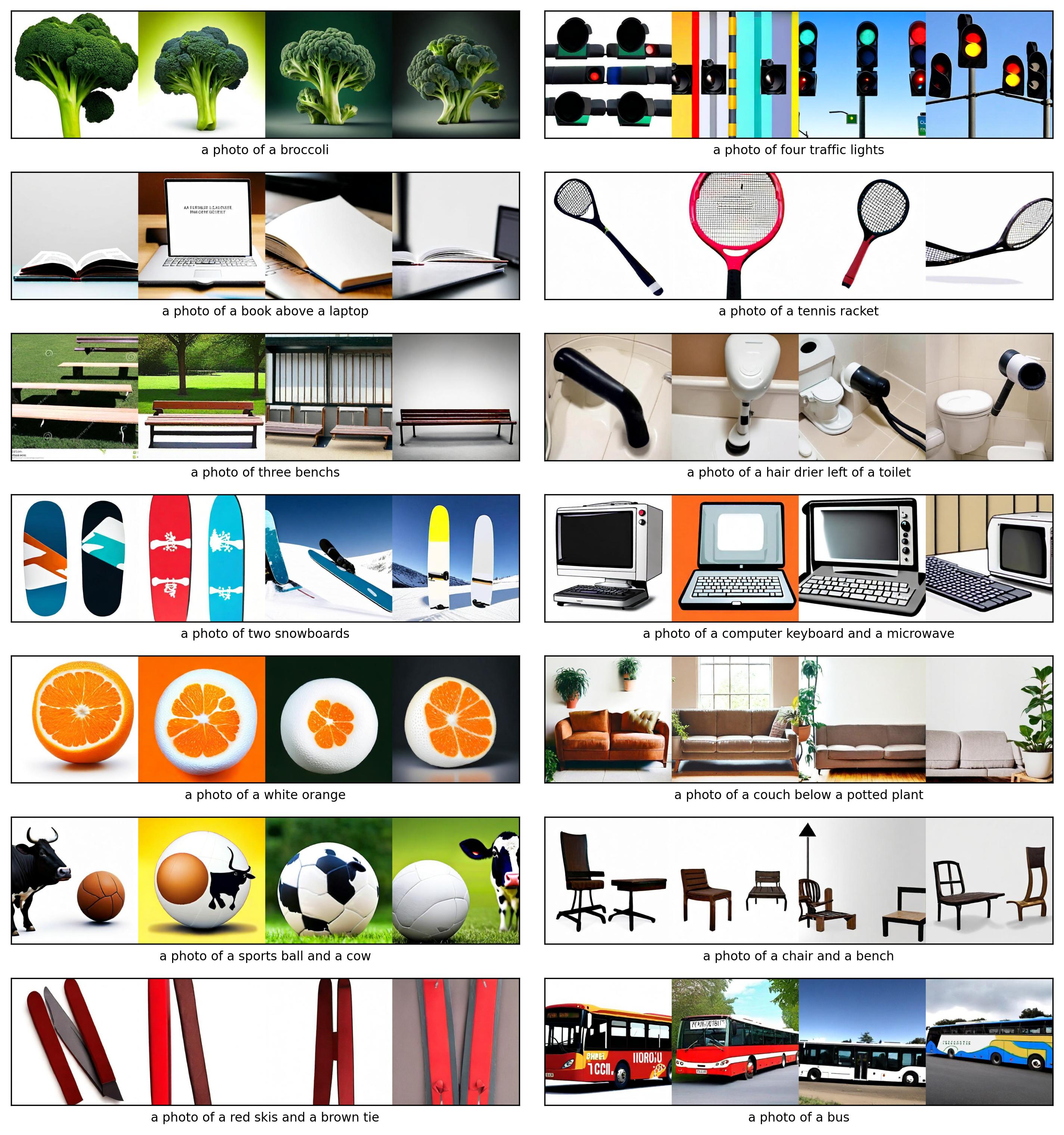}
    \caption{Synthesized images by our model using randomly selected prompts from GenEval~\citep{ghosh2024geneval}.}
\end{figure}

\begin{figure}[!htb]
    \centering
    \begin{subfigure}[b]{0.49\linewidth}
         \centering
         \includegraphics[width=\linewidth]{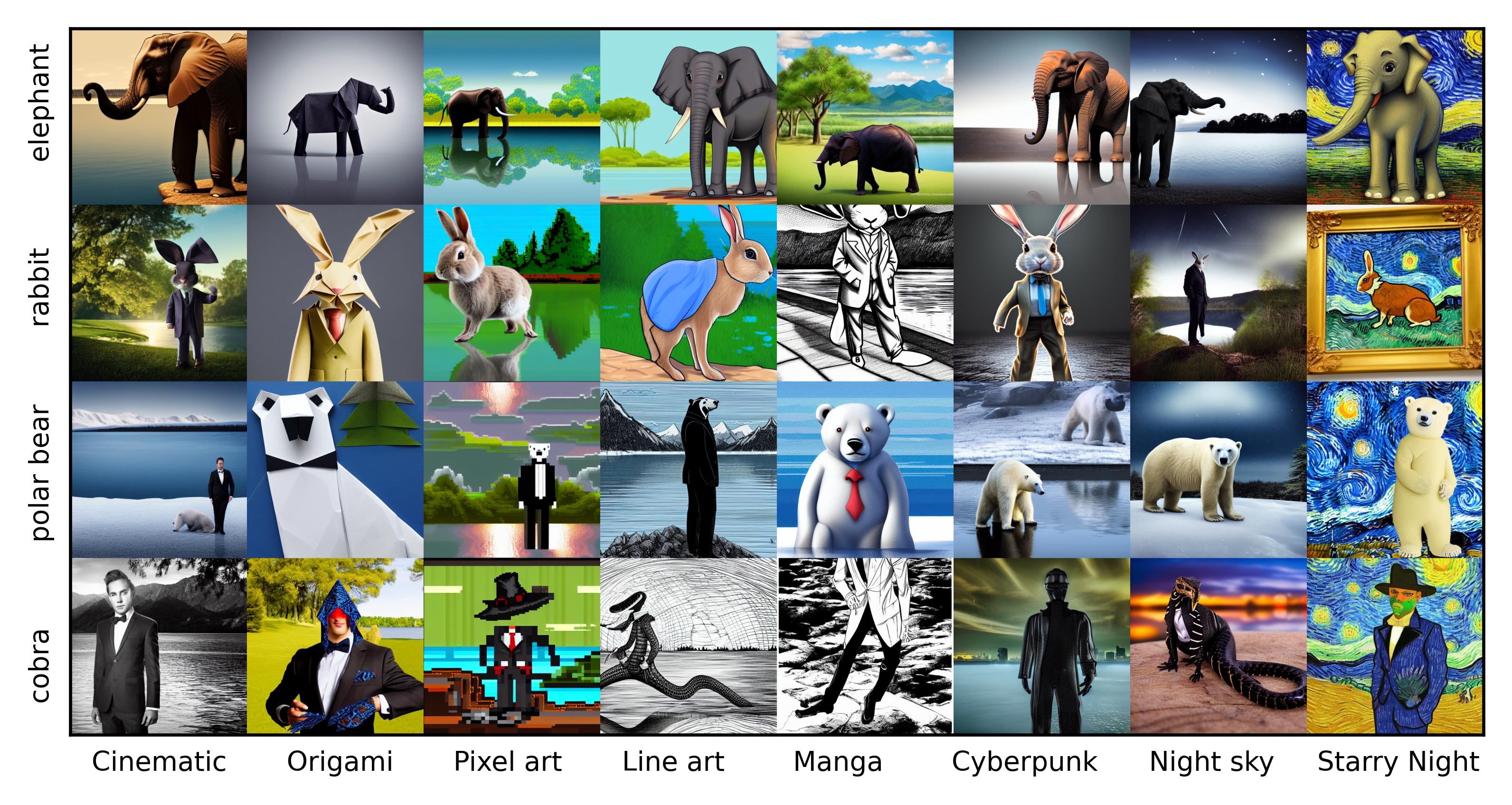}
         \caption{Stable-Diffusion-1.5}
    \end{subfigure}
    \hfill
    \begin{subfigure}[b]{0.49\linewidth}
         \centering
         \includegraphics[width=\linewidth]{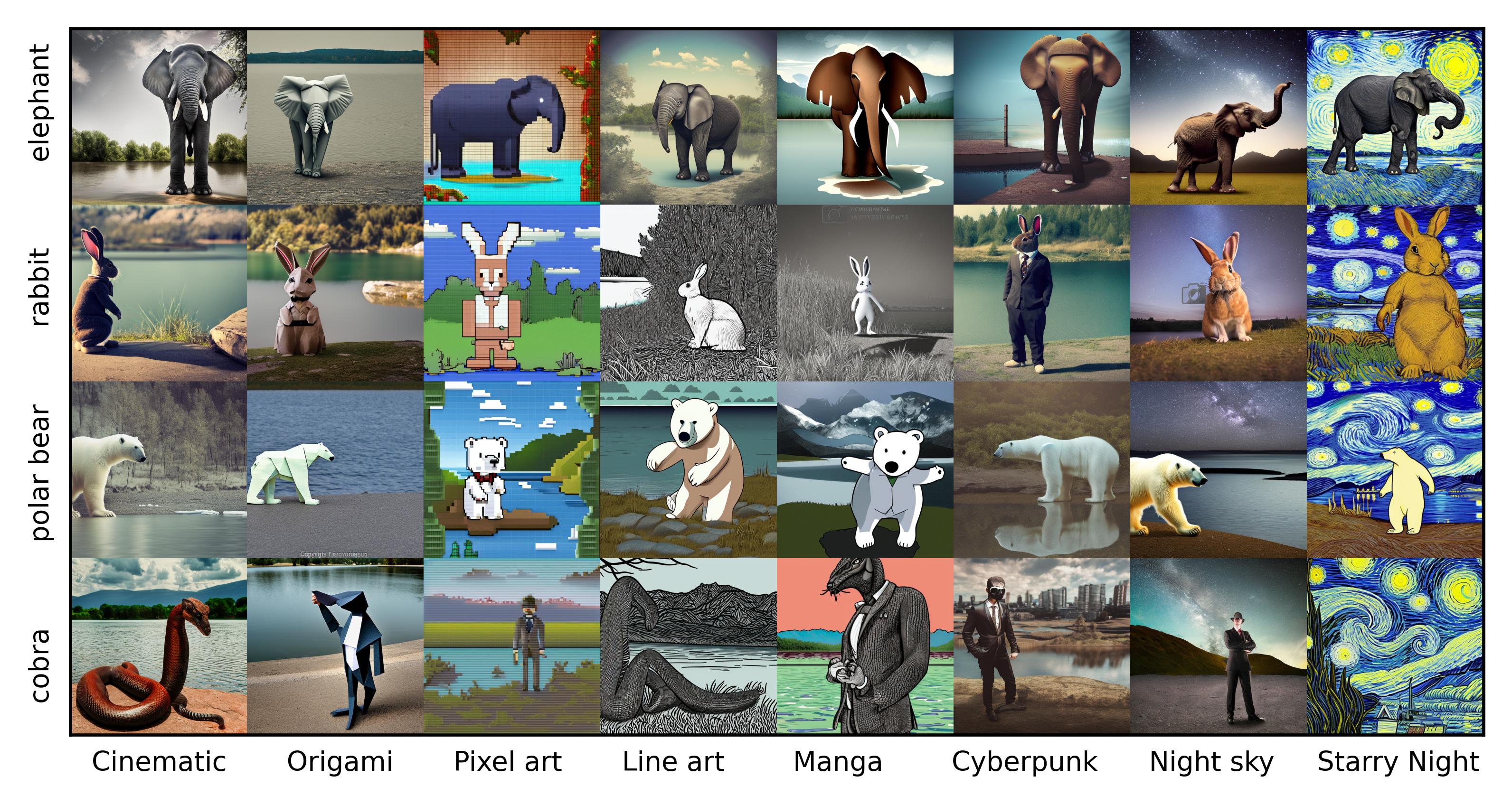}
         \caption{Stable-Diffusion-2.1}
    \end{subfigure}
    \begin{subfigure}[b]{0.49\linewidth}
         \centering
         \includegraphics[width=\linewidth]{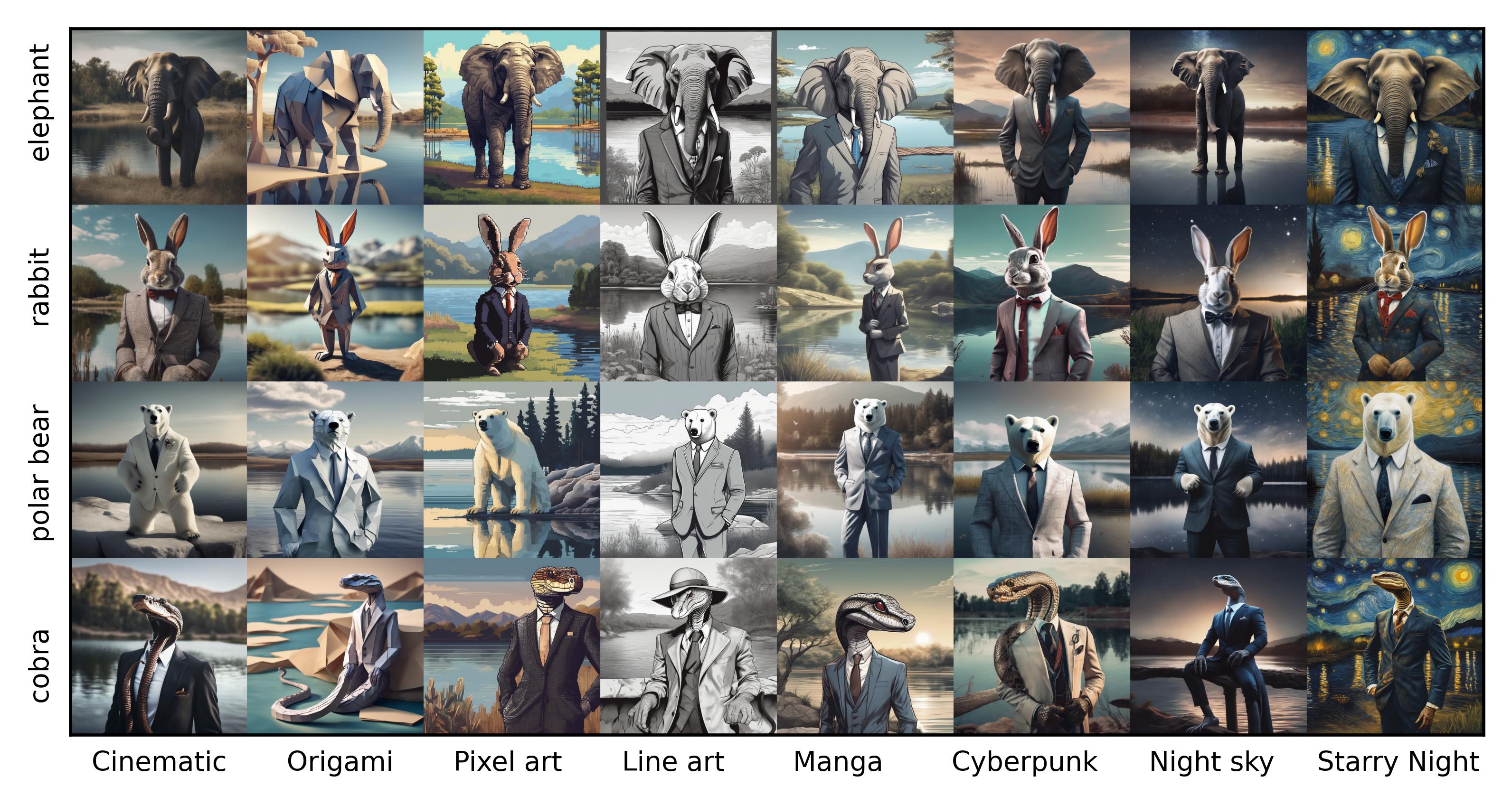}
         \caption{Stable-Diffusion-XL}
    \end{subfigure}
    \hfill
    \begin{subfigure}[b]{0.49\linewidth}
         \centering
         \includegraphics[width=\linewidth]{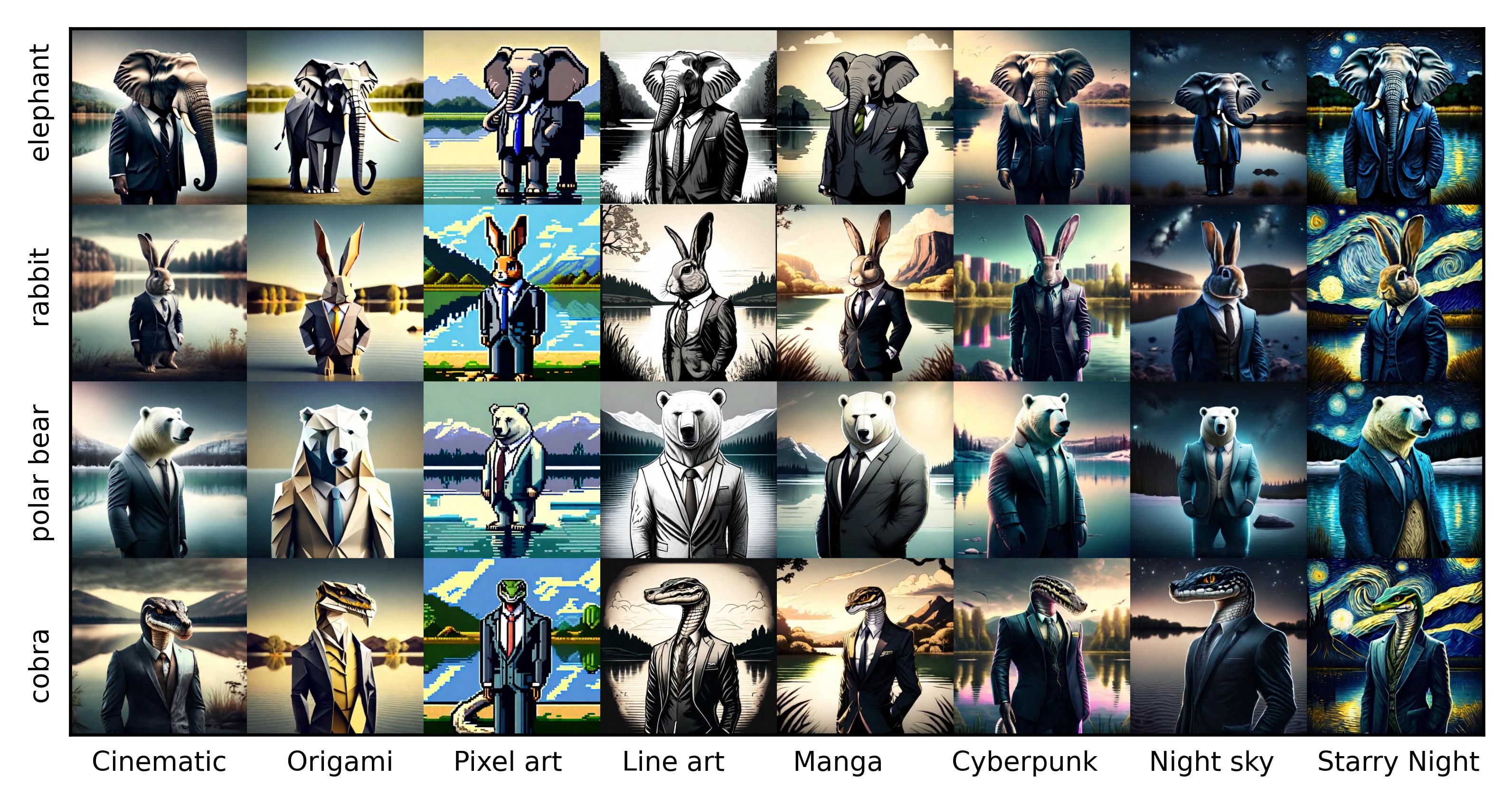}
         \caption{Ours}
    \end{subfigure}
     \caption{Evaluating the ability to generate diverse styles. \textit{Prompt}: A \rule{0.5cm}{0.15mm} dressed in a suit posing for a photo in \rule{0.5cm}{0.15mm} style. Natural lake landscape in background, detailed light and shadow, high detail.}
\end{figure}

\begin{figure}[!htb]
    \centering
    \begin{subfigure}[b]{0.49\linewidth}
         \centering
         \includegraphics[width=\linewidth]{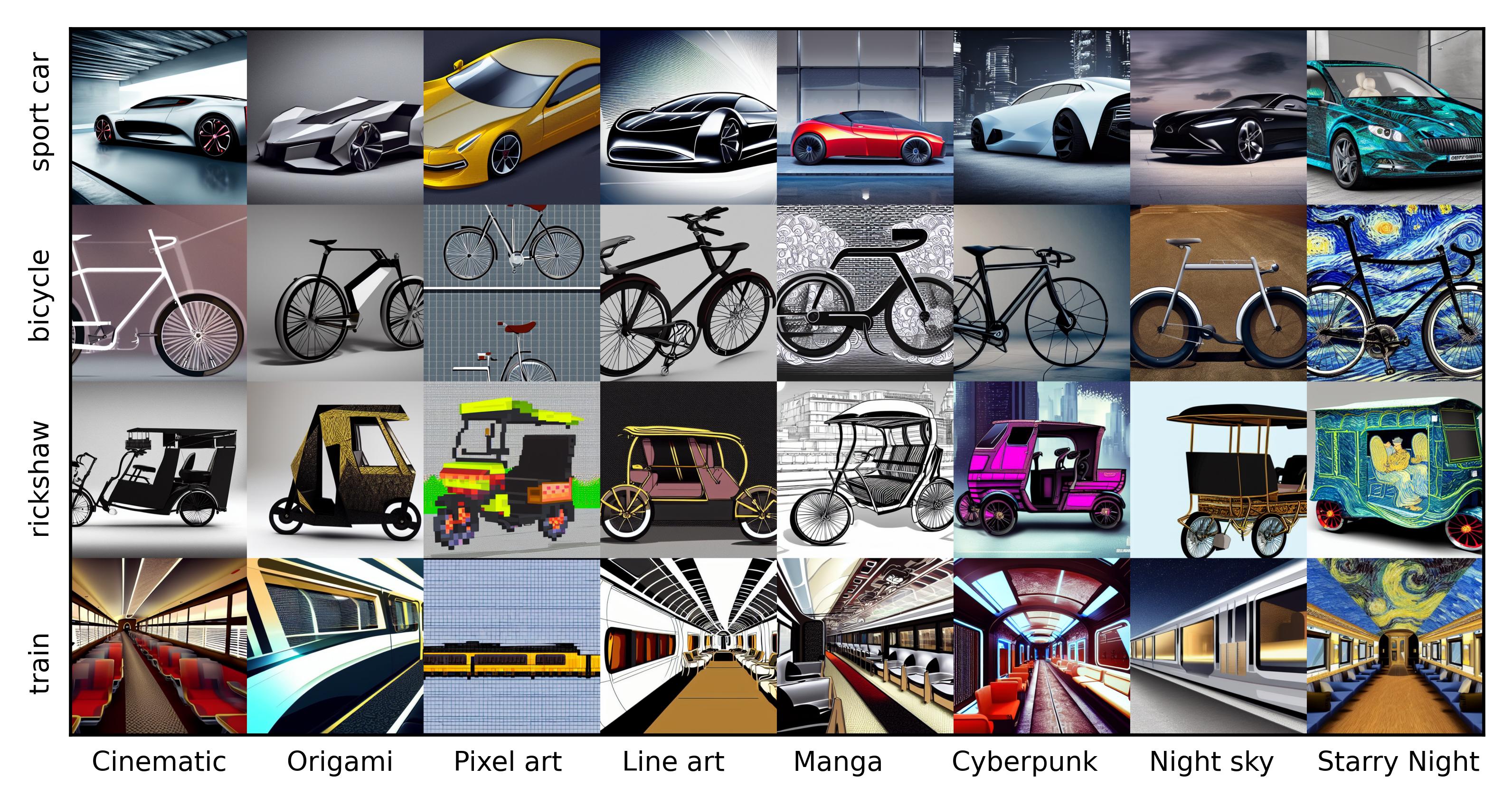}
         \caption{Stable-Diffusion-1.5}
    \end{subfigure}
    \hfill
    \begin{subfigure}[b]{0.49\linewidth}
         \centering
         \includegraphics[width=\linewidth]{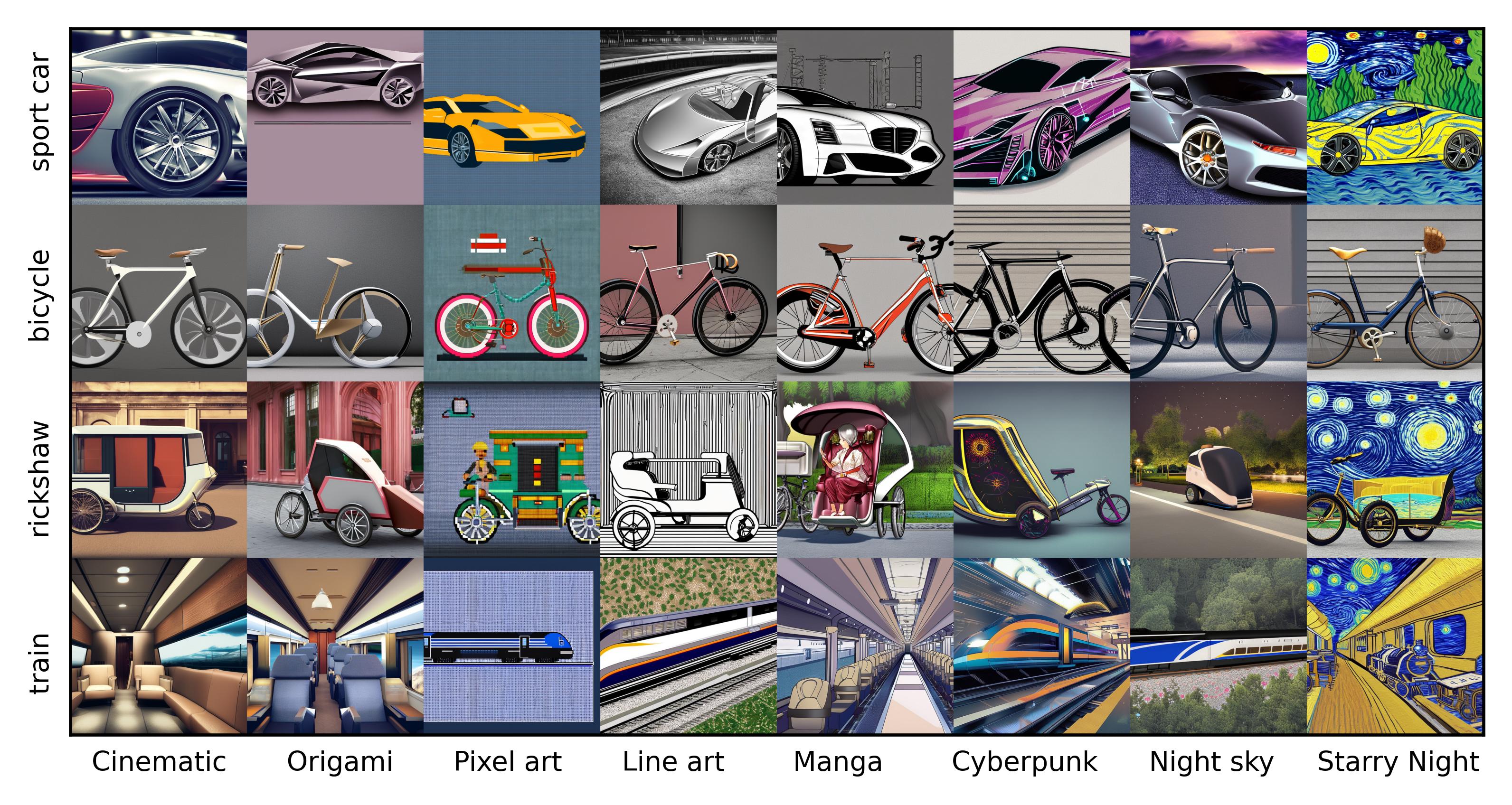}
         \caption{Stable-Diffusion-2.1}
    \end{subfigure}
    \begin{subfigure}[b]{0.49\linewidth}
         \centering
         \includegraphics[width=\linewidth]{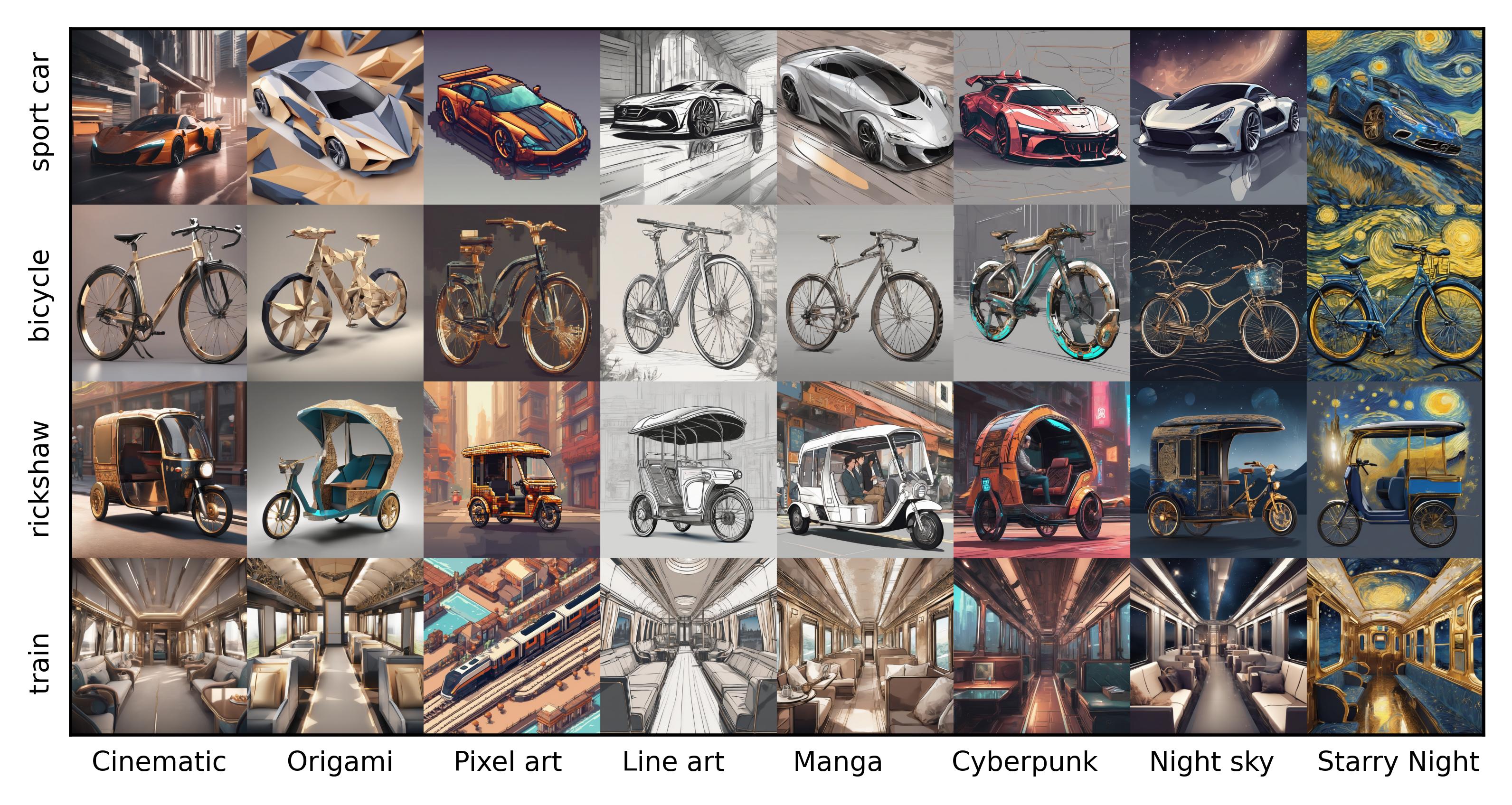}
         \caption{Stable-Diffusion-XL}
    \end{subfigure}
    \hfill
    \begin{subfigure}[b]{0.49\linewidth}
         \centering
         \includegraphics[width=\linewidth]{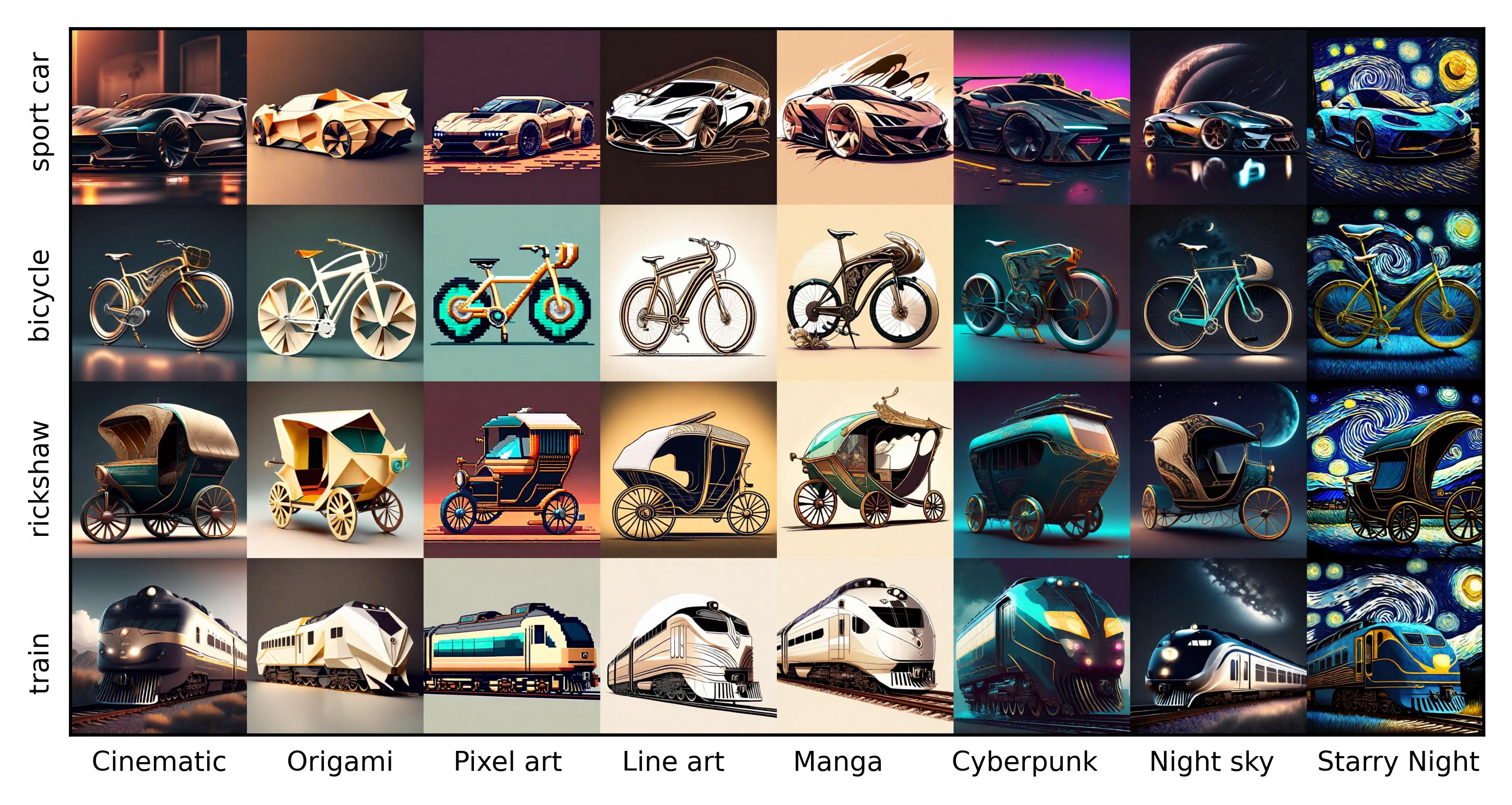}
         \caption{Ours}
    \end{subfigure}
     \caption{Evaluating the ability to generate diverse styles. \textit{Prompt}: A luxurious \rule{0.5cm}{0.15mm} with a next-generation modern design illustrated in \rule{0.5cm}{0.15mm} style.}
\end{figure}

\begin{figure}[!htb]
    \centering
    \begin{subfigure}[b]{0.49\linewidth}
         \centering
         \includegraphics[width=\linewidth]{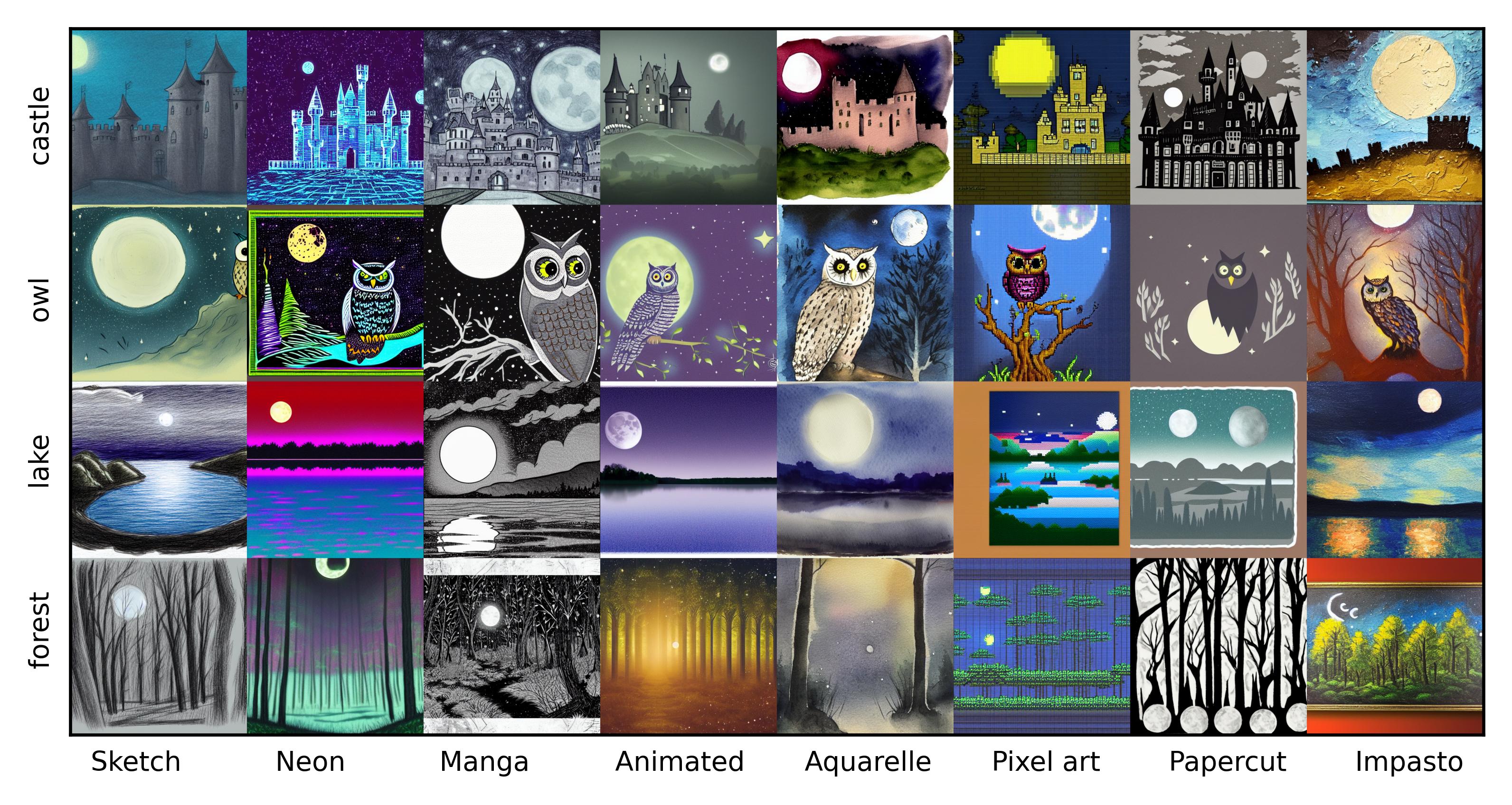}
         \caption{Stable-Diffusion-1.5}
    \end{subfigure}
    \hfill
    \begin{subfigure}[b]{0.49\linewidth}
         \centering
         \includegraphics[width=\linewidth]{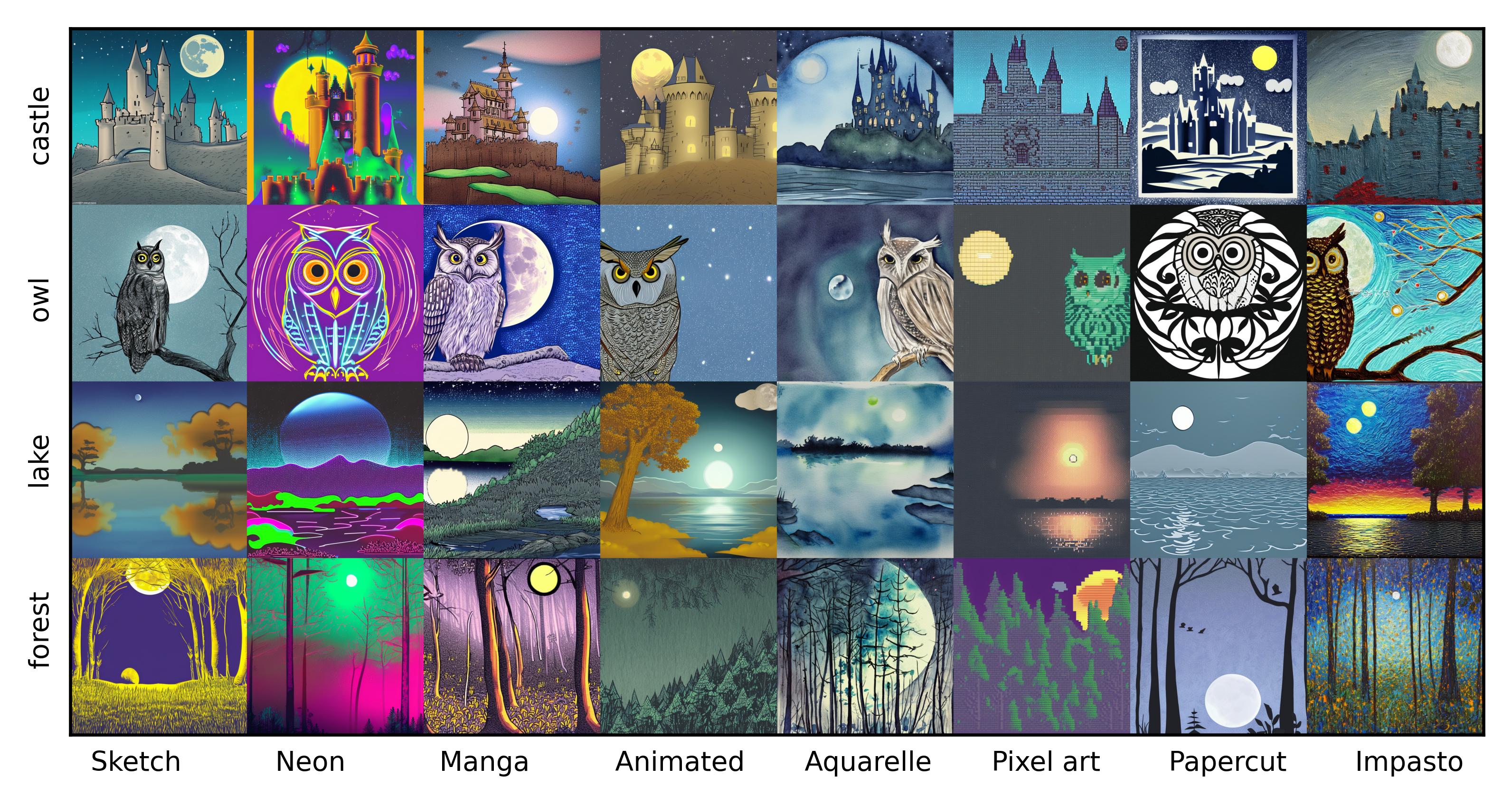}
         \caption{Stable-Diffusion-2.1}
    \end{subfigure}
    \begin{subfigure}[b]{0.49\linewidth}
         \centering
         \includegraphics[width=\linewidth]{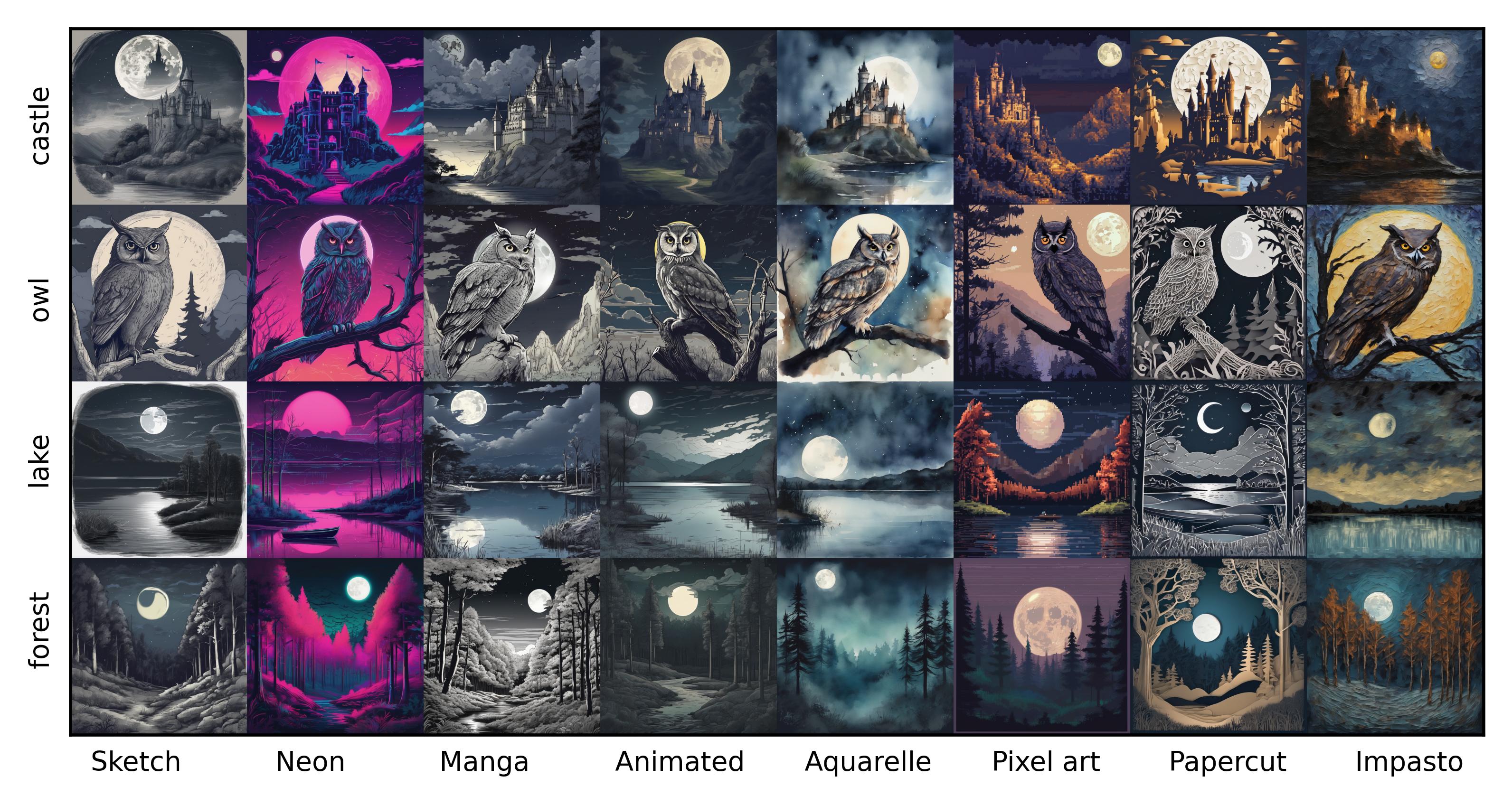}
         \caption{Stable-Diffusion-XL}
    \end{subfigure}
    \hfill
    \begin{subfigure}[b]{0.49\linewidth}
         \centering
         \includegraphics[width=\linewidth]{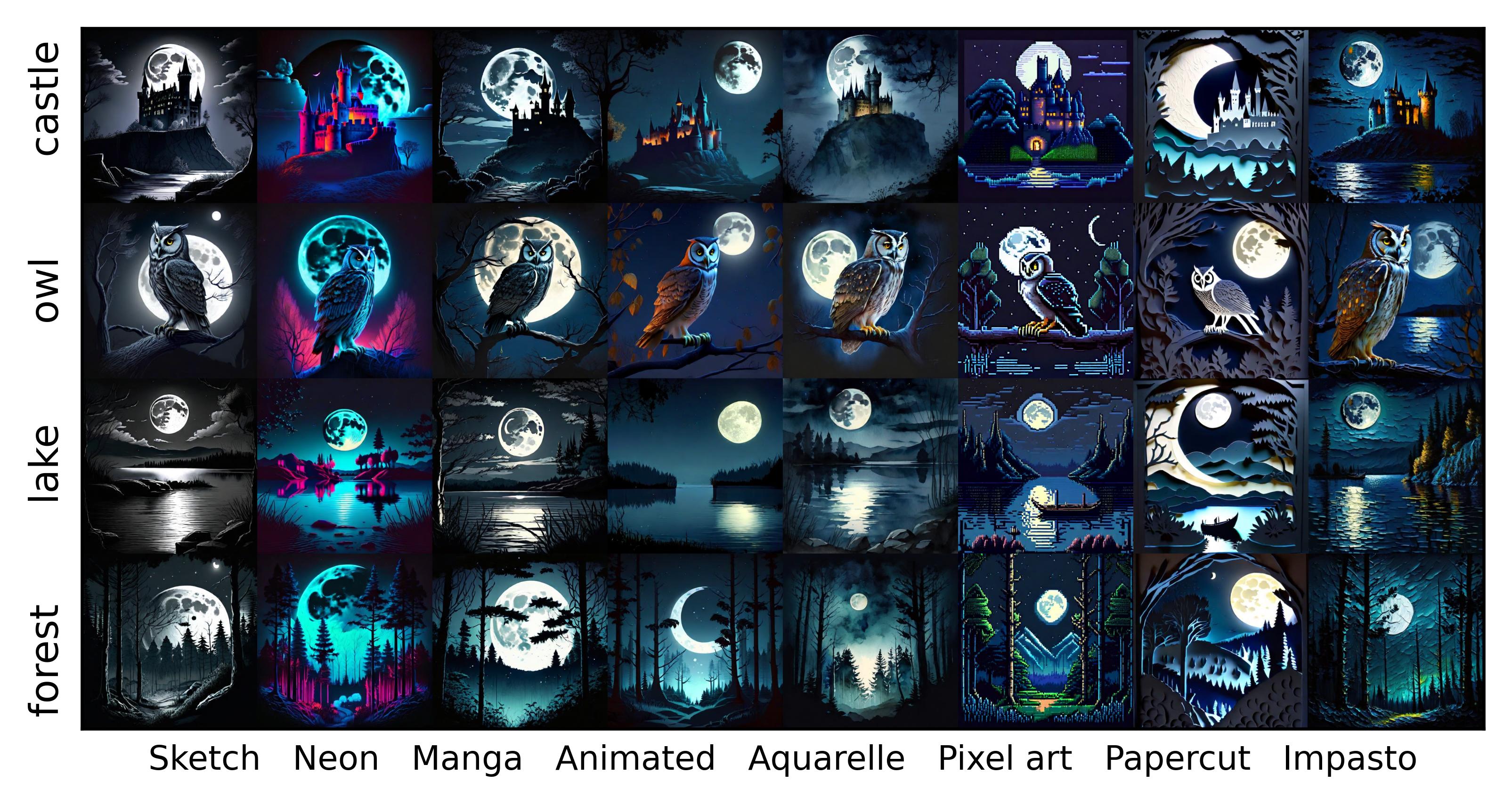}
         \caption{Ours}
    \end{subfigure}
     \caption{Evaluating the ability to generate diverse styles. \textit{Prompt}: A moonlit night over a \rule{0.5cm}{0.15mm}, mysteriously rendered in \rule{0.5cm}{0.15mm} style.}
\end{figure}

\begin{figure}[!htb]
    \centering
    \begin{subfigure}[b]{0.49\linewidth}
         \centering
         \includegraphics[width=\linewidth]{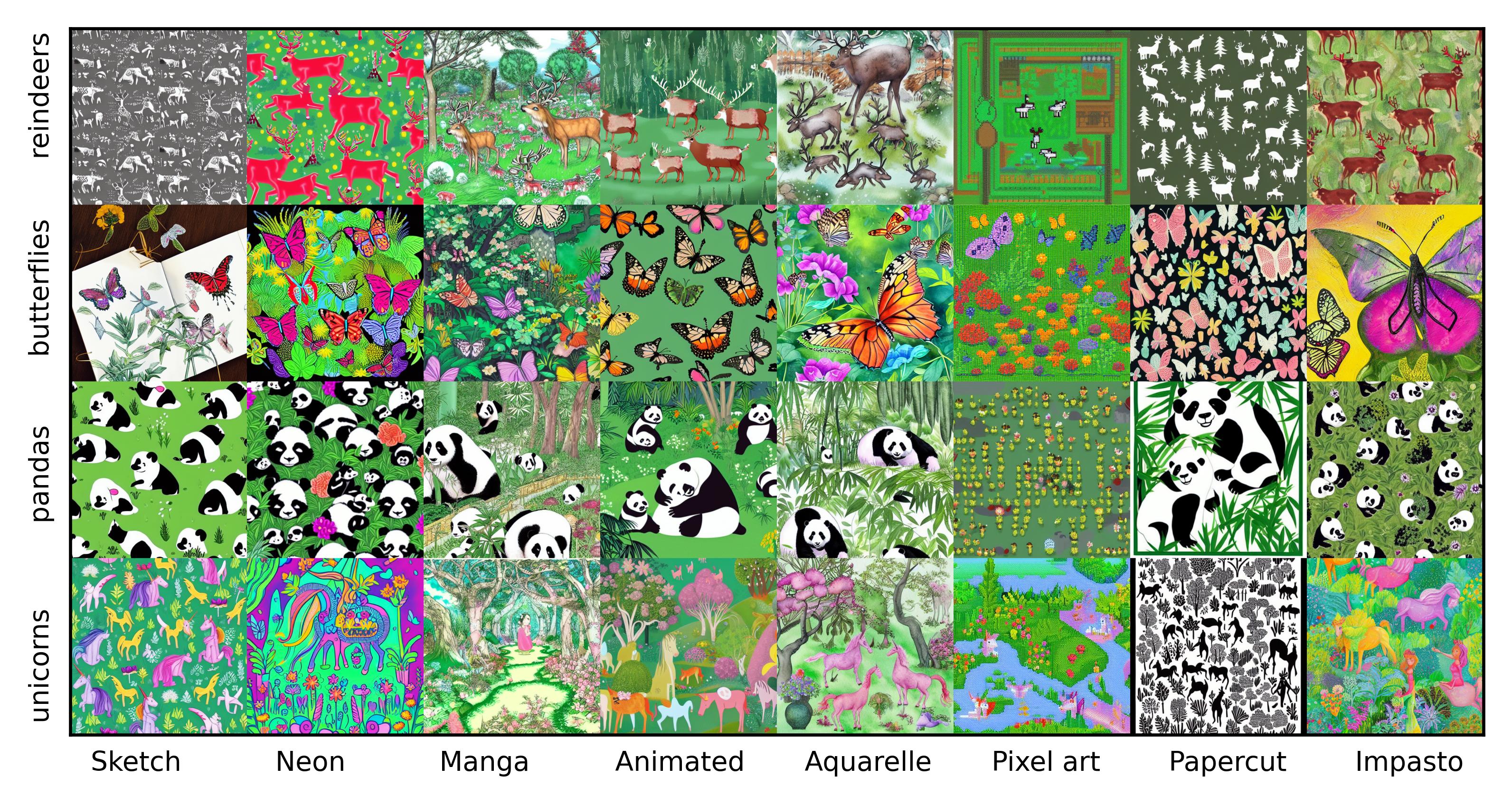}
         \caption{Stable-Diffusion-1.5}
    \end{subfigure}
    \hfill
    \begin{subfigure}[b]{0.49\linewidth}
         \centering
         \includegraphics[width=\linewidth]{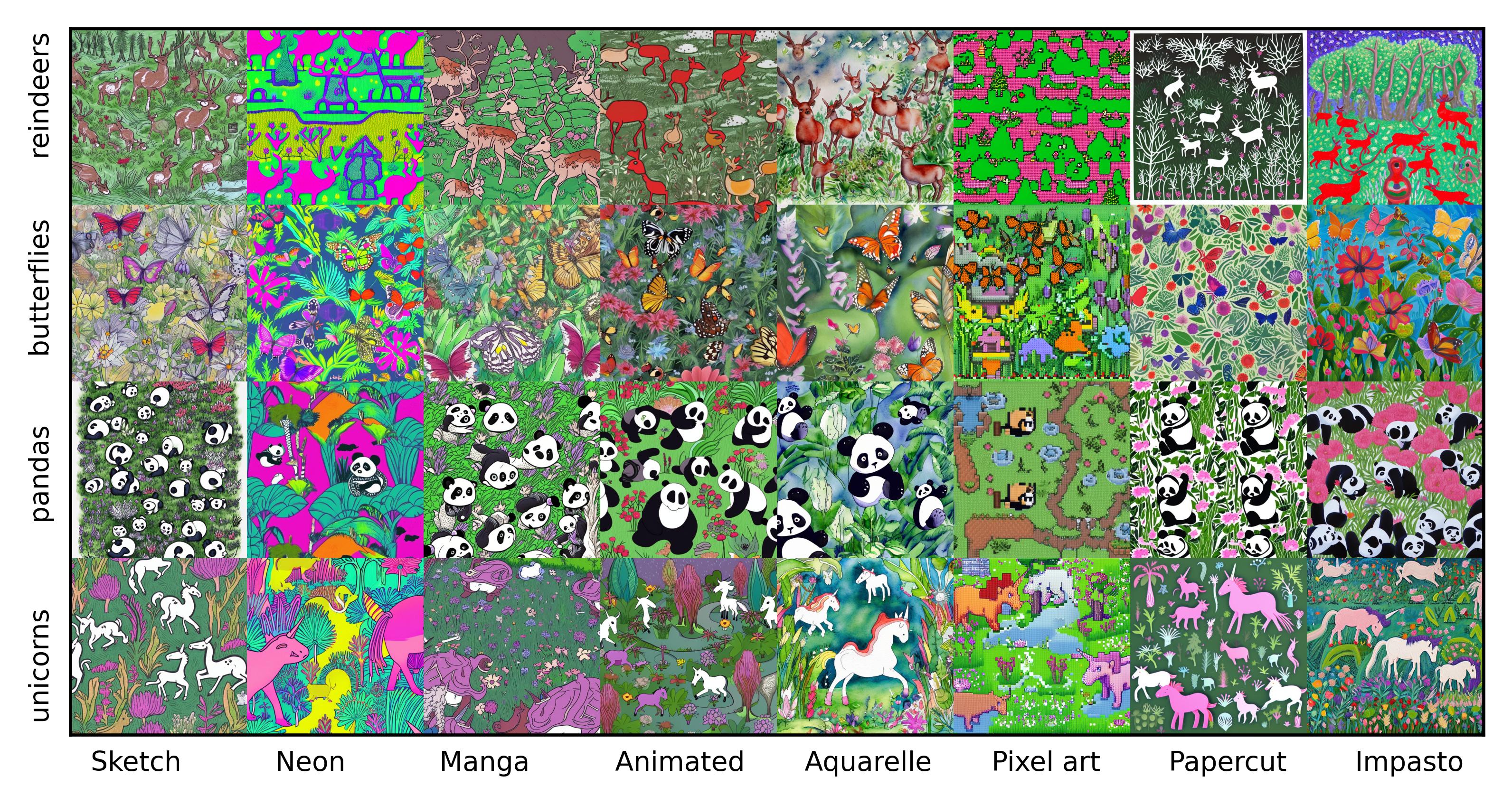}
         \caption{Stable-Diffusion-2.1}
    \end{subfigure}
    \begin{subfigure}[b]{0.49\linewidth}
         \centering
         \includegraphics[width=\linewidth]{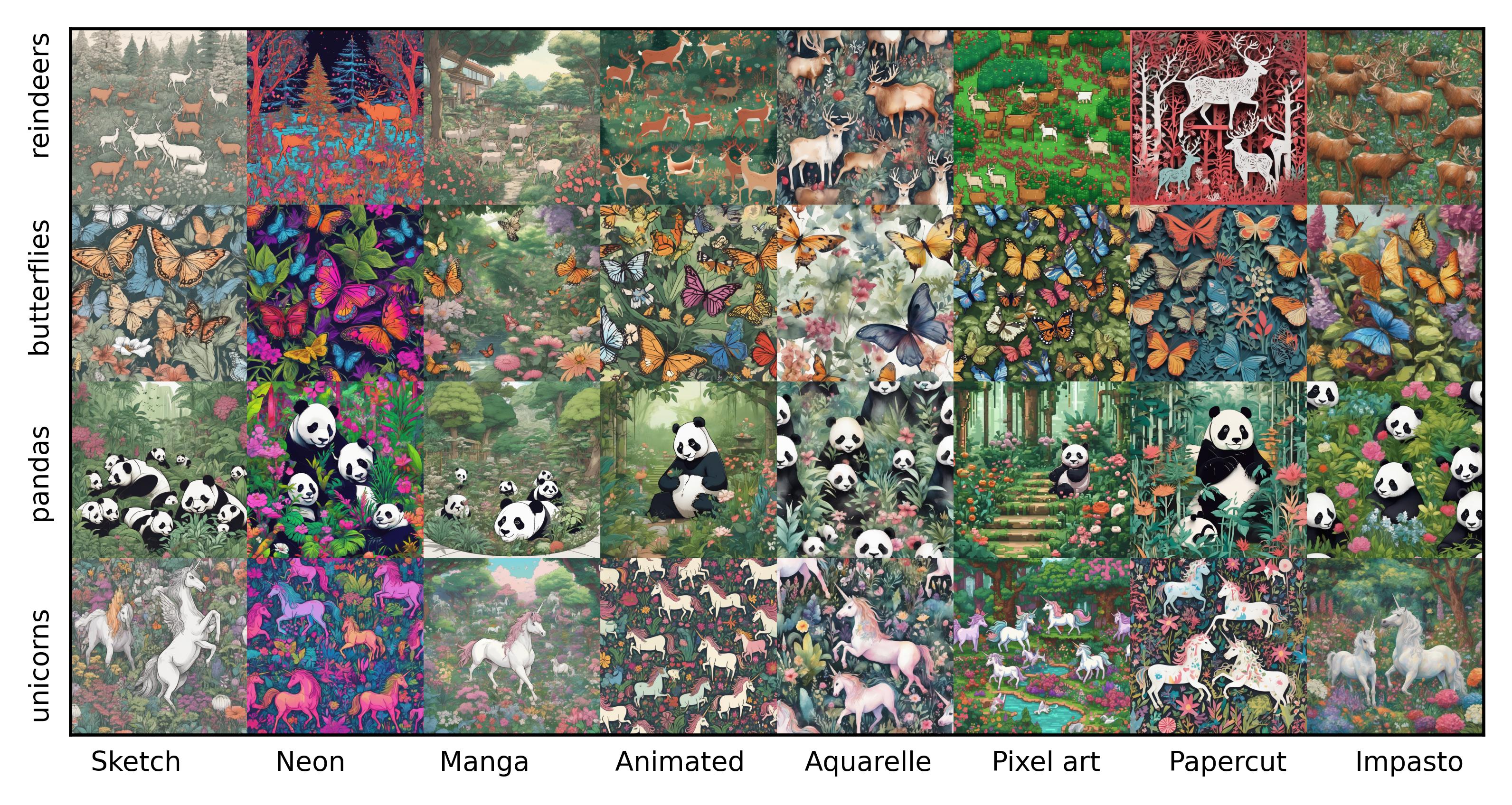}
         \caption{Stable-Diffusion-XL}
    \end{subfigure}
    \hfill
    \begin{subfigure}[b]{0.49\linewidth}
         \centering
         \includegraphics[width=\linewidth]{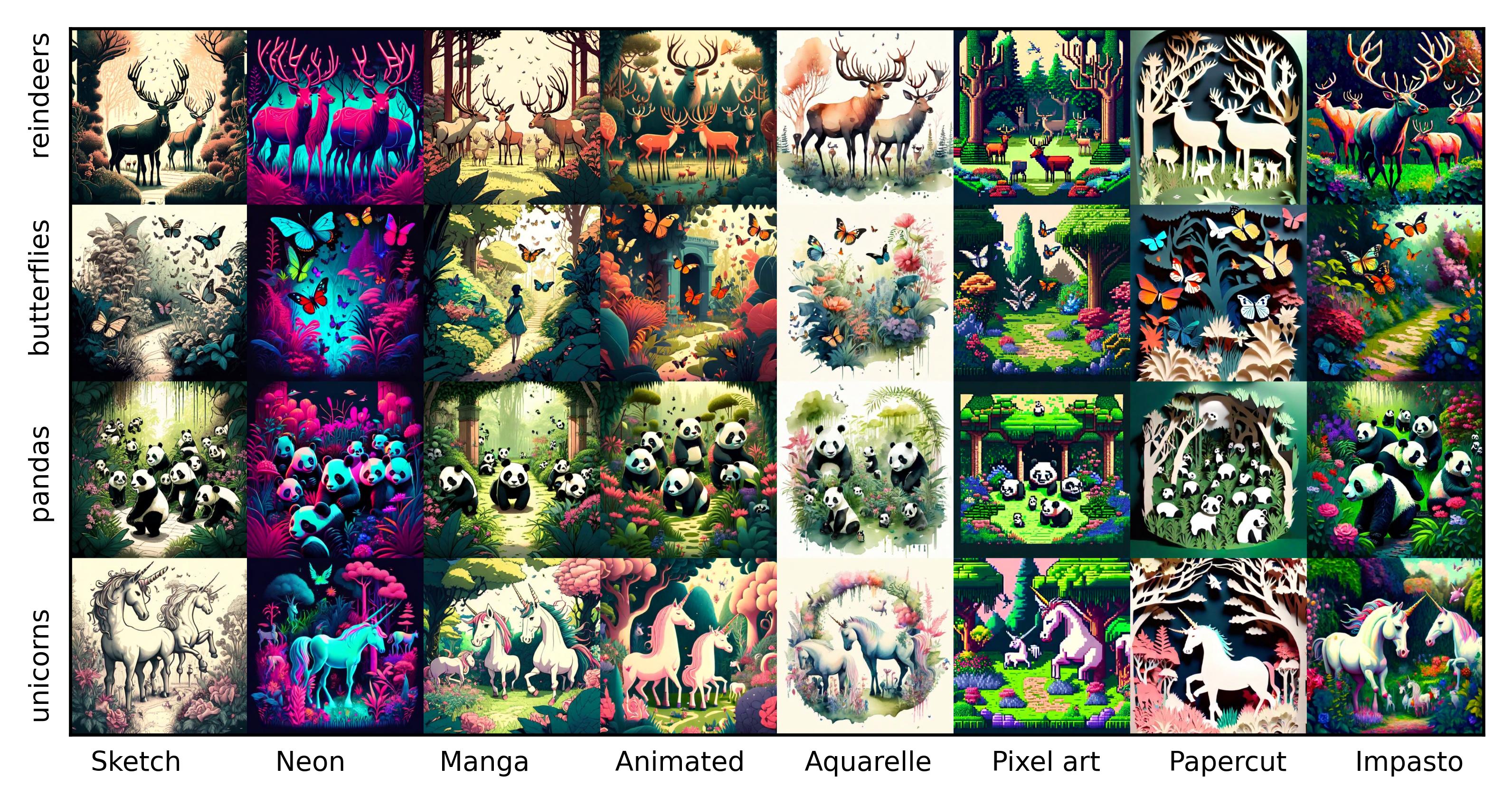}
         \caption{Ours}
    \end{subfigure}
     \caption{Evaluating the ability to generate diverse styles. \textit{Prompt}: A lush garden full of \rule{0.5cm}{0.15mm}, vividly illustrated in \rule{0.5cm}{0.15mm} style.}
\end{figure}

\begin{figure}
    \centering
    \includegraphics[width=\linewidth]{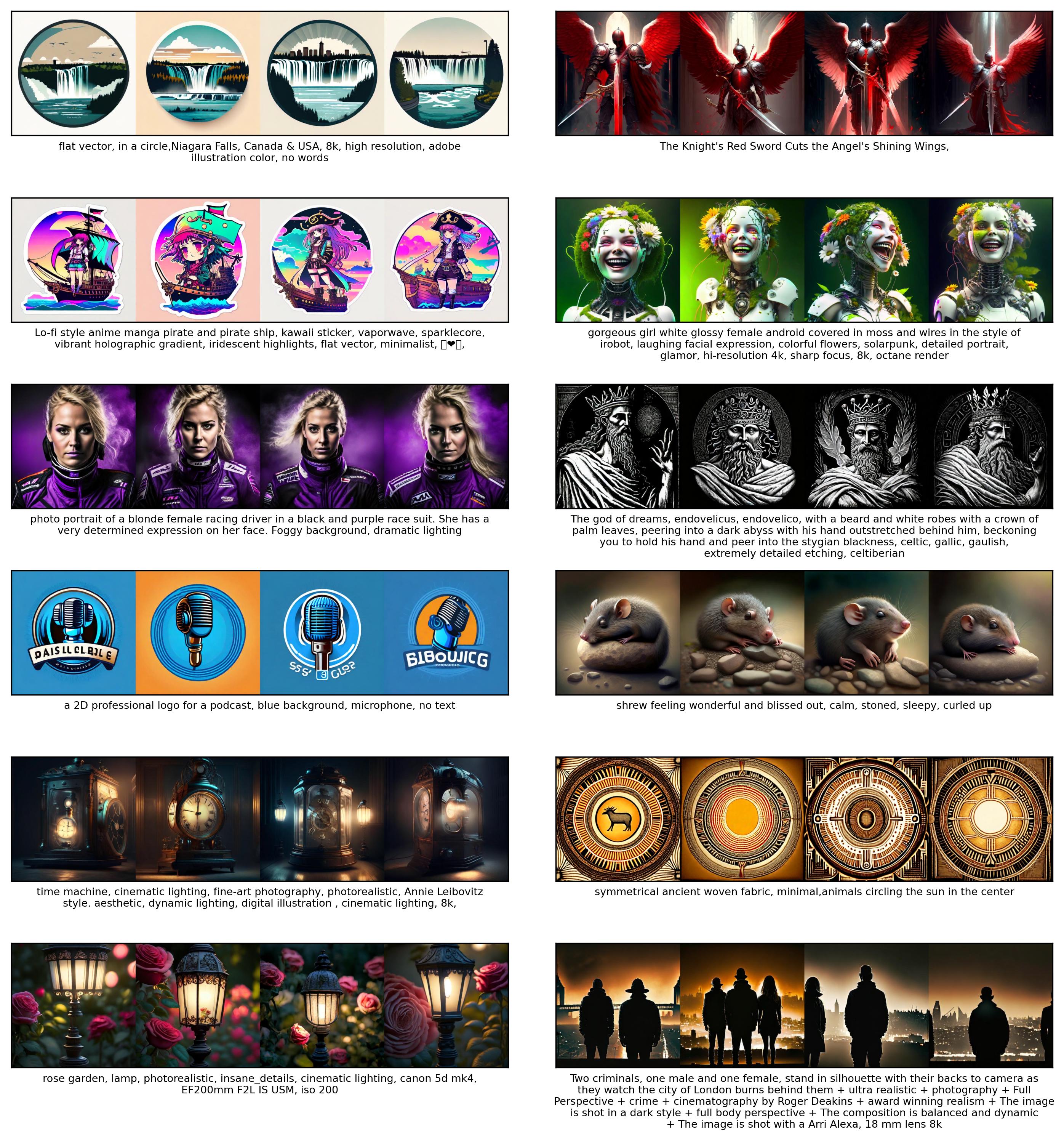}
    \caption{Synthesized images by our model using randomly selected prompts from the test set of JourneyDB dataset~\citep{ghosh2024geneval}.}
\end{figure}

\begin{figure}
    \centering
    \includegraphics[width=\linewidth]{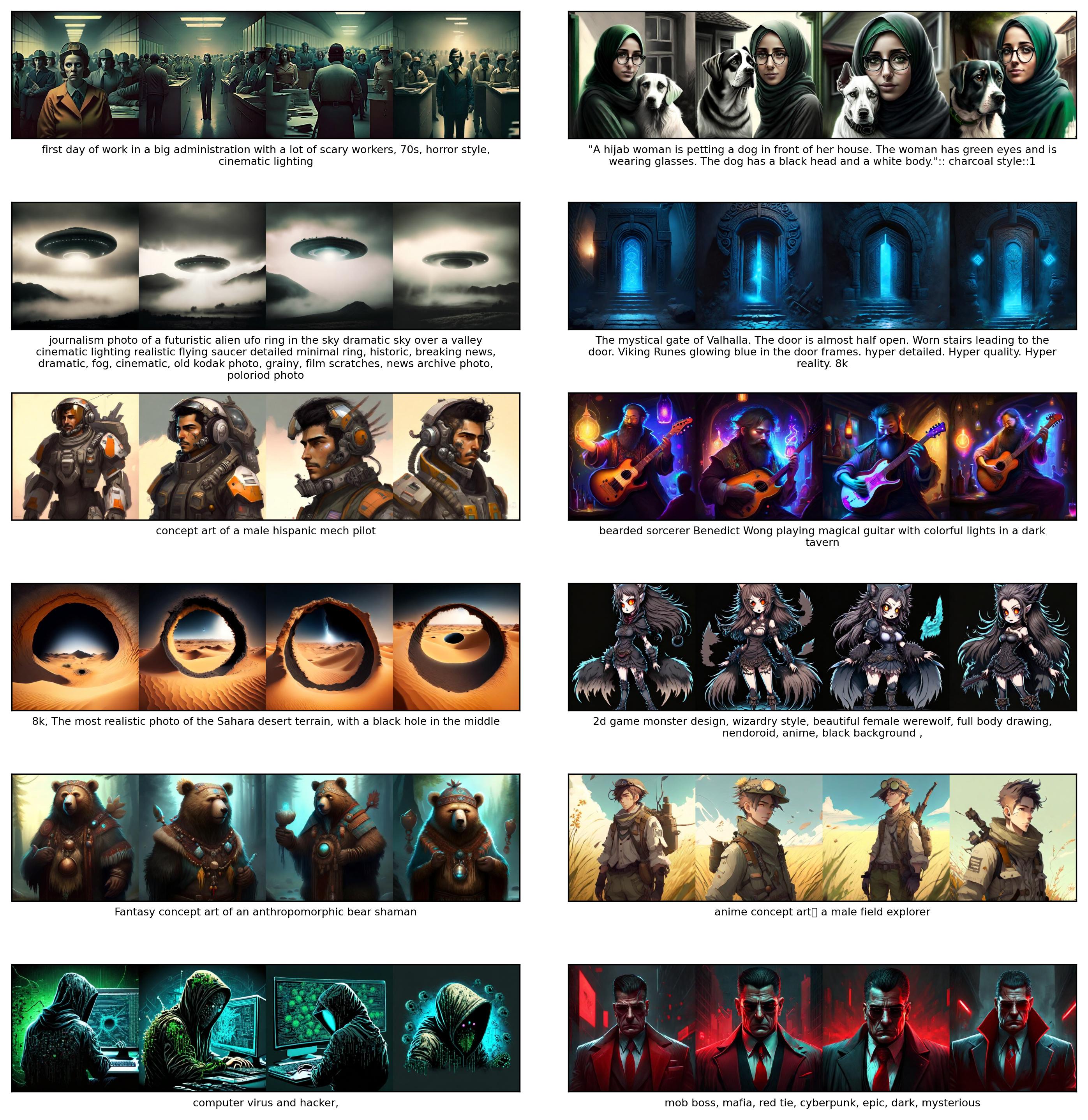}
    \caption{Synthesized images by our model using randomly selected prompts from the test set of JourneyDB dataset~\citep{ghosh2024geneval}.}
\end{figure}

\begin{figure}
    \centering
    \includegraphics[width=\linewidth]{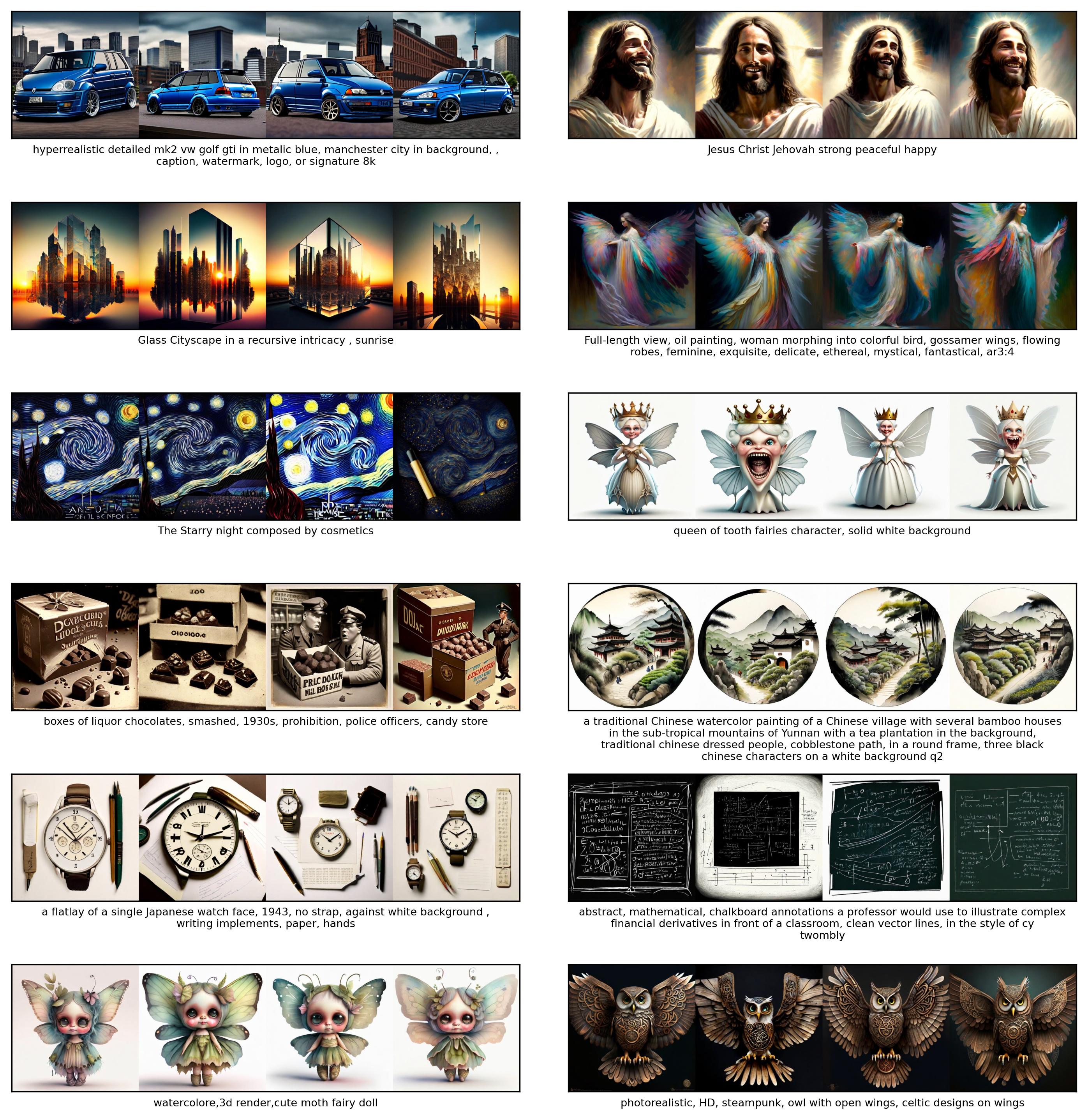}
    \caption{Synthesized images by our model using randomly selected prompts from the test set of JourneyDB dataset~\citep{ghosh2024geneval}.}
\end{figure}

\begin{figure}
    \centering
    \includegraphics[width=\linewidth]{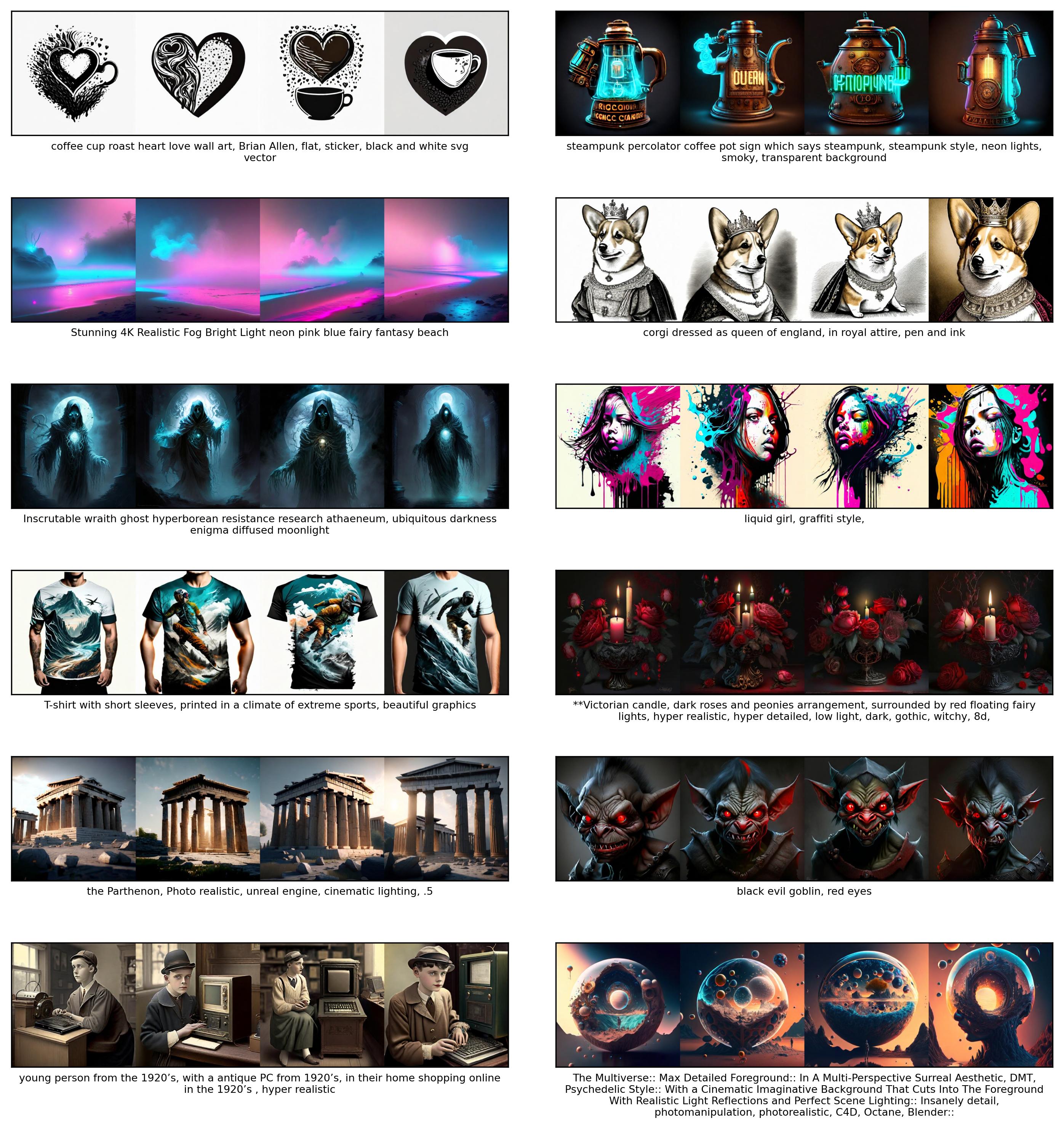}
    \caption{Synthesized images by our model using randomly selected prompts from the test set of JourneyDB dataset~\citep{ghosh2024geneval}.}
\end{figure}

\begin{figure}
    \centering
    \includegraphics[width=0.85\linewidth]{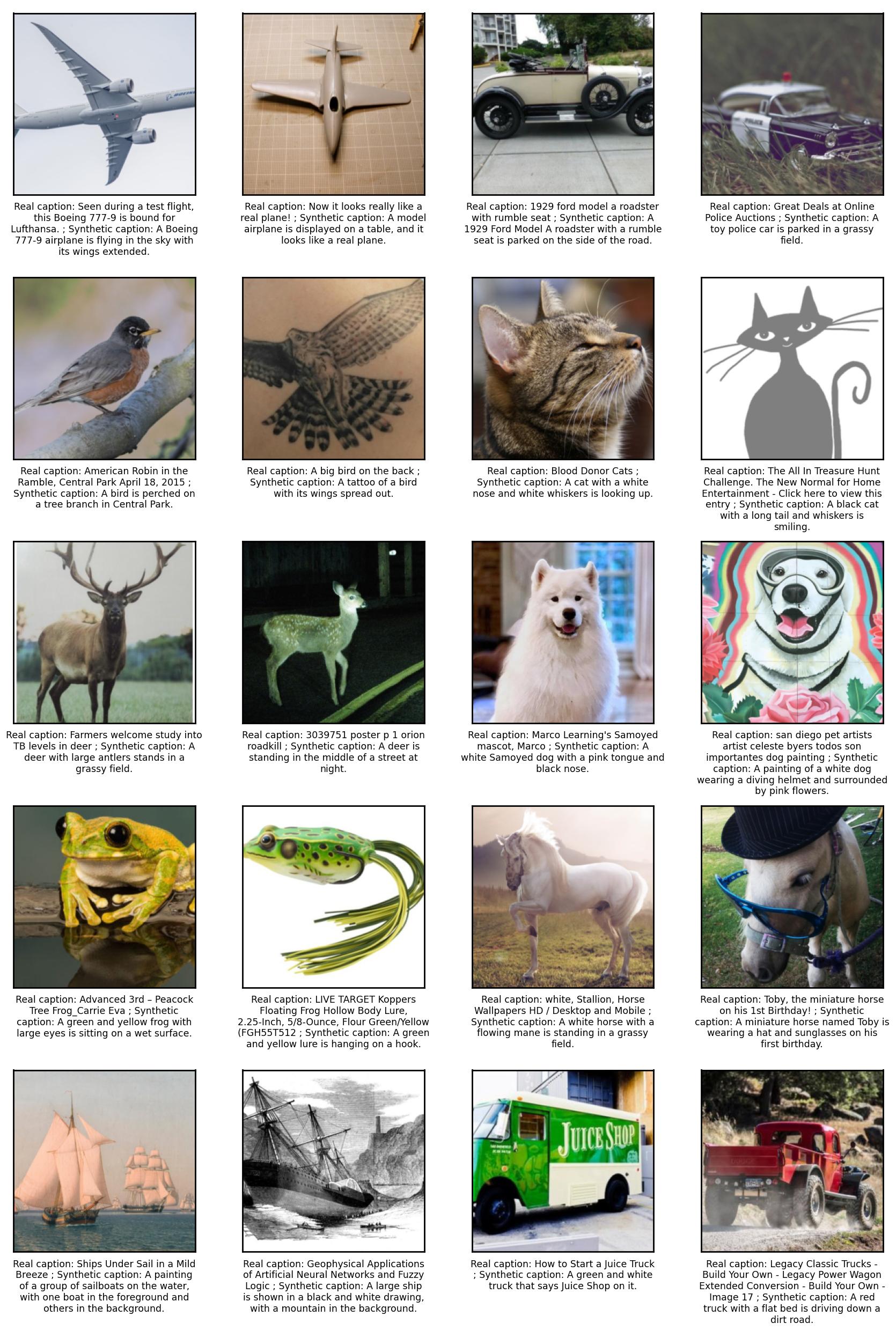}
    \caption{Selected images with corresponding real and synthetic captions from the \cifarcaps dataset. We created the \cifarcaps dataset, imitating the widely used CIFAR-10 dataset~\citep{krizhevsky2009cifar10}, to enable small scale experimentation on the text-to-image generative models.}
    \label{fig: cifar_cap_examples}
\end{figure}

\end{document}